%% file: main.tex
\documentclass{article} %
\usepackage{iclr2024_conference,times}
\usepackage[pagebackref=true,breaklinks=true,colorlinks,bookmarks=false,citecolor=blue,linkcolor=blue]{hyperref}
\usepackage[utf8]{inputenc} %
\usepackage[T1]{fontenc}    %
\usepackage{multirow}
\usepackage{url}            %
\usepackage{booktabs}       %
\usepackage{amsfonts}       %
\usepackage{nicefrac}       %
\usepackage{microtype}      %
\usepackage{xcolor}         %
\usepackage{graphicx}
\usepackage{xcolor}
\usepackage{algorithm}
\usepackage{algpseudocode}
\usepackage{listings}
\usepackage{subcaption}
\usepackage{amsthm}
\input{math_commands}

\input{defns}
\input{macros}

\usepackage[conf={restate,no link to proof, text proof=Proof}]{proof-at-the-end}

\title{Training-free linear image inverses via flows}

\author{Ashwini Pokle$^{1}$ \quad Matthew J. Muckley$^{2}$ \quad  Ricky T. Q. Chen$^{2}$ \quad Brian Karrer$^{2}$ \\
$^1$Carnegie Mellon University \quad $^2$ FAIR at Meta \\
\texttt{apokle@andrew.cmu.edu}, \texttt{\{mmuckley, rtqichen, briankarrer\}@meta.com}    
}

\iclrfinalcopy %
\begin{document}

\maketitle

\input{sections/abstract}
\input{sections/introduction}
\input{sections/background}

\input{sections/methods}

\input{sections/results}
\input{sections/related_work}

\input{sections/conclusion}

\bibliography{iclr2024_conference}
\bibliographystyle{iclr2024_conference}

\appendix
\input{sections/appendix}

\end{document}

%% file: math_commands.tex
\usepackage{amsmath,amsfonts,bm}

\def\eqref#1{equation~\ref{#1}}

\def\1{\bm{1}}

\def\rvy{{\mathbf{y}}}

\def\vg{{\bm{g}}}

\def\vu{{\bm{u}}}
\def\vv{{\bm{v}}}

\def\vx{{\bm{x}}}
\def\vy{{\bm{y}}}
\def\vz{{\bm{z}}}

\def\mA{{\bm{A}}}

\def\mI{{\bm{I}}}

\def\mX{{\bm{X}}}

\DeclareMathAlphabet{\mathsfit}{\encodingdefault}{\sfdefault}{m}{sl}
\SetMathAlphabet{\mathsfit}{bold}{\encodingdefault}{\sfdefault}{bx}{n}

\newcommand{\E}{\mathbb{E}}

%% file: macros.tex
\usepackage[capitalize]{cleveref}
\crefname{equation}{Eq.}{Eq.}
\crefname{figure}{Figure}{Figure~}
\crefname{table}{Table}{Table~}
\crefname{section}{Sec.}{Sec.~}
\crefname{algorithm}{Algorithm}{Algorithm~}
\crefname{thm}{Theorem}{Theorem~}
\crefname{lemma}{Lemma}{Lemma~}
\crefname{appendix}{Appendix}{Appendix~}

\def\ie{\textit{i.e.,~}}
\def\iid{\textit{i.i.d.~}}
\def\eg{\textit{e.g.,~}}

\definecolor{prune}{rgb}{0.44, 0.11, 0.11}
\definecolor{myblue}{rgb}{0, .5, 1}
\definecolor{maroon}{rgb}{0.5450, 0, 0}
\definecolor{darkred}{rgb}{0.5450, 0, 0}
\definecolor{RoyalBlue}{RGB}{0,100,170}
\definecolor{DarkBlue}{RGB}{20,70,200}
\definecolor{peach}{rgb}{1, 0.56, 0.56}
\definecolor{NotionGreen}{RGB}{15,123,108}
\definecolor{NotionOrange}{RGB}{217,115,13}
\definecolor{NotionRed}{RGB}{224,62,62}
\definecolor{MontrealBlue}{RGB}{0, 30, 98}

\def\shownotes{1}  %
\ifnum\shownotes=1
\newcommand{\authnote}[2]{{$\ll$\textsf{\footnotesize #1: #2}$\gg$}}
\else
\newcommand{\authnote}[2]{}
\fi

%% file: sections/abstract.tex
\begin{abstract}

Solving inverse problems without any training involves using a pretrained generative model and making appropriate modifications to the generation process to avoid finetuning of the generative model. 
While recent methods have explored the use of diffusion models, they still require the manual tuning of many hyperparameters for different inverse problems. 
In this work, we propose a training-free method for solving linear inverse problems by using pretrained flow models, leveraging the simplicity and efficiency of Flow Matching models, using theoretically-justified weighting schemes, and thereby significantly reducing the amount of manual tuning.
In particular, we draw inspiration from two main sources: adopting prior gradient correction methods to the flow regime, and a solver scheme based on conditional Optimal Transport paths.
As pretrained diffusion models are widely accessible, we also show how to practically adapt diffusion models for our method.
Empirically, our approach requires no problem-specific tuning across an extensive suite of noisy linear inverse problems on high-dimensional datasets, ImageNet-64/128 and AFHQ-256, and we observe that our flow-based method for solving inverse problems improves upon closely-related diffusion-based methods in most settings.

\end{abstract}

%% file: sections/introduction.tex
\section{Introduction}

\begin{figure}[!b]
    \begin{subfigure}[b]{0.49\textwidth}
    \centering
    \includegraphics[width=0.49\textwidth]{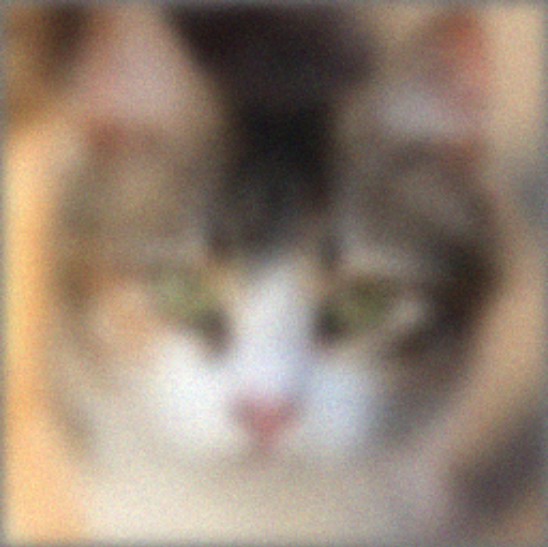}
        \includegraphics[width=0.49\textwidth]{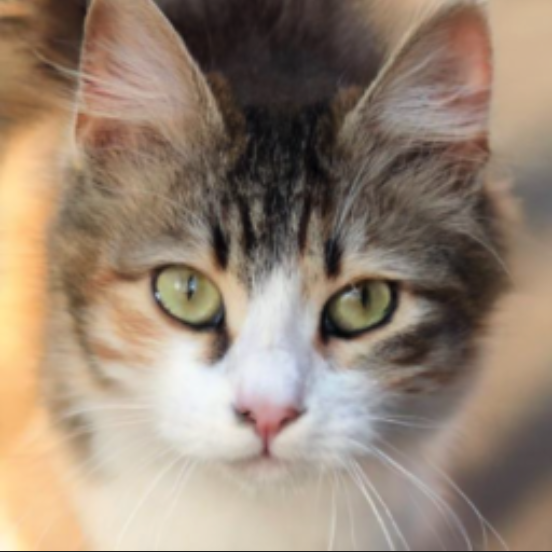}
        \caption{Gaussian deblur}
    \end{subfigure}
    \hfill
    \begin{subfigure}[b]{0.49\textwidth}
        \centering
        \includegraphics[width=0.49\textwidth]{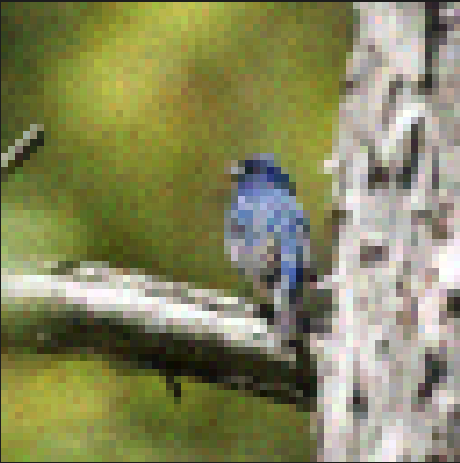}
        \includegraphics[width=0.49\textwidth]{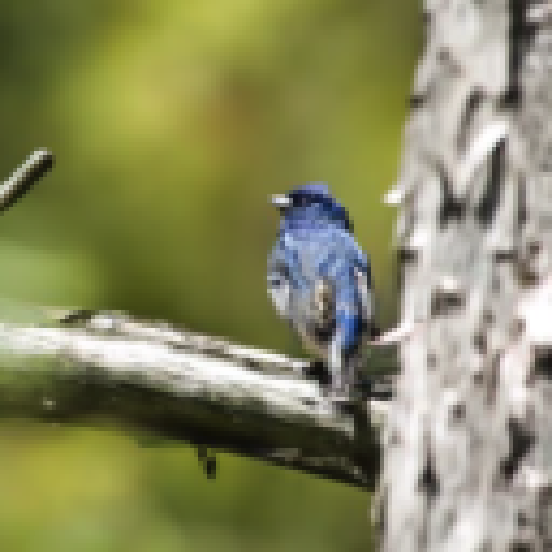}
        \caption{$2\times$ Super-resolution }
    \end{subfigure}\\
    \begin{subfigure}[b]{0.49\textwidth}
        \centering
        \includegraphics[width=0.49\textwidth]{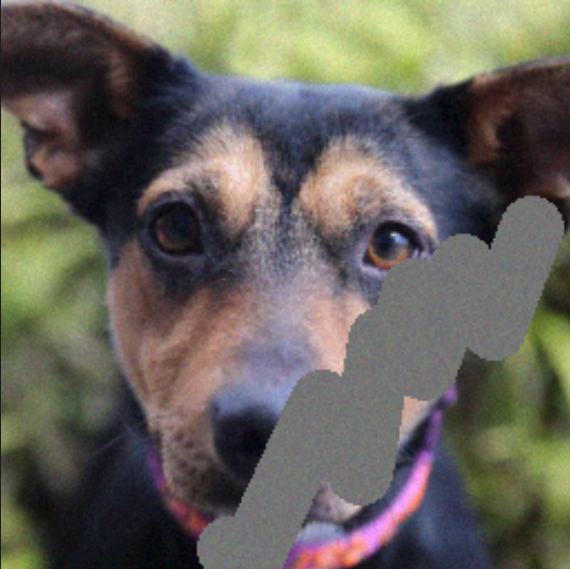}
        \includegraphics[width=0.49\textwidth]{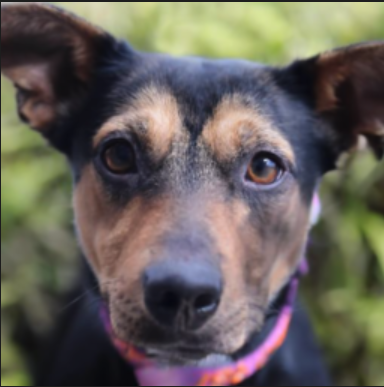}
        \caption{Inpainting}
    \end{subfigure}
    \hfill
    \begin{subfigure}[b]{0.49\textwidth}
        \centering
        \includegraphics[width=0.49\textwidth]{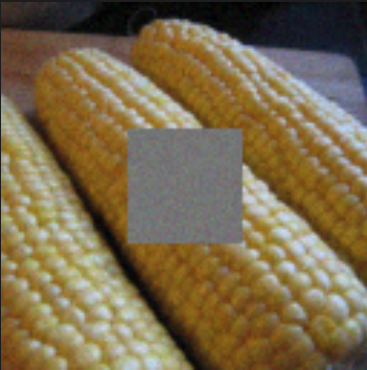}
        \includegraphics[width=0.49\textwidth]{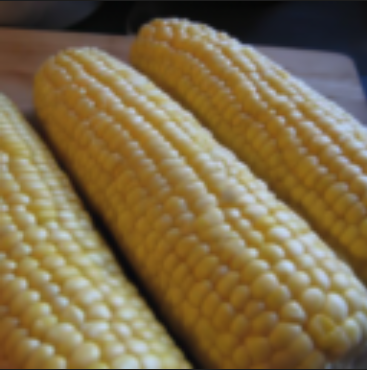}
        \caption{Inpainting}
    \end{subfigure}
    \caption{Corrected images by solving linear inverse problems with flow models. For each pair of images, we show the noisy measurement (left) and the reconstruction (right).}
    \label{fig:cifar-samples}
    \vspace{-0.2cm}
\end{figure}

Solving an inverse problem involves recovering a clean signal from noisy measurements generated by a known degradation model. 
Many interesting image processing tasks can be cast as an inverse problem. Some instances of these problems are super-resolution, inpainting, deblurring, colorization, denoising etc.
Diffusion models or score-based generative models~\citep{diffusion,ddpm, score,score-sde} have emerged as a leading family of generative models for solving inverse problems for images~\citep{sr3, palette, ddnm, dps, pigdm, reddiff}. 
However, sampling with diffusion models is known to be slow, and the quality of generated images is affected by the curvature of SDE/ODE solution trajectories~\citep{edm}.  While \cite{edm} observed ODE sampling for image generation could produce better results, sampling via SDE is still common for solving  inverse problems, whereas ODE sampling has been rarely considered, perhaps due to the use of diffusion probability paths.

Continuous Normalizing Flow (CNF)~\citep{neuralode} trained with Flow Matching~\citep{flowmatching} has been recently proposed as a powerful alternative to diffusion models. 
CNF (hereafter denoted flow model) has the ability to model arbitrary probability paths, and includes diffusion probability paths as a special case. 
Of particular interest to us are Gaussian probability paths that correspond to optimal transport (OT) displacement~\citep{ot-displacement}. 
Recent works~\citep{flowmatching,stochastic-interpolants,rectified-flow,shaul2023kinetic} have shown that these conditional OT probability paths are straighter than diffusion paths, which results in faster training and sampling with these models.
Due to these properties, conditional OT flow models are an appealing alternative to diffusion models for solving inverse problems.

In this work, we introduce a training-free method to utilize pretrained flow models for solving linear inverse problems.  
Our approach adds a correction term to the vector field that takes into account knowledge from the degradation model.
Specifically, we introduce an algorithm that incorporates $\Pi$GDM~\citep{pigdm} gradient correction to flow models.  Given the wide availability of pretrained diffusion models, we also present a way to convert these models to arbitrary paths for our sampling procedure. Empirically, we observe images restored via a conditional OT path exhibit perceptual quality better than that achieved by the model's original diffusion path, as well as recently proposed diffusion %
approaches such as $\Pi$GDM~\citep{pigdm} and RED-Diff~\citep{reddiff}, particularly in noisy settings. To summarize, our key contributions are:
\begin{itemize}
\item We present a training-free approach for solving linear inverse problems with pretrained conditional OT flow models that adapts the $\Pi$GDM correction, proposed for diffusion models, to flow sampling.
\item We offer a way to convert between flow models and diffusion models, enabling the use of pretrained continuous-time diffusion models for conditional OT sampling, and vice versa. 
\item We demonstrate that images restored via our algorithm using conditional OT probability paths have perceptual quality that is largely on par with, or better than that achieved by diffusion probability paths, and other recent methods like $\Pi$GDM and RED-Diff, without the need for problem-specific hyperparameter tuning.
\end{itemize}
\vspace{-5pt}

%% file: sections/background.tex
\section{Preliminaries}
\everypar{\looseness=-1}
We introduce relevant background knowledge and notation from conditional diffusion and flow modeling, as well as training-free inference with diffusion models.

\paragraph{Notation.}
Both diffusion and flow models consider two distinct processes indexed by time between $[0,1]$ that convert data to noise and noise to data. 
Here, we follow the temporal notation used in prior work~\citep{flowmatching} where the distribution at $t=1$ is the data distribution and $t=0$ is a standard Gaussian distribution. 
Note that this is opposite of the prevalent notation used in diffusion model literature~\citep{score,ddpm,ddim,score-sde}. 
We let $\vx_t$ denote the value %
at time $t$, without regard to which process (\ie diffusion or flow) it was drawn from.  
The probability density for the data to noise process is denoted $q$ and the parameterized probability density for the noise to data process is denoted $p_{\theta}$. Expectations with respect to $q$ are denoted via $\E_q$. We generally keep function arguments of $t$ implicit (i.e. $f(\vx_t, t)$ is informally written as $f(\vx_t)$.)

\paragraph{Conditional diffusion models.}
Conditional diffusion uses a class of processes $q$, and defines data to noise process $p_{\theta}$ as a Markov process that in continuous time obeys a stochastic differential equation (SDE) \citep{score, ddpm, score-sde}.
The parameters of $p_{\theta}$ are learned via minimizing a regression loss with respect to $\widehat{\vx_1}$
\begin{align}
L_{\text{diffusion}}(\widehat{\vx_1}, q) = \int_{0}^{1} w(t) \E_q[(\widehat{\vx_1}(\vx_t, \vy) - \vx_1)^2]dt
\label{eq:diffusion_reg_loss}
\end{align}
where $w(t)$ are positive weights~\citep{kingma2021variational, song2021maximum,kingma2023vdm}, %
$\vy$ is conditioning (i.e. a noisy image), and $\vx_1$ is noiseless data.  The optimal solution for $\widehat{\vx_1}$ is $\E_q[\vx_1 | \vx_t, \vy]$, and hence we refer to $\widehat{\vx_1}$ as a denoiser. 
 Many equivalent parameterizations exist and have known conversions to denoising.  Sampling using $p_{\theta}$ proceeds via starting from $\vx_0 \sim p_{\theta}(\vx_0 | \vy)$ and integrating the SDE to $t=1$.  If $p_{\theta}(\vx_0 | \vy) = q(\vx_0 | \vy)$, the SDE is integrated exactly, and $\widehat{\vx_1} = \E_q[\vx_1 | \vx_t, \vy]$, the resulting $\vx_1 \sim q(\vx_1 | \vy)$.

\paragraph{Conditional flow models} %
Alternatively, continuous normalizing flow models~\citep{neuralode} define the data generation process through an ODE. This leads to simpler formulations and does not introduce extra noise during intermediate steps of sample generation. Recently, simulation-free training algorithms have been designed specifically for such models~\citep{flowmatching,rectified-flow,stochastic-interpolants}, an example being the Conditional Flow Matching loss~\citep{flowmatching},
\begin{align}
L_{\text{cfm}}(\widehat{\vv}, q) = \int_{0}^{1} \E_q\left[ \left(\widehat{\vv}(\vx_t, \vy) - \E_q\left[\frac{d\vx_t}{dt} \bigg| \vx_t, \vy, \vx_1\right]\right)^2 \right] dt.
\end{align}
where $\widehat{\vv}(\vx_t, \vy)$ denotes a parameterized vector field defining the ODE
\begin{align}
\frac{d\vx_t}{dt} = \widehat{\vv}(\vx_t, \vy).
\end{align}
If trained perfectly, the marginal distributions of $\vx_t$, denoted $p_{\theta}(\vx_t | \vy)$, will match the marginal distributions of $q(\vx_t | \vy)$. Hence sampling from $q(\vx_t | \vy)$ involves starting from an initial value $\vx_{t'} \sim q(\vx_{t'} | \vy)$ and integrating the ODE from $t'$ to $t$. Typically, one samples from $t' = 0$ since $q(\vx_0 | \vy)$ is a tractable distribution. For general Gaussian probability paths $q(\vx_t | \vx_1 , \vy) = \mathcal{N}(\mu_t(\vy, \vx_1), \sigma_t(\vy, \vx_1)^2 \mI)$, one can set~\citep{flowmatching}
\begin{align}
\E_q\left[\frac{d\vx_t}{dt} \bigg| \vx_t, \vy, \vx_1\right] = \frac{d\mu_t}{dt} + \frac{d\sigma_t}{dt}\left(\frac{\vx_t - \mu_t}{\sigma_t}\right).
\label{eq:opt_vector_field}
\end{align}

\paragraph{Gaussian probability paths.}
The time-dependent distributions $q(\vx_t | \vy, \vx_1)$ are referred to as conditional probability paths.  We focus on the class of affine Gaussian probability paths of the form
\begin{align}
q(\vx_t | \vy, \vx_1) = q(\vx_t | \vx_1) = \mathcal{N}(\alpha_t \vx_1, \sigma_t^2 \mI)
\label{eq:gaussian_prob_paths}
\end{align}
where non-negative $\alpha_t$ and $\sigma_t$ are monotonically increasing and decreasing respectively.  This class includes the probability paths for conditional diffusion as well as the conditional Optimal Transport (OT) path~\citep{flowmatching}, where $\alpha_t = t$ and $\sigma_t = 1-t$.  The conditional OT path used by flow models has been demonstrated to have good empirical properties, including faster inference and better sampling in practice, and has theoretical support in high-dimensions~\citep{shaul2023kinetic}. As emphasized in \cite{lin2023common}, a desirable property for probability paths, obeyed by conditional OT but not commonly used diffusion paths, is to ensure $q(\vx_0 | \vy)$ is known (i.e. $\mathcal{N}(0, \mI)$), as otherwise one cannot exactly sample $\vx_0$ which can add substantial error. 

\paragraph{Training-free conditional inference using unconditional diffusion.}
Given pretrained \textit{unconditional} diffusion models that are trained to approximate $\E_q[\vx_1 | \vx_t]$, training-free approaches aim to approximate $\E_q[\vx_1 | \vx_t, \vy]$. Under Gaussian probability paths, the two terms are related as Tweedie's identity~\citep{tweedies} expresses $\E_q[\vx_1 | \vx_t, \vy] = (
\vx_t + \sigma_t^2 \nabla_{
\vx_t} \ln q(\vx_t | \vy)) / \alpha_t$.  Applying this identity (twice for both $\E_q[\vx_1 | \vx_t, \vy]$ and $\E_q[\vx_1 | \vx_t]$) and simplifying gives
\begin{align}
\E_q[\vx_1 | \vx_t, \vy] = \E_q[\vx_1 | \vx_t] + \frac{\sigma_t^2}{\alpha_t} \nabla_{\vx_t} \ln q(\vy | \vx_t).
\label{eq:tweedie}
\end{align}       
Following Eq.~\ref{eq:tweedie}, past approaches (\eg \cite{dps, pigdm}) have used the pretrained model for the first term and approximated the second intractable term to produce an approximate $\widehat{\vx_1}(\vx_t, \vy)$.  %
Diffusion posterior sampling (DPS)~\citep{dps} proposed to approximate $q(\vy | \vx_t)$ via $q(\vy | \vx_1 = \widehat{\vx_1}(\vx_t))$. 
Later, Pseudo-inverse Guided Diffusion Models ($\Pi$GDM)~\citep{pigdm} improved upon DPS for linear noisy observations where $\vy = \mA\vx + \sigma_y\epsilon$, where $\mA$ is some measurement matrix and $\epsilon \sim \mathcal{N}(0, \mI)$, by approximating $q(\vy | \vx_t)$ as $\mathcal{N}(\mA \widehat{\vx_1}(\vx_t), \sigma_\vy^2 \mI + r_t^2 \mA \mA^{T})$, derived via first approximating $q(\vx_1 | \vx_t)$ as $\mathcal{N}(\widehat{\vx_1}(\vx_t), r_t^2 \mI)$.  $\Pi$GDM also suggested adaptive weighting, replacing $\sigma_t^2 / \alpha_t$ with another function of time to account for the approximation. %

%% file: sections/methods.tex
\section{Solving Linear Inverse Problems without Training via Flows}

In the standard setup of a linear inverse problem, we observe measurements $\vy \in \mathbb{R}^n$ such that 
\begin{align}
    \vy = \mA\vx_1 + \epsilon
    \label{eqn:linear-inversion}
\end{align}
where $\vx_1 \in \mathbb{R}^m$ is drawn from an unknown data distribution $q(\vx_1)$, $\mA \in \mathbb{R}^{n \times m}$ is a known measurement matrix, and $\epsilon \sim \mathcal{N}(0, \sigma_y^2\mI)$ is unknown \iid Gaussian noise with known standard deviation $\sigma_y$.  Given a pretrained flow model with $\widehat{\vv}(\vx_t)$ that can sample from $q(\vx_1)$, and measurements $\vy$, our goal is to produce clean samples from the posterior $q(\vx_1 |\vy) \propto q(\vy | \vx_1) q(\vx_1)$ without training a problem-specific conditional flow model defined by $\widehat{\vv}(\vx_t, \vy)$.  Proofs are in Appendix~\ref{sec:proofs}.

\subsection{Converting between diffusion models and flow models}
\label{subsec:conversion}

While our proposed algorithm is based on flow models, this poses no limitation on needing pretrained flow models. We first
take a brief detour outside of inverse problems to demonstrate continuous-time diffusion models parameterized by %
$\widehat{\vx_1}$ can be converted to flow models $\widehat{\vv}$
that have Gaussian probability paths described by Eq.~\ref{eq:gaussian_prob_paths}. 
We separate training and sampling such that one can train with Gaussian probability path 
$q'$ 
and then perform sampling with a different Gaussian probability path 
$q$.
Equivalent expressions to this subsection have been derived in~\cite{edm}, leveraging a more general conversion from SDE to probability flow ODE from~\cite{score-sde}. \cite{edm} similarly separate training and sampling for Gaussian probability paths, and demonstrate that using alternative sampling paths was beneficial.  Our derivations avoid the SDE via a flow perspective and exposes a subtlety when swapping probability paths using pretrained models.

\begin{theoremEnd}[category=conversion]{lemma}%
\label{lem:cfm_conversion}
For Gaussian probability path $q$ given by Eq.~\ref{eq:gaussian_prob_paths}, the optimal solution for $\widehat{\vv}(\vx_t, \vy)$ is known given $\E_q[\vx_1 | \vx_t, \vy]$, and vice versa.
\end{theoremEnd}

\begin{proofEnd}
\quad Inserting Gaussian probability paths defined by Eq.~\ref{eq:gaussian_prob_paths} into Eq.~\ref{eq:opt_vector_field} gives
\begin{align}
\E_q\left[\frac{d\vx_t}{dt} \bigg| \vx_t, \vy, \vx_1\right] = \frac{d\alpha_t}{dt} \vx_1 + \frac{d\sigma_t}{dt}\left(\frac{\vx_t - \alpha_t \vx_1}{\sigma_t}\right).
\label{eq:optimal_vf}
\end{align}
So we reparameterize vector field $\widehat{\vv}(\vx_t, \vy) = \frac{d\alpha_t}{dt} \widehat{\vx_1}(\vx_t, \vy) + \frac{d\sigma_t}{dt}\left(\frac{\vx_t - \alpha_t \widehat{\vx_1}(\vx_t,\vy)}{\sigma_t}\right)$.  Inserting these expressions into the Conditional Flow Matching loss gives
\begin{align}
\int_{0}^{1} \left(\alpha_t \frac{d \ln(\alpha_t / \sigma_t)}{dt}\right)^2 \E_q[(\widehat{\vx_1} - \vx_1)^2]dt,
\end{align}
recovering the denoising loss with particular $w(t)$.  The optimal solution does not depend on the weights and is $\E_q[\vx_1 | \vx_t, \vy]$.  
\end{proofEnd}

In particular, a diffusion model's denoiser $\widehat{\vx_1}(\vx_t, \vy)$ trained using Gaussian probability path $q$ can be interchanged with a flow model's $\widehat{\vv}(\vx_t, \vy)$ with the same $q$ via
\begin{align}
\widehat{\vv} = \left(\alpha_t \frac{d \ln (\alpha_t / \sigma_t)}{dt}\right)\widehat{\vx_1} + \frac{d \ln \sigma_t}{dt}\vx_t.
\label{eq:interchange}
\end{align}
Furthermore, $\widehat{\vx_1}(\vx_t, \vy)$ trained under $q'$ can be used for another Gaussian $q$ during sampling.

\begin{theoremEnd}[category=conversion]{lemma}%
Consider two Gaussian probability paths $q$ and $q'$ defined by Eq.~\ref{eq:gaussian_prob_paths} with $\alpha_t$, $\sigma_t$ and $\alpha'_t$, $\sigma'_t$ respectively.  Define $t'(t)$ as the unique solution to $\frac{\sigma_t}{\alpha_t} = \frac{\sigma'_{t'}}{\alpha'_{t'}}$ when it exists for given $t$.  Then
\begin{align}
\E_q[\vx_1 | \vx_t, \vy] = \E_{q'}[\vx_1 | \mX'_{t'(t)} = \alpha'_{t'(t)} \vx_t / \alpha_t, \vy].
\end{align}
\end{theoremEnd}

\begin{proofEnd}
\quad We first note that $q$ and $q'$ share $q(\vx_1 | \vy) = q'(\vx_1 | \vy)$.  Then algebraically for Gaussian distributions $q$ and $q'$, $q(\mX_t = \vx_t | \vx_1, \vy) = q'(\mX'_{t'(t)} = \alpha'_{t'(t)} \vx_t / \alpha_t | \vx_1, \vy)$ when $t'(t)$ exists.  The solution for $t'(t)$ is unique due to the monotonicity requirements of both $\alpha$ and $\sigma$.  Since the joint densities are therefore identical, $\E_q[\vx_1 | \vx_t, \vy] = \E_{q'}[\vx_1 | \mX'_{t'(t)} = \alpha'_{t'(t)} \vx_t / \alpha_t,, \vy]$.
\end{proofEnd}

So if $\widehat{\vx_1}$ was trained under $q'$ it can be used for sampling under $q$ via evaluating at $\widehat{\vx_1}(\alpha'_{t'(t)} \vx_t / \alpha_t, t'(t), \vy)$ (with explicit time for clarity) whenever $t'(t)$ exists.  %
In particular, if pre-trained denoiser was trained with $q'$ and we perform conditional OT sampling, we utilize
\begin{align}
t'(t) = \text{SNR}_{q'}^{-1}(\text{SNR}_{q}(t))) = \text{SNR}_{q'}^{-1}\left(\frac{t}{1-t}\right).
\end{align}  
where signal-to-noise ratio $\text{SNR}(t) = \alpha_t / \sigma_t$.  %
The main avenue for non-existence for $t'(t)$ is if the model under $q'$ is trained using a minimum SNR above zero, which induces a minimum $t$ for which $t'(t)$ exists.  When a minimum $t$ exists, we can only perform sampling with $q$ starting from $\vx_t \sim q(\vx_t | \vy)$. Approximating this sample is entirely analogous to approximating $\vx_0 \sim q'(\vx_0 | \vy)$.  This error already exists for $q'$ because $q'(\vx_0 | \vy)$ is not $\mathcal{N}(0, \mI)$ unless $q'$ is trained to zero SNR. An initialization problem cannot be avoided if $q'$ has limited SNR range by switching paths to $q$. 

\subsection{Correcting the vector field of flow models}

To solve linear inverse problems without any training via  flow models, we derive an expression similar to Eq.~\ref{eq:tweedie} that relates conditional vector fields under Gaussian probability paths to unconditional vector fields.

\begin{theoremEnd}[category=conversion]{theorem}%
\label{thm:vector_field_corr}
Let $q$ be a Gaussian probability path described by Eq.~\ref{eq:gaussian_prob_paths}. 
Assume we observe $\vy \sim q(\vy | \vx_1)$ for arbitrary $q(\vy | \vx_1)$ and $v(\vx_t)$ is a vector field enabling sampling $\vx_t \sim q(\vx_t)$.  Then a vector field $v(\vx_t, \vy)$ enabling sampling $\vx_t \sim q(\vx_t | \vy)$ can be written
\begin{align}
v(\vx_t, \vy) = v(\vx_t) + \sigma_t^2 \frac{d \ln(\alpha_t / \sigma_t)}{dt} \nabla_{\vx_t} \ln q(\vy | \vx_t).
\label{eq:vec_corr}
\end{align}
\end{theoremEnd}

\begin{proofEnd}
\quad The optimal $v(\vx_t, \vy)$ for the Conditional Flow Matching loss is $\E_q[\frac{d\vx_t}{dt} | \vx_t, \vy]$ and for Gaussian probability paths described by Eq.~\ref{eq:gaussian_prob_paths} can be written using Eq.~\ref{eq:optimal_vf} as
\begin{align}
v(\vx_t, \vy) = \frac{d\alpha_t}{dt} \E_q[\vx_1 | \vx_t, \vy] + \frac{d\sigma_t}{dt}\left(\frac{\vx_t - \alpha_t \E_q[\vx_1 | \vx_t, \vy]}{\sigma_t}\right),
\end{align}
where the identical expression without $\vy$ holds for $v(\vx_t)$.  Inserting the result from Eq.~\ref{eq:tweedie} and simplifying gives Eq.~\ref{eq:vec_corr}.
\end{proofEnd}

We turn Theorem~\ref{thm:vector_field_corr} into an training-free algorithm for solving linear inverse using flows by adapting ${\Pi}$GDM's approximation.  In particular, given $\widehat{\vv}(\vx_t)$ (or $\widehat{\vx_1}(\vx_t)$), our approximation will be 
\begin{align}
\widehat{\vv}(\vx_t, \vy) = \widehat{\vv}(\vx_t) + \sigma_t^2 \frac{d \ln(\alpha_t / \sigma_t)}{dt}\gamma_t \nabla_{\vx_t} \ln q^{app}(\vy | \vx_t),
\label{eq:vec_alg_corr}
\end{align}
where following ${\Pi}$GDM terminology, we refer to $\gamma_t = 1$ as unadaptive and other choices as adaptive weights.  In general, we view adaptive weights $\gamma_t \neq 1$ as an adjustment for error in $q^{app}(\vy | \vx_t)$.

For $q^{app}(\vy | \vx_t)$, we generalize ${\Pi}$GDM to any Gaussian probability path described by Eq.~\ref{eq:gaussian_prob_paths} via updating $r_t^2$.  We still have $q^{app}(\vy | \vx_t)$ is $\mathcal{N}(\mA\widehat{\vx_1}(\vx_t), \sigma^2_y I + r^2_t \mA \mA^\top)$ when $q(\vx_1 | \vx_t)$ is approximated as $\mathcal{N}(\widehat{\vx_1}(\vx_t), r_t^2 I)$.  We choose $r_t^2$ by following ${\Pi}$GDM's derivation, noting that if $q(\vx_1)$ is $\mathcal{N}(0, \mI)$ and $q(\vx_t | \vx_1)$ is $\mathcal{N}(\alpha_t \vx_1, \sigma_t^2 I)$, then
\begin{align}
r_t^2 = \frac{\sigma_t^2}{\sigma_t^2 + \alpha_t^2}. \label{eqn:r_t_sqr_flow_value}
\end{align}
When $\alpha_t = 1$, we recover ${\Pi}$GDM's $r_t^2$ as expected under their Variance-Exploding path specification.%

\paragraph{Starting flow sampling at time $t$ > 0.} Initializing conditional diffusion model sampling at $t > 0$ has been proposed by~\cite{chung2022come}.  For flows, we similarly want $\vx_t \sim q(\vx_t | \vy)$ at initialization time $t$.  In our experiments, we examine (approximately) initializing at different times $t > 0$ using
\begin{align}
\vx_t = \alpha_t \vy + \sigma_t \epsilon
\label{eq:initialization}
\end{align}
for $\epsilon \sim \mathcal{N}(0, \mI)$ when $\vy$ is the correct shape.  For super-resolution, we use nearest-neighbor interpolation on $\vy$ instead. We also consider using $\mA^{\dagger}y$ as an ablation in the~\cref{sec:appendix-ablation} (where $\mA^{\dagger}$ is the pseudo-inverse of $\mA$~\citep{pigdm}). We may be forced to use this initialization for flow sampling due to converting a diffusion model not trained to zero SNR.  However as shown in~\citep{chung2022come} for diffusion, this initialization can improve results more generally.  Conceptually, if the resulting $\vx_t$ is closer to $\vx_t \sim q(\vx_t | \vy)$ than achieved via starting from an earlier time $t'$ and integrating, then this initialization can result in less overall error.

\paragraph{Algorithm summary.}
Putting this altogether, our proposed approach using flow sampling and conditional OT probability paths is succinctly summarized in ~\cref{alg:cond-ot-flow-image-inversion}, derived via inserting $\alpha_t = t$ and $\sigma_t = 1-t$. Unlike $\Pi$GDM, we propose unadaptive weights $\gamma_t = 1$.  By default, we set initialization time $t=0.2$.  The algorithm therefore has no additional hyperparameters to tune over traditional diffusion or flow sampling.  In Appendix~\ref{sec:alg_for_arbitrary_path}, we detail our algorithm for other Gaussian probability paths, and the equivalent formulation when a pretrained vector field is available instead.     
\vspace{-5pt}

\begin{algorithm}
\caption{Solving linear inverse problems via flows using conditional OT probability path} %
\label{alg:cond-ot-flow-image-inversion}
\begin{algorithmic}[1]
\Require{Pretrained denoiser $\widehat{\vx_1}(\vx_t)$ converted to conditional OT probability path using Section~\ref{subsec:conversion}, noisy measurement $\vy$, measurement matrix $\mA$, initial time $t$, and std $\sigma_y$}
\State{Initialize $\vx_t = t \rvy + (1 - t)\epsilon$, where $\epsilon \sim \mathcal{N}(0, \mI)$}
\Comment{Initialize $\vx_t$, \cref{eq:initialization}}
\State{$\vz_t = \vx_t$}
\For{each time step $t'$ of ODE integration} \Comment{Integrate ODE from $t' = t$ to $1$.}
    \State{$r_{t'}^2 = \frac{(1 - {t'})^2}{{t'}^2 + (1 - {t'})^2}$} \Comment{Value of $r_t^2$ from \cref{eqn:r_t_sqr_flow_value}}
    \State{$\widehat{\vv} = \frac{\widehat{\vx_1}(\vz_{t'}) - \vz_{t'}}{1-t'}$} \Comment{Convert $\widehat{\vx_1}$ to vector field, \cref{eq:interchange}}
    \State{$\vg = (\vy - \mA \widehat{\vx_1})^\top (r_{t'}^2\mA \mA^\top + \sigma_y^2\mI)^{-1} \mA \frac{\partial \widehat{\vx_1}}{\partial \vz_{t'}}$} \Comment{$\Pi$GDM correction}
    \State{$\widehat{\vv}_{\text{corrected}} = \widehat{\vv} + \frac{1-{t'}}{{t'}}\vg$} \Comment{Correct unconditional vector field $\widehat{\vv}$, \cref{eq:vec_alg_corr}}
\EndFor
\end{algorithmic}
\end{algorithm}

%% file: sections/results.tex
\section{Experiments}
\paragraph{Datasets.} We verify the effectiveness of our proposed approach on three datasets: face-blurred ImageNet $64\times64$ and $128\times128$~\citep{imagenet-orig, imagenet, fb-imagenet}, and AnimalFacesHQ (AFHQ) $256\times256$~\citep{afhq}. %
We report our results on $10$K randomly sampled images from validation split of ImageNet, and $1500$ images from test split of AFHQ. %

\paragraph{Tasks.} We report results on the following linear inverse problems: inpainting (center-crop), Gaussian deblurring, super-resolution, and denoising. The exact details of the measurement operators are: 1) For inpainting, we use centered mask of size %
$20\times20$ for ImageNet-64, $40\times40$ for ImageNet-128, and $80\times80$ for AFHQ. %
In addition, for images of size $256 \times 256$, we also use free-form masks simulating brush strokes similar to the ones used in \cite{palette, pigdm}.
2) For super-resolution, we apply bicubic interpolation to downsample images by $4\times$ for datasets that have images with resolution $256\times256$ and downsample images by $2\times$ otherwise.
3) For Gaussian deblurring, we apply Gaussian blur kernel of size %
$61\times61$ with intensity value $1$ for ImageNet-$64$ and ImageNet-$128$, and $61\times61$ with  intensity value $3$ for AFHQ. %
4) For denoising, we add \iid Gaussian noise with $\sigma_y = 0.05$ to the images.
For tasks besides denoising, we consider \iid Gaussian noise with $\sigma_y = 0$ and $0.05$ to the images. 
Images $\vx_1$ are normalized to range $[-1,1]$.  

\paragraph{Implementation details.} We trained our own continuous-time conditional VP-SDE model, and conditional Optimal Transport (conditional OT) flow model from scratch on the above datasets following the hyperparameters and training procedure outlined in \citet{score-sde} and \citet{flowmatching}.  
These models are conditioned on class labels, not noisy images.  All derivations hold with class label $c$ since $q(\vy | c, \vx_1) = q(\vy | \vx_1)$ (i.e. the noisy image is independent of class label given the image).  We use the open-source implementation of the Euler method provided in torchdiffeq library~\citep{torchdiffeq} to solve the ODE in our experiments.  Our choice of Euler is intentionally simple, as we focus on flow sampling with the conditional OT path, and not on the choice of ODE solver.

\paragraph{Metrics.} We follow prior works~\citep{dps,ddrm} and report Fr\'echet Inception Distance (FID)~\citep{fid}, Learned Perceptual Image Patch Similarity (LPIPS)~\citep{lpips}, peak signal-to-noise ratio (PSNR), and structural similarity index (SSIM). We use open-source implementations of these metrics in the TorchMetrics library~\citep{torchmetrics}. 

\paragraph{Methods and baselines.} %
We use our two pretrained model checkpoints--- a conditional OT flow model and continuous VP-SDE diffusion model, and perform flow sampling with both conditional OT and Variance-Preserving (VP) paths, labeling our methods as OT-ODE and VP-ODE respectively.  Because qualitative results are identical and quantitative results similar, we only include the VP-SDE diffusion model in the main text, and include the conditional OT flow model in \cref{sec:empirical-results}.  We compare our OT-ODE and VP-ODE methods against $\Pi$GDM~\citep{pigdm} and RED-Diff~\citep{reddiff} as relevant baselines.  We selected these baselines because they achieve state-of-the-art performance in solving linear inverse problems using diffusion models. The code for both baseline methods is available on github, and we make minimal changes while reimplementing these methods in our codebase.  A fair comparison between methods requires considering the number of function evaluations (NFEs) used during sampling.  %
We utilize at most $100$ NFEs for our OT-ODE and VP-ODE sampling (see \cref{sec:empirical-results}), and utilize $100$ for $\Pi$GDM as recommended in \citet{pigdm}.  
We allow RED-Diff $1000$ NFEs since it does not require gradients of $\widehat{\vx_1}$.  
For OT-ODE following~\cref{alg:cond-ot-flow-image-inversion}, we use $\gamma_t = 1$ and initial $t=0.2$ for all datasets and tasks.  
For VP-ODE following~\cref{alg:arb-flow-image-inversion-denoiser} in the Appendix, we use $\gamma_t = \sqrt{\frac{\alpha_t}{\alpha_t^2 + \sigma_t^2}}$ and initial $t=0.4$ for all datasets and tasks.  
Ablations of these mildly tuned hyperparameters are shown in \cref{sec:appendix-ablation}. We extensively tuned hyperparameters for RED-Diff and $\Pi$GDM as described in \cref{sec:baselines}, including different hyperparameters per dataset and task.  %

\subsection{Experimental Results}
\begin{figure}[!t]
    \centering
    \begin{minipage}[b]{0.95\textwidth}
        \includegraphics[width=0.49\textwidth]{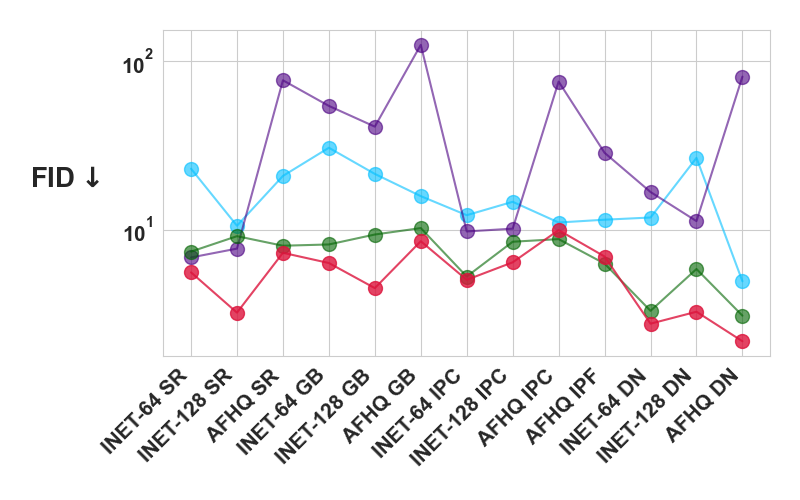}
        \includegraphics[width=0.49\textwidth]{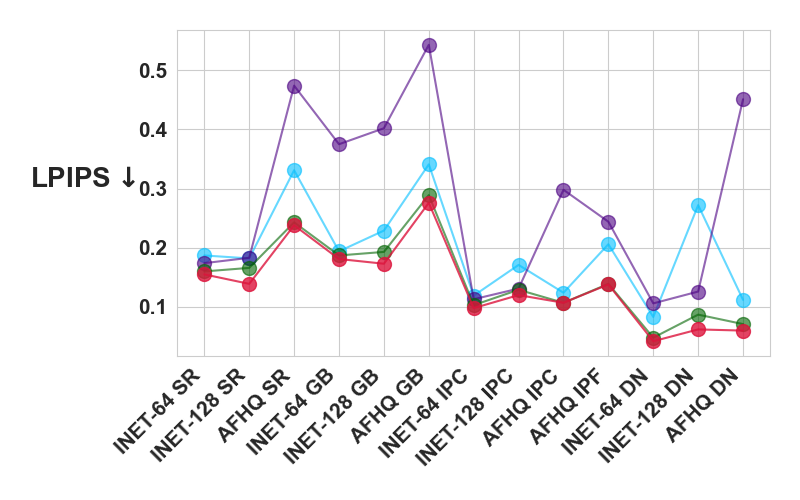}
        \includegraphics[width=0.49\textwidth]{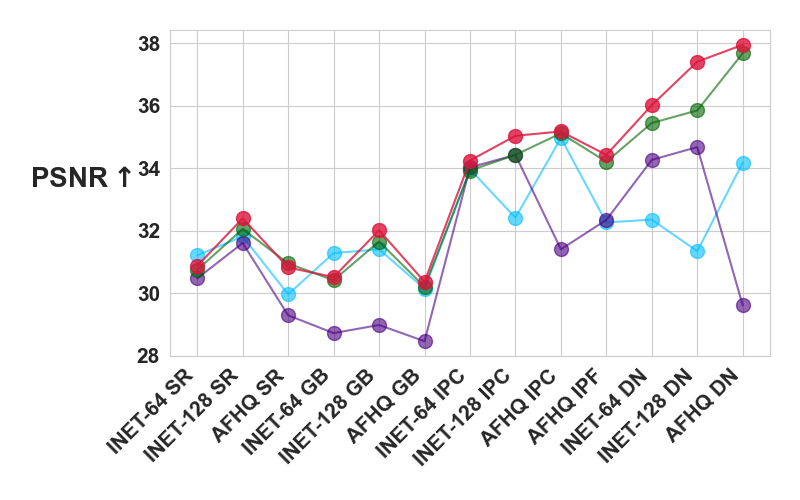}
        \includegraphics[width=0.49\textwidth]{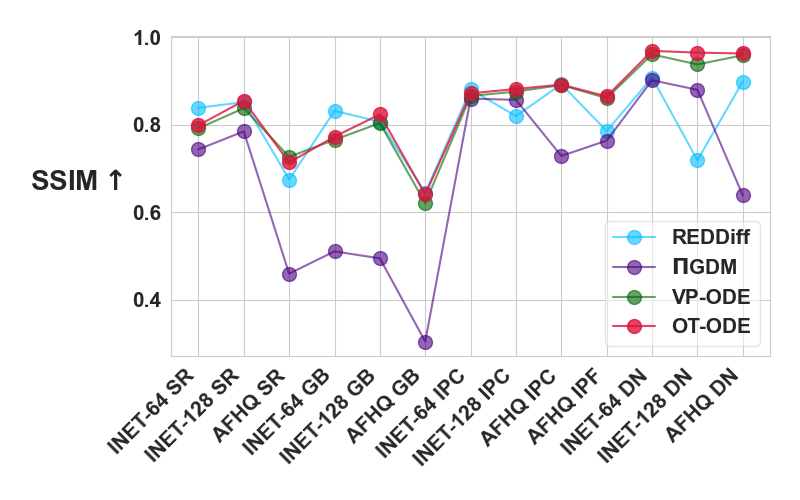}
    \end{minipage}
    \caption{Quantitative evaluation of pretrained VP-SDE model for solving linear inverse problems on super-resolution (SR), gaussian deblurring (GB), inpainting with centered (IPC) and freeform mask (IPF), and denoising (DN) with $\sigma_y=0.05$. We present results on face-blurred ImageNet-64 (INET-64), face-blurred ImageNet-128 (INET-128), and AFHQ.}
    \label{fig:sigma-0.05-ddpm-ckpt}
    \vspace{-10pt}
\end{figure}

We report quantitative results for the VP-SDE model, across all datasets and  linear measurements, in \cref{fig:sigma-0.05-ddpm-ckpt} for $\sigma_y=0.05$, and in \cref{fig:sigma-0-ddpm-ckpt} within~\cref{sec:empirical-results} for $\sigma_y=0$. Additionally, we report results for the conditional OT flow model in \cref{fig:sigma-0.05-cond-OT} and \cref{fig:sigma-0-ot-ckpt} for $\sigma_y=0.05$ and $\sigma_y=0$, respectively, in~\cref{sec:empirical-results}. Exact numerical values for all the metrics across all datasets and tasks can also be found in~\cref{sec:empirical-results}.  Qualitative images have been selected for demonstration purposes.

\begin{figure}[!ht]
    \centering
    \begin{subfigure}[t]{0.15\textwidth}
        \includegraphics[width=\textwidth]{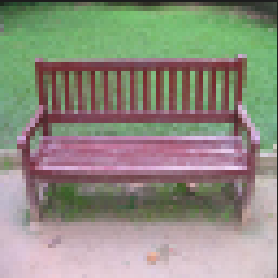}
    \end{subfigure}
    \begin{subfigure}[t]{0.15\textwidth}
        \includegraphics[width=\textwidth]{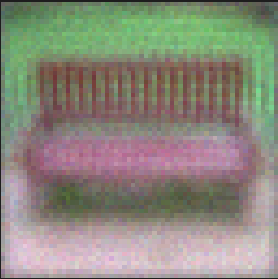}
    \end{subfigure}
    \begin{subfigure}[t]{0.15\textwidth}
        \includegraphics[width=\textwidth]{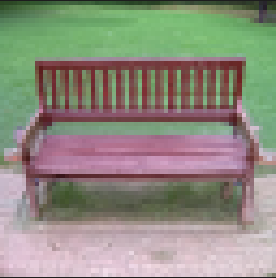}
    \end{subfigure}
    \begin{subfigure}[t]{0.15\textwidth}
    \includegraphics[width=\textwidth]{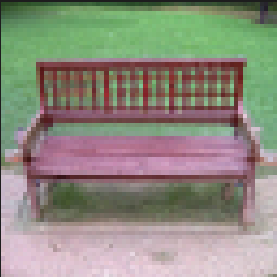}
    \end{subfigure}
    \begin{subfigure}[t]{0.15\textwidth}    
    \includegraphics[width=\textwidth]{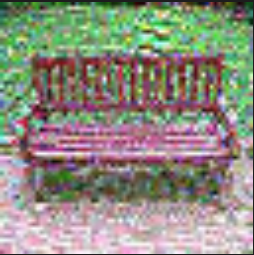}
    \end{subfigure} 
    \begin{subfigure}[t]{0.15\textwidth}    
        \includegraphics[width=\textwidth]{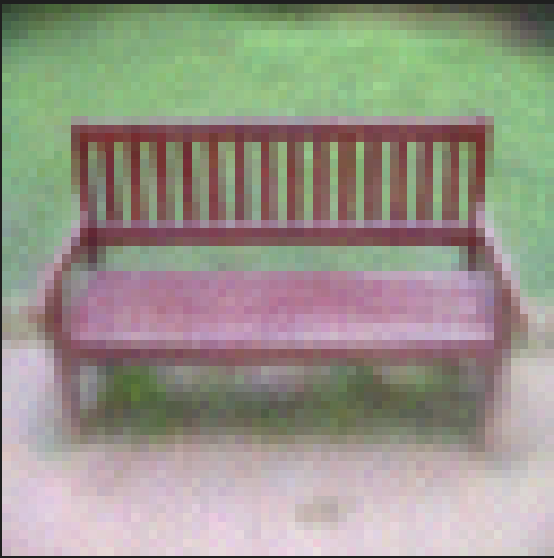}
    \end{subfigure}
    \\
        \begin{subfigure}[t]{0.15\textwidth}
        \includegraphics[width=\textwidth]{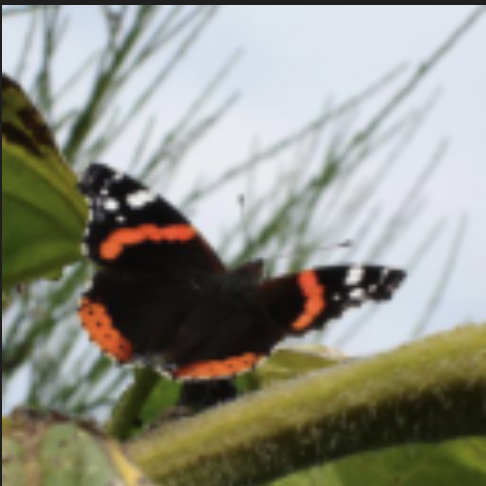}
    \end{subfigure}
    \begin{subfigure}[t]{0.15\textwidth}
        \includegraphics[width=\textwidth]{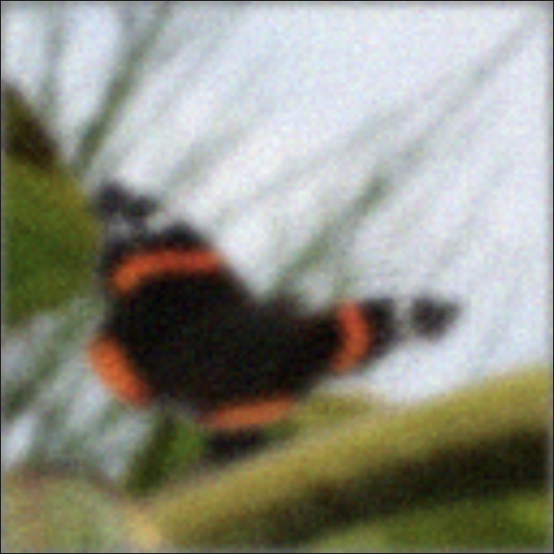}
    \end{subfigure}
    \begin{subfigure}[t]{0.15\textwidth}
        \includegraphics[width=\textwidth]{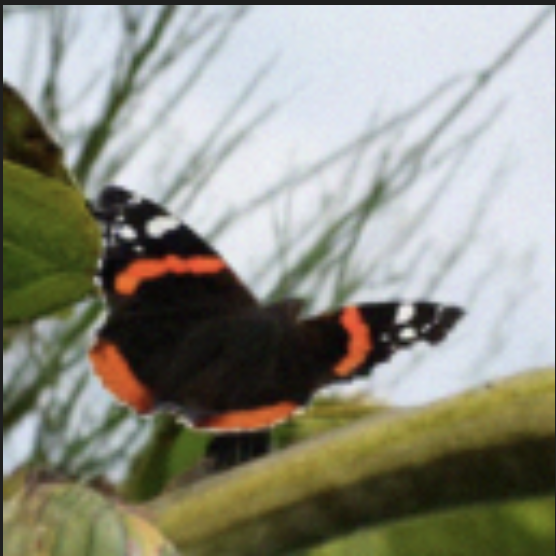}
    \end{subfigure}
    \begin{subfigure}[t]{0.15\textwidth}
        \includegraphics[width=\textwidth]{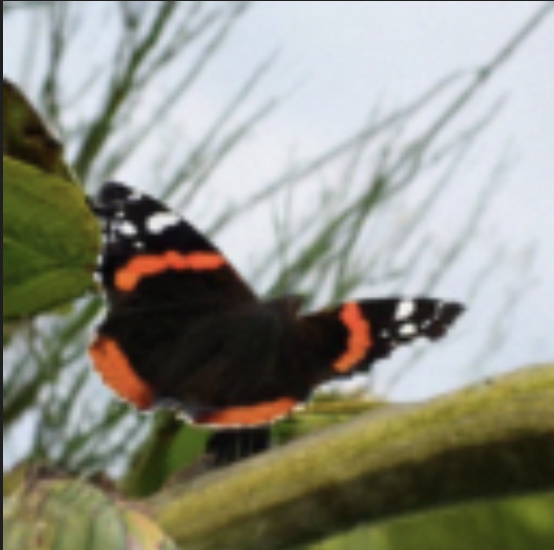}
    \end{subfigure}
    \begin{subfigure}[t]{0.15\textwidth}    
        \includegraphics[width=\textwidth]{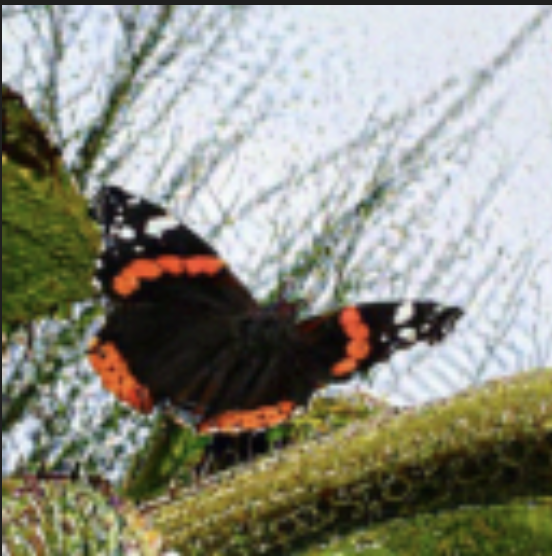}
    \end{subfigure}
    \begin{subfigure}[t]{0.15\textwidth}    
        \includegraphics[width=\textwidth]{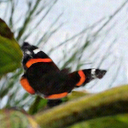}
    \end{subfigure}
    \\
    \begin{subfigure}[t]{0.15\textwidth}
        \includegraphics[width=\textwidth]{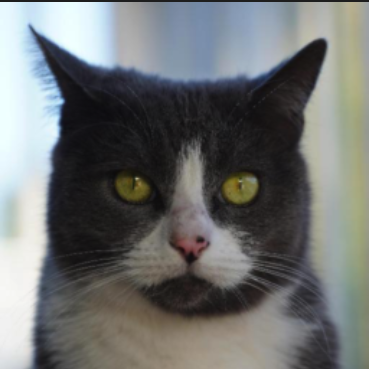}
        \caption{Reference}
    \end{subfigure}
    \begin{subfigure}[t]{0.15\textwidth}
        \includegraphics[width=\textwidth]{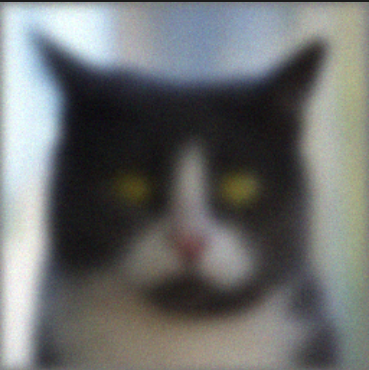}
        \caption{Distorted}
    \end{subfigure}
    \begin{subfigure}[t]{0.15\textwidth}
        \includegraphics[width=\textwidth]{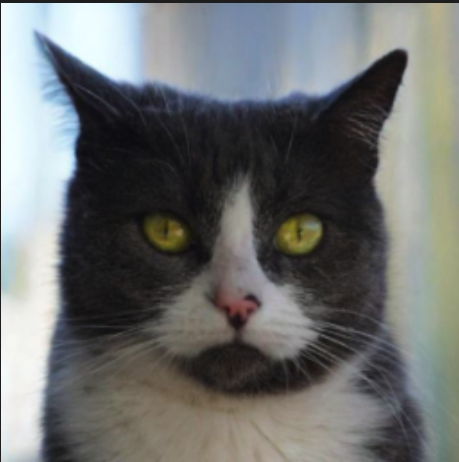}
        \caption{OT-ODE}
    \end{subfigure}
    \begin{subfigure}[t]{0.15\textwidth}
        \includegraphics[width=\textwidth]{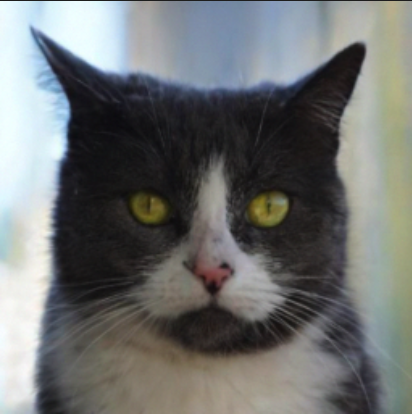}
        \caption{VP-ODE}
    \end{subfigure}
    \begin{subfigure}[t]{0.15\textwidth}    
        \includegraphics[width=\textwidth]{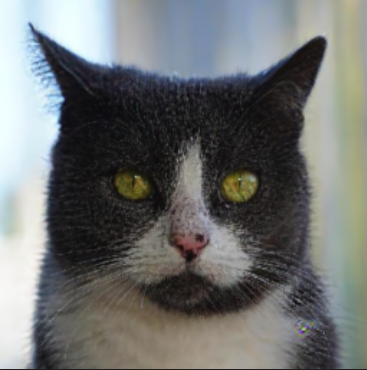}
        \caption{$\Pi$GDM}
    \end{subfigure}
    \begin{subfigure}[t]{0.15\textwidth}    
        \includegraphics[width=\textwidth]{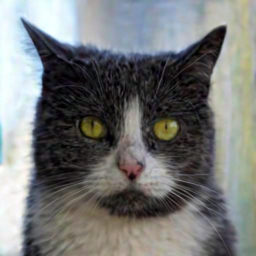}
        \caption{REDDiff}
    \end{subfigure}
    \caption{Results for Gaussian deblurring with VP-SDE model and $\sigma_y=0.05$ for (\textbf{first row}) ImageNet-64, (\textbf{second row}) ImageNet-128, and (\textbf{third row}) AFHQ.} 
    \label{fig:gb-noisy-ddpm-vis}
\end{figure}

\paragraph{Gaussian deblurring.} We report qualitative noisy results for the VP-SDE model in \cref{fig:gb-noisy-ddpm-vis} and for the conditional OT flow (cond-OT) model in \cref{fig:gb-noisy-ot-vis}.  We observe that OT-ODE and VP-ODE outperforms $\Pi$GDM and RED-Diff, both qualitatively and quantitatively, across all datasets for $\sigma_y=0.05$.  As shown in these figures, $\Pi$GDM tends to sharpen the images, which sometimes results in unnatural textures in the images. Further, we also observe some unnatural textures and background noise with RED-Diff for $\sigma_y=0.05$.  For $\sigma_y=0$, OT-ODE has better FID and LPIPS, but $\Pi$GDM shows improved PSNR and SSIM.  \cref{fig:gb-noiseless-ddpm-vis} and \cref{fig:gb-noiseless-ot-vis} show qualitative examples for $\sigma_y=0$.

\begin{figure}[!ht]
    \centering
    \begin{subfigure}[t]{0.15\textwidth}
        \includegraphics[width=\textwidth]{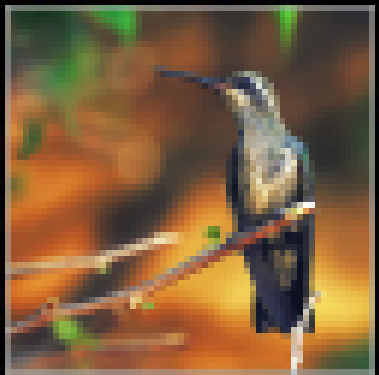}
    \end{subfigure}
    \begin{subfigure}[t]{0.15\textwidth}
        \includegraphics[width=\textwidth]{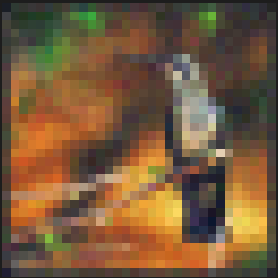}
    \end{subfigure}
    \begin{subfigure}[t]{0.15\textwidth}
        \includegraphics[width=\textwidth]{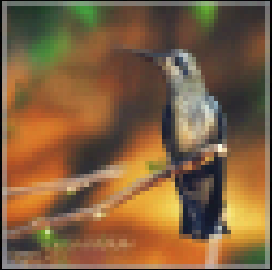}
    \end{subfigure}
    \begin{subfigure}[t]{0.15\textwidth}
        \includegraphics[width=\textwidth]{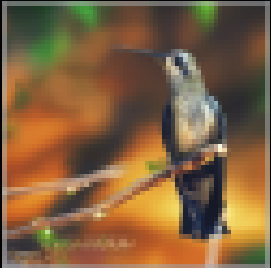}
    \end{subfigure}
    \begin{subfigure}[t]{0.15\textwidth}    
        \includegraphics[width=\textwidth]{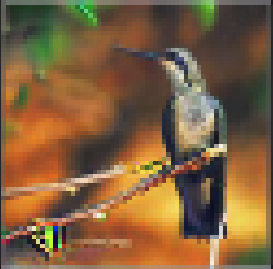}
    \end{subfigure} 
    \begin{subfigure}[t]{0.15\textwidth}    
        \includegraphics[width=\textwidth]{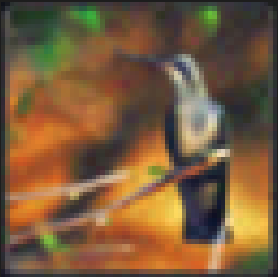}
    \end{subfigure}
    \\
    \begin{subfigure}[t]{0.15\textwidth}
        \includegraphics[width=\textwidth]{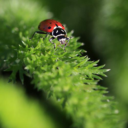}
    \end{subfigure}
    \begin{subfigure}[t]{0.15\textwidth}
        \includegraphics[width=\textwidth]{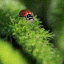}
    \end{subfigure}
    \begin{subfigure}[t]{0.15\textwidth}
        \includegraphics[width=\textwidth]{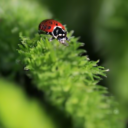}
    \end{subfigure}
    \begin{subfigure}[t]{0.15\textwidth}
        \includegraphics[width=\textwidth]{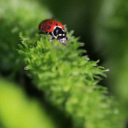}
    \end{subfigure}
    \begin{subfigure}[t]{0.15\textwidth}    
        \includegraphics[width=\textwidth]{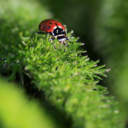}
    \end{subfigure}
    \begin{subfigure}[t]{0.15\textwidth}    
        \includegraphics[width=\textwidth]{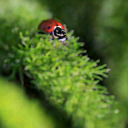}
    \end{subfigure}
    \\
    \begin{subfigure}[t]{0.15\textwidth}
        \includegraphics[width=\textwidth]{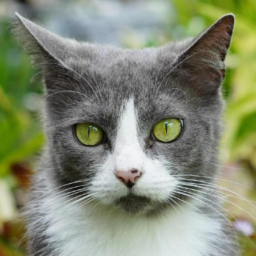}
        \caption{Reference}
    \end{subfigure}
    \begin{subfigure}[t]{0.15\textwidth}
        \includegraphics[width=\textwidth]{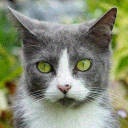}
        \caption{Distorted}
    \end{subfigure}
    \begin{subfigure}[t]{0.15\textwidth}
        \includegraphics[width=\textwidth]{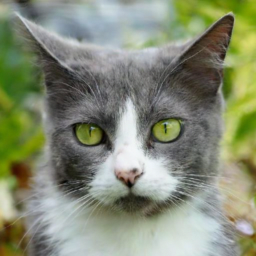}
        \caption{OT-ODE}
    \end{subfigure}
    \begin{subfigure}[t]{0.15\textwidth}
        \includegraphics[width=\textwidth]{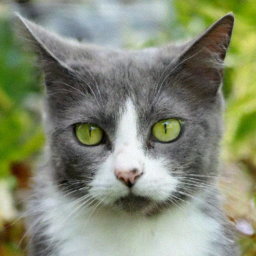}
        \caption{VP-ODE}
    \end{subfigure}
    \begin{subfigure}[t]{0.15\textwidth}    
        \includegraphics[width=\textwidth]{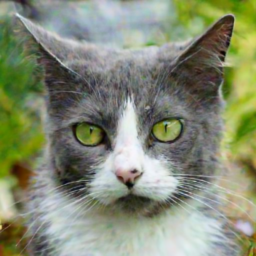}
        \caption{$\Pi$GDM}
    \end{subfigure}
    \begin{subfigure}[t]{0.15\textwidth}    
        \includegraphics[width=\textwidth]{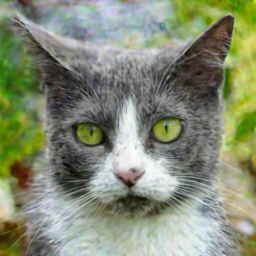}
        \caption{REDDiff}
    \end{subfigure}
    \caption{Results for super-resolution with VP-SDE model and $\sigma_y=0.05$ for (\textbf{first row}) ImageNet-64 $2\times$, (\textbf{second row}) ImageNet-128 $2\times$, and (\textbf{third row}) AFHQ $4\times$.} 
    \label{fig:super-resolution-noisy-ddpm-vis}
    \vspace{-10pt}
\end{figure}

\paragraph{Super-resolution.} We report qualitative noisy results for the VP-SDE model in~\cref{fig:super-resolution-noisy-ddpm-vis} and for the cond-OT model in~\cref{fig:super-resolution-noisy-ot-vis}. OT-ODE consistently achieves better FID, LPIPS and PSNR metrics compared to other methods for $\sigma_y=0.05$ (See \cref{fig:sigma-0.05-ddpm-ckpt,fig:sigma-0.05-cond-OT}). Similar to Gaussian deblurring, $\Pi$GDM tends to produce sharper edges. This is certainly desirable to achieve good super-resolution, but sometimes this results in unnatural textures in the images (See~\cref{fig:super-resolution-noisy-ddpm-vis}). RED-Diff for $\sigma_y=0.05$ gives slightly blurry images. In our experiments, we observe RED-Diff is sensitive to the values of $\sigma_y$, and we get good quality results for smaller values of $\sigma_y$, but the performance deteriorates with increase in value of $\sigma_y$. For $\sigma_y=0$, as shown in \cref{fig:sr-noiseless-ot-vis} and \cref{fig:sr-noiseless-ddpm-vis}, all the methods achieve comparable performance and the method declared best varies per metric and dataset. 

\begin{figure}[!ht]
    \centering
    \begin{subfigure}[t]{0.15\textwidth}
        \includegraphics[width=\textwidth]{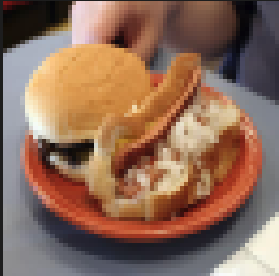}
    \end{subfigure}
    \begin{subfigure}[t]{0.15\textwidth}
        \includegraphics[width=\textwidth]{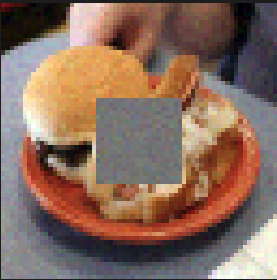}
    \end{subfigure}
    \begin{subfigure}[t]{0.15\textwidth}
        \includegraphics[width=\textwidth]{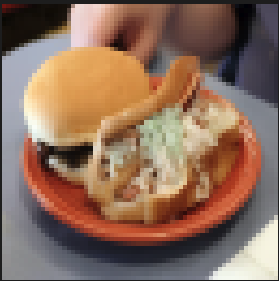}
    \end{subfigure}
    \begin{subfigure}[t]{0.15\textwidth}
    \includegraphics[width=\textwidth]{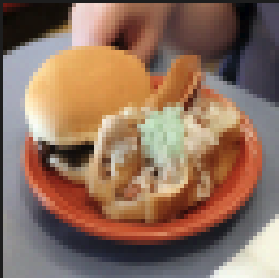}
    \end{subfigure}
    \begin{subfigure}[t]{0.15\textwidth}    
    \includegraphics[width=\textwidth]{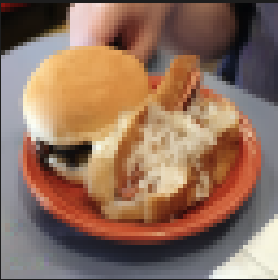}
    \end{subfigure} 
    \begin{subfigure}[t]{0.15\textwidth}    
        \includegraphics[width=\textwidth]{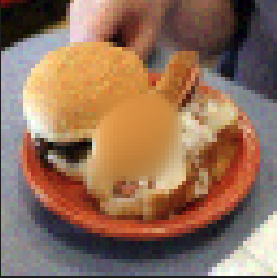}
    \end{subfigure}
    \\
        \begin{subfigure}[t]{0.15\textwidth}
        \includegraphics[width=\textwidth]{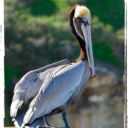}
    \end{subfigure}
    \begin{subfigure}[t]{0.15\textwidth}
        \includegraphics[width=\textwidth]{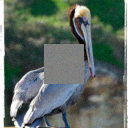}
    \end{subfigure}
    \begin{subfigure}[t]{0.15\textwidth}
        \includegraphics[width=\textwidth]{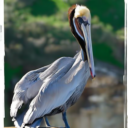}
    \end{subfigure}
    \begin{subfigure}[t]{0.15\textwidth}
        \includegraphics[width=\textwidth]{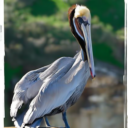}
    \end{subfigure}
    \begin{subfigure}[t]{0.15\textwidth}    
        \includegraphics[width=\textwidth]{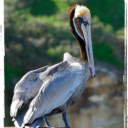}
    \end{subfigure}
    \begin{subfigure}[t]{0.15\textwidth}    
        \includegraphics[width=\textwidth]{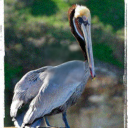}
    \end{subfigure}
    \\
    \begin{subfigure}[t]{0.15\textwidth}
        \includegraphics[width=\textwidth]{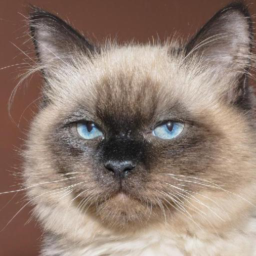}
        \caption{Reference}
    \end{subfigure}
    \begin{subfigure}[t]{0.15\textwidth}
        \includegraphics[width=\textwidth]{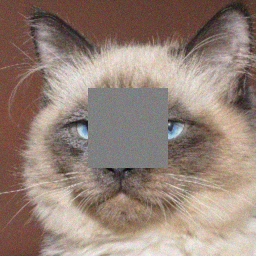}
        \caption{Distorted}
    \end{subfigure}
    \begin{subfigure}[t]{0.15\textwidth}
        \includegraphics[width=\textwidth]{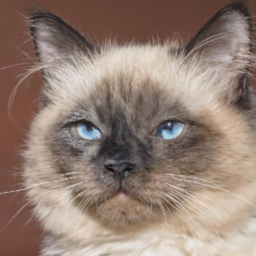}
        \caption{OT-ODE}
    \end{subfigure}
    \begin{subfigure}[t]{0.15\textwidth}
        \includegraphics[width=\textwidth]{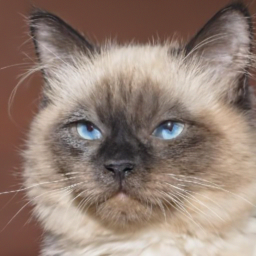}
        \caption{VP-ODE}
    \end{subfigure}
    \begin{subfigure}[t]{0.15\textwidth}    
        \includegraphics[width=\textwidth]{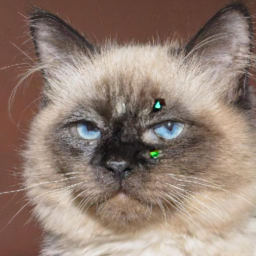}
        \caption{$\Pi$GDM}
    \end{subfigure}
    \begin{subfigure}[t]{0.15\textwidth}    
        \includegraphics[width=\textwidth]{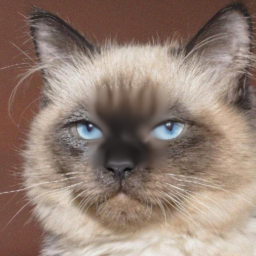}
        \caption{REDDiff}
    \end{subfigure}
    \caption{Results for inpainting (centered mask) with VP-SDE model and $\sigma_y=0.05$ for (\textbf{first row}) ImageNet-64, (\textbf{second row}) ImageNet-128, and (\textbf{third row}) AFHQ.} 
    \label{fig:ipc-noisy-ddpm-vis}
    \vspace{-12pt}
\end{figure}

\paragraph{Inpainting.} For centered mask inpainting, OT-ODE outperforms $\Pi$GDM and RED-Diff in terms of LPIPS, PSNR and SSIM across all datasets at $\sigma_y=0.05$. 
Regarding FID, OT-ODE performs comparably to or better than VP-ODE (See~\cref{fig:sigma-0.05-cond-OT,fig:sigma-0.05-ddpm-ckpt}). Similar observations hold true for inpainting with freeform mask on AFHQ. 
We present qualitative noisy results for the VP-SDE model in \cref{fig:ipc-noisy-ddpm-vis} and the cond-OT model in \cref{fig:inpainting-noisy-ot-vis}. As evident in these images, OT-ODE can result in more semantically meaningful inpainting (for instance, the shape of bird's neck, and shape of hot-dog bread in \cref{fig:ipc-noisy-ddpm-vis}). In contrast, the inpainted regions generated by RED-Diff tend be blurry and less semantically meaningful.  However, we note that OT-ODE (and VP-ODE) inpainting occasionally produces artifacts in the inpainted region as the resolution of image increases. We show examples of such negative inpainting results in \cref{sec:failure-modes-inpainting}. %
Empirically, we observe that performance of RED-Diff  and $\Pi$GDM improves as $\sigma_y$ decreases.  For $\sigma_y=0$, RED-Diff achieves higher PSNR and SSIM, but performs worse than OT-ODE in terms of FID and LPIPS (Refer to~\cref{fig:sigma-0-ddpm-ckpt}). 
OT-ODE's tendency to produce inpainting artifacts for higher resolution images remains for $\sigma_y=0$, and can occur for the same images as $\sigma_y=0.05$.  These artifacts 
can significantly degrade the pixel-based metrics PSNR and SSIM more than the perceptual metrics such as FID and LPIPS. %
We further note that noiseless inpainting for OT-ODE can be improved by incorporating null-space decomposition~\citep{ddnm}. We describe this adjustment in \cref{app:null_space}. 
\vspace{-2pt}

%% file: sections/related_work.tex
\section{Related Work}
The challenge of solving noisy linear inverse problems without any training has been tackled in many ways, often with other solution concepts than posterior sampling~\citep{elad2023image}.  Utilizing a diffusion model has a host of recent research that we build upon.  
Our state-of-the-art baselines $\Pi$GDM~\citep{pigdm} and RED-Diff~\citep{reddiff} correspond to lines of research in gradient-based corrections and variational inference.  

Earlier gradient-based corrections that approximate $\nabla_{\vx_t} \ln q(\vy | \vx_t)$ in various ways include Diffusion Posterior Sampling (DPS)~\citep{dps}, Manifold Constrained Gradient~\citep{chung2022improving}, and an annealed approximation~\citep{jalal2021robust}.  $\Pi$GDM out-performs earlier methods combining adaptive weights and Gaussian posterior approximation with discrete-time denoising diffusion implicit model (DDIM) sampling~\citep{ddim}.  Here we adapt $\Pi$GDM to all Gaussian probability paths and to flow sampling.  Our results show adaptive weights are unnecessary for strongly performing conditional OT flow sampling.  Denoising Diffusion Null Models (DDNM)~\citep{ddnm} proposed an alternative approximation of $\E_q[\vx_1 | \vx_t, \vy]$ using a null-space decomposition specific to linear inverse problems, which has been explored in combination with our method in Appendix~\ref{app:null_space}.%

RED-Diff~\citep{reddiff} approximates intractable $q(\vx_1 | \vy)$ directly using variational inference, solving for parameters via optimization.  RED-Diff was reported to have mode-seeking behavior confirmed by our results where RED-Diff performed better for noiseless inference.  Another earlier variational inference method %
is Denoising Diffusion Restoration Models (DDRM)~\citep{ddrm}.  DDRM showed SVD can be memory-efficient for image applications, and we adapt their SVD implementations for super-resolution and blur.  DDRM incorporates noiseless method ILVR~\citep{choi2021ilvr}, and leverages a measurement-dependent forward process (i.e. $q(\vx_t | \vy, \vx_1) \neq q(\vx_t | \vx_1)$) like earlier SNIPS~\citep{kawar2021snips}.  SNIPS collapses in special cases to variants proposed in \citet{score, score-sde, kadkhodaie2020solving} for linear inverse problems. 
\vspace{-10pt}

%% file: sections/conclusion.tex
\section{Discussion, Limitations, and Future work}

We have presented a training-free approach to solve linear inverse problems using flows that can leverage either pretrained diffusion or flow models.  
The algorithm is simple, stable, and requires no hyperparameter tuning when used with conditional OT probability paths.  
Our method combines past ideas from diffusion including $\Pi$GDM and early starting with the conditional OT probability path from flows, and our results demonstrate that this combination can solve inverse problems for both noisy and noiseless cases across a variety of datasets. 
Our algorithm using the conditional OT path (OT-ODE) produced results superior to the VP path (VP-ODE) and also to $\Pi$GDM and REDDiff for noisy inverse problems.  For the noiseless case, the perceptual quality from OT-ODE is on par with $\Pi$GDM for super-resolution and gaussian deblurring, but lagging for inpainting due to image artifacts.

Another important limitation, shared with most of the past related research, is a restriction to linear observations with scalar variance.  Our method can extend to arbitrary covariance, but non-linear observations are more complex.  %
Non-linear observations occur with image inverse tasks when utilizing latent, not pixel-space, diffusion or flow models.  
Applying our approach to such measurements requires devising an alternative $q^{app}(\vy | \vx_t)$.  
Another shared limitation is that we consider the non-blind setting with known $\mA$ and $\sigma_y$.  

Future research could tackle these limitations.  For non-linear observations in latent space, we could perhaps build upon \citet{rout2023solving} that uses a latent diffusion model for linear inverses.  For the blind setting, %
we might start from blind extensions to DPS and DDRM~\citep{Chung_2023_CVPR, murata2023gibbsddrm}.  As demonstrated here, we may be able to adapt and possibly improve these approaches via conversion to flow sampling using conditional OT paths.

%% file: sections/appendix.tex
\clearpage
\section{Proofs}
\label{sec:proofs}

For clarity, we restate Theorems and Lemmas from the main text before giving their proof.

\printProofs[conversion]

\section{Our method for any Gaussian probability path}
\label{sec:alg_for_arbitrary_path}

Algorithm~\ref{alg:cond-ot-flow-image-inversion} in the main text is specific to conditional OT probability paths.  Here we provide Algorithm~\ref{alg:arb-flow-image-inversion-denoiser} for any Gaussian probability path specified by Eq.~\ref{eq:gaussian_prob_paths}.  Algorithm~\ref{alg:cond-ot-flow-image-inversion} and Algorithm~\ref{alg:arb-flow-image-inversion-denoiser} are written assuming a denoiser $\widehat{\vx_1}(\vx_t)$ is provided from a pretrained diffusion model.  For completeness, we also include equivalent Algorithm~\ref{alg:arb-flow-image-inversion} that assumes $\widehat{\vv}(\vx_t)$ is provided from a pretrained flow model.  In all cases, the vector field or denoiser is evaluated only once per iteration.  

Our VP-ODE sampling results correspond to $\alpha_t$ and $\sigma_t$ given from the Variance-Preserving path, which can be found in~\citep{flowmatching}.  

\begin{algorithm}
\caption{A training-free approach to solve inverse problems via flows with a pretrained denoiser}\label{alg:arb-flow-image-inversion-denoiser}
\begin{algorithmic}[1]
\Require{Pretrained denoiser $\widehat{\vx_1}(\vx_t)$ converted to Gaussian probability path with $\alpha_t$ and $\sigma_t$ using Section~\ref{subsec:conversion}, noisy measurement $\vy$, measurement matrix $\mA$, initial time $t$, adaptive weights $\gamma_t$, and std $\sigma_y$}
\State{Initialize $\vx_t = \alpha_t \rvy + \sigma_t\epsilon$, where $\epsilon \sim \mathcal{N}(0, \mI)$}
\Comment{Initialize $\vx_t$, \cref{eq:initialization}}
\State{$\vz_t = \vx_t$}
\For{each time step $t'$ of ODE integration} \Comment{Integrate ODE from $t' = t$ to $1$.}
    \State{$r_{t'}^2 = \frac{\sigma_{t'}^2}{\sigma_{t'}^2 + \alpha_{t'}^2}$} \Comment{Value of $r_t^2$ from \cref{eqn:r_t_sqr_flow_value}}
    \State{$\widehat{\vv} = \left(\alpha_t \frac{d \ln (\alpha_t / \sigma_t)}{dt}\right)\widehat{\vx_1} + \frac{d \ln \sigma_t}{dt}\vz_t$} \Comment{Convert $\widehat{\vx_1}$ to vector field, \cref{eq:interchange}}
    \State{$\vg = (\vy - \mA \widehat{\vx_1})^\top (r_{t'}^2\mA \mA^\top + \sigma_y^2\mI)^{-1} \mA \frac{\partial \widehat{\vx_1}}{\partial \vz_{t'}}$} \Comment{$\Pi$GDM correction}
    \State{$\widehat{\vv}_{\text{corrected}} = \widehat{\vv} + \sigma_t^2 \frac{d \ln(\alpha_t / \sigma_t)}{dt}\gamma_t \vg$} \Comment{Correct unconditional vector field $\widehat{\vv}$, \cref{eq:vec_alg_corr}}
\EndFor
\end{algorithmic}
\end{algorithm}

\begin{algorithm}
\caption{A training-free approach to solve inverse problems via flows with a pretrained vector field}\label{alg:arb-flow-image-inversion}
\begin{algorithmic}[1]
\Require{Pretrained vector field $\widehat{\vv}(\vx_t)$ converted to Gaussian probability path with $\alpha_t$ and $\sigma_t$ using Section~\ref{subsec:conversion}, noisy measurement $\vy$, measurement matrix $\mA$, initial time $t$, adaptive weights $\gamma_t$, and std $\sigma_y$}
\State{Initialize $\vx_t = \alpha_t \rvy + \sigma_t\epsilon$, where $\epsilon \sim \mathcal{N}(0, \mI)$}
\Comment{Initialize $\vx_t$, \cref{eq:initialization}}
\State{$\vz_t = \vx_t$}
\For{each time step $t'$ of ODE integration} \Comment{Integrate ODE from $t' = t$ to $1$.}
    \State{$\widehat{\vv} = \widehat{\vv}(\vz_{t'})$} \Comment{$\vz_{t'}$ is value of $\vx_t$ at time $t'$ during ODE integration}
    \State{$r_{t'}^2 = \frac{\sigma_{t'}^2}{\sigma_{t'}^2 + \alpha_{t'}^2}$} \Comment{Value of $r_t^2$ from \cref{eqn:r_t_sqr_flow_value}}
    \State{$\widehat{\vx_1} = \left(\alpha_t \frac{d \ln (\alpha_t / \sigma_t)}{dt}\right)^{-1}\left(\widehat{\vv}  - \frac{d \ln \sigma_t}{dt}\vz_t\right)$} \Comment{Convert vector field to $\widehat{\vx_1}$, \cref{eq:interchange}}
    \State{$\vg = (\vy - \mA \widehat{\vx_1})^\top (r_{t'}^2\mA \mA^\top + \sigma_y^2\mI)^{-1} \mA \frac{\partial \widehat{\vx_1}}{\partial \vz_{t'}}$} \Comment{$\Pi$GDM correction}
    \State{$\widehat{\vv}_{\text{corrected}} = \widehat{\vv} + \sigma_t^2 \frac{d \ln(\alpha_t / \sigma_t)}{dt}\gamma_t \vg$} \Comment{Correct unconditional vector field $\widehat{\vv}$, \cref{eq:vec_alg_corr}}
\EndFor
\end{algorithmic}
\end{algorithm}

\clearpage
\section{Ablation Study} \label{sec:appendix-ablation}
\paragraph{Choice of initialization.} We initialize the flow at time $t > 0$ as 
$\vx_t = \alpha_t \vy + \sigma_t \epsilon$ (y-init) where $\epsilon \sim \mathcal{N}(0, \mI)$. Another choice of initialization is to use $
\vx_t = \alpha_t \mA^\dagger \vy + \sigma_t \epsilon$. However, empirically we find that this initialization performs worse that y-init on cond-OT model with OT-ODE sampling. We summarize the results of our ablation study in \cref{tab:abl-afhq-ot-init}. We find that on Gaussian deblurring, initialization with  $\mA^\dagger \vy$ does worse than y-init, while the performance of both the initializations is comparable for super-resolution. In all our experiments, we use y-init, due to its better performance on Gaussian deblurring.
\begin{table*}[!htbp]
    \centering
    \caption{Quantitative evaluation of choice of initialization for conditional OT flow model with OT-ODE sampling on AFHQ dataset. We find that y-init outperforms $\mA^\dagger \vy$ on Gaussian deblurring.}
    \label{tab:abl-afhq-ot-init}
    \resizebox{\textwidth}{!}{
    \begin{tabular}{ccccccccccc}
    \toprule 
    \multirow{3}{*}{Initialization} & \multirow{3}{*}{Start time} & \multirow{3}{*}{NFEs $\downarrow$} &\multicolumn{4}{c}{Gaussian deblur, $\sigma_y=0.05$} &\multicolumn{4}{c}{SR 4$\times$, $\sigma_y=0.05$} \\
    \cmidrule(lr){4-7}\cmidrule(lr){8-11}
    & & & FID $\downarrow$ & LPIPS $\downarrow$ & PSNR $\uparrow$ &  SSIM $\uparrow$ & FID $\downarrow$ & LPIPS $\downarrow$ & PSNR $\uparrow$ &  SSIM $\uparrow$\\
    \midrule
    y init & 0.2 & 100 & \textbf{7.57} & \textbf{0.268} & \textbf{30.28} & \textbf{0.626} & \textbf{6.03} & \textbf{0.219} & \textbf{31.12} & \textbf{0.739} \\
    $\mA^\dagger \vy$ & 0.1 & 100 & 41.22 & 0.449 & 28.79 & 0.392 & 12.93 & 0.292 & 30.46 & 0.664\\
    $\mA^\dagger \vy$ & 0.2 & 100 & 56.42 & 0.554 & 28.11 & 0.249 & 6.09 & 0.219 & 31.12 & 0.739 \\
    \bottomrule
    \end{tabular}
    }
\end{table*}

\paragraph{Ablation over $\gamma_t$ for VP-ODE sampling.} We compare the performance of $\gamma_t=1$ against $\gamma_t = \sqrt{\frac{\alpha_t}{\alpha_t^2 + \sigma_t^2}}$. We show results of VP-ODE sampling with VP-SDE model in \cref{tab:abl-inet128-gamma_t-vpode-1} and \cref{tab:abl-inet128-gamma_t-vpode-2}. As seen our choice of $\gamma_t$ outperform $\gamma_t=1$ across all the metrics on face-blurred ImageNet-128.

\begin{table*}[!htbp]
    \centering
    \caption{Quantitative evaluation of value of $\gamma_t$ in VP-ODE sampling with VP-SDE model on face-blurred ImageNet-$128$ dataset.}
    \label{tab:abl-inet128-gamma_t-vpode-1}
    \resizebox{\textwidth}{!}{
    \begin{tabular}{ccccccccccc}
    \toprule 
    \multirow{3}{*}{$\gamma_t$} & \multirow{3}{*}{Start time} & \multirow{3}{*}{NFEs $\downarrow$} &\multicolumn{4}{c}{SR 2$\times$, $\sigma_y=0.05$} &\multicolumn{4}{c}{Gaussian deblur, $\sigma_y=0.05$} \\
    \cmidrule(lr){4-7}\cmidrule(lr){8-11}
    & & & FID $\downarrow$ & LPIPS $\downarrow$ & PSNR $\uparrow$ &  SSIM $\uparrow$ & FID $\downarrow$ & LPIPS $\downarrow$ & PSNR $\uparrow$ &  SSIM $\uparrow$\\
    \midrule
    $1$ & 0.4 & 60 &  32.66 & 0.371 & 29.06 & 0.530 & 29.31 & 0.346 & 29.12 & 0.554 \\
    $\sqrt{\frac{\alpha_t}{\alpha_t^2 + \sigma_t^2}}$ & 0.4 & 60 & \textbf{9.14} &  \textbf{0.167} & \textbf{32.06} & \textbf{0.838} & \textbf{10.14} & \textbf{0.196} & \textbf{31.59} & \textbf{0.800} \\
    \bottomrule
    \end{tabular}
    }
\end{table*}

\begin{table*}[!htbp]
    \centering
    \caption{Quantitative evaluation of value of $\gamma_t$ in VP-ODE sampling with VP-SDE model on face-blurred ImageNet-$128$ dataset.}
    \label{tab:abl-inet128-gamma_t-vpode-2}
    \resizebox{\textwidth}{!}{
    \begin{tabular}{ccccccccccc}
    \toprule 
    \multirow{3}{*}{$\gamma_t$} & \multirow{3}{*}{Start time} & \multirow{3}{*}{NFEs $\downarrow$} &\multicolumn{4}{c}{Inpainting-\textit{Center}, $\sigma_y=0.05$} &\multicolumn{4}{c}{Denoising, $\sigma_y=0.05$} \\
    \cmidrule(lr){4-7}\cmidrule(lr){8-11}
    & & & FID $\downarrow$ & LPIPS $\downarrow$ & PSNR $\uparrow$ &  SSIM $\uparrow$ & FID $\downarrow$ & LPIPS $\downarrow$ & PSNR $\uparrow$ &  SSIM $\uparrow$\\
    \midrule
    $1$ & 0.3 & 70 & 53.03 & 0.285 & 31.55 & 0.737 & 28.37 & 0.238 & 31.63 & 0.786 \\
    $\sqrt{\frac{\alpha_t}{\alpha_t^2 + \sigma_t^2}}$ & 0.3 & 70 & \textbf{8.47} & \textbf{0.129} & \textbf{34.43} & \textbf{0.876} & \textbf{5.83} & \textbf{0.087} & \textbf{35.85} & \textbf{0.938} \\
    \bottomrule
    \end{tabular}
    }
\end{table*}

\begin{figure}[!htbp]
    \centering
    \begin{minipage}[b]{0.65\textwidth}
        \includegraphics[width=0.44\textwidth]{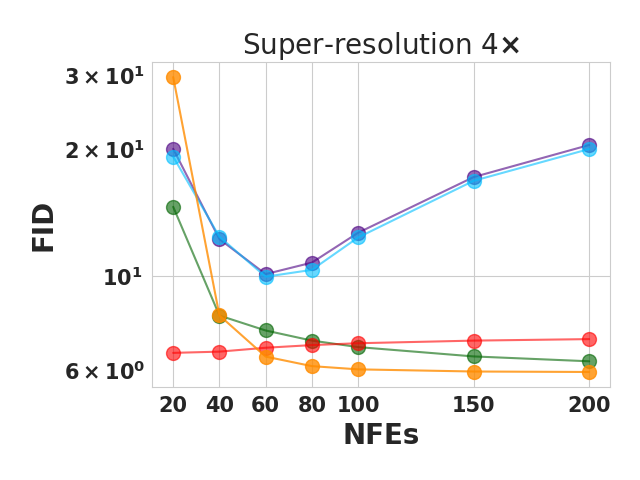}
        \includegraphics[width=0.44\textwidth]{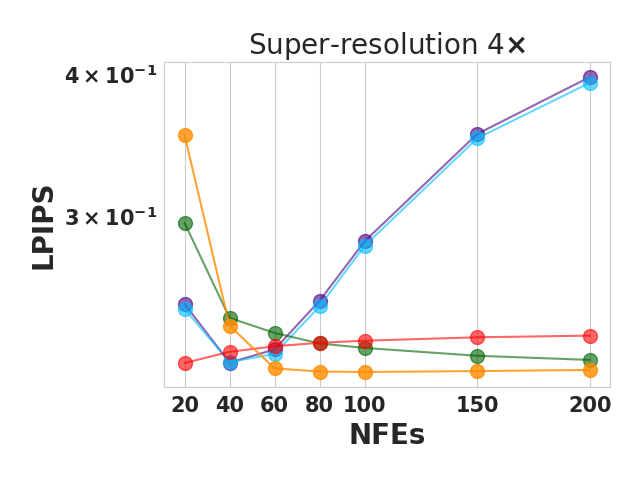}
        \includegraphics[width=0.44\textwidth]{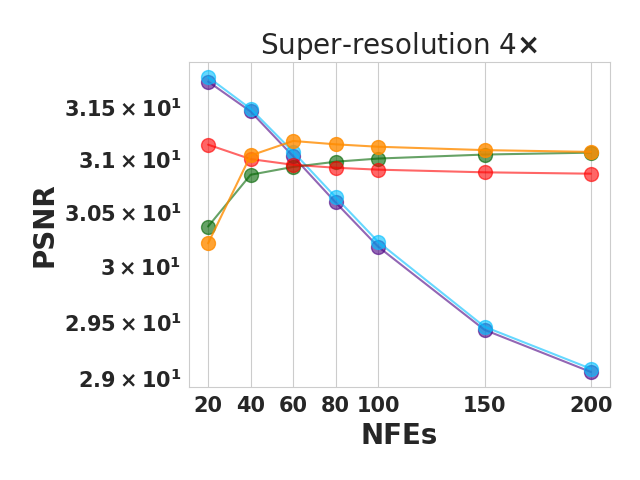}
        \includegraphics[width=0.44\textwidth]{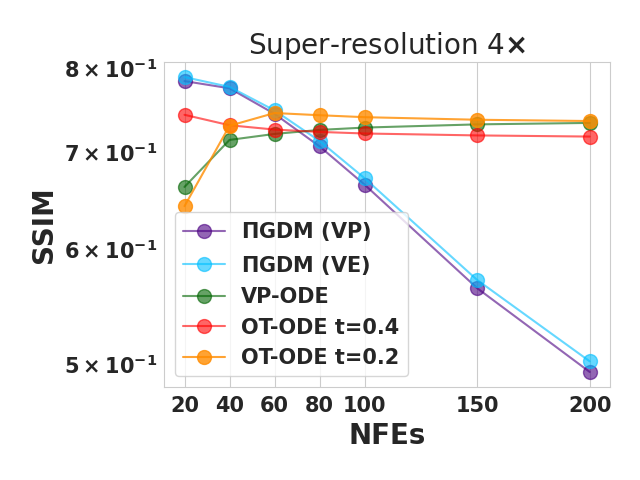}
    \end{minipage}
    \centering
    \begin{minipage}[b]{0.65\textwidth}
        \includegraphics[width=0.44\textwidth]{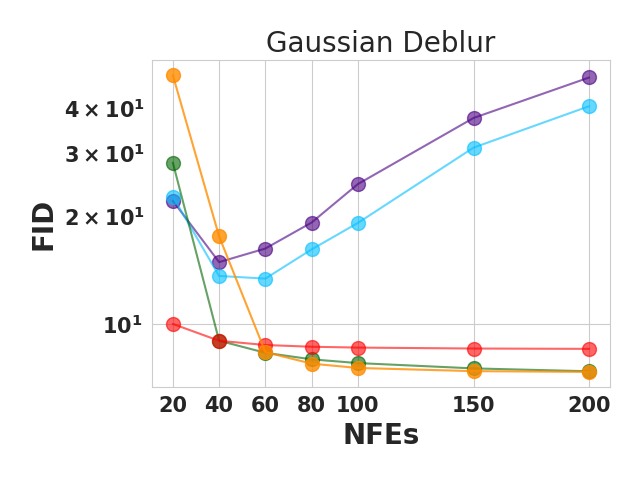}
        \includegraphics[width=0.44\textwidth]{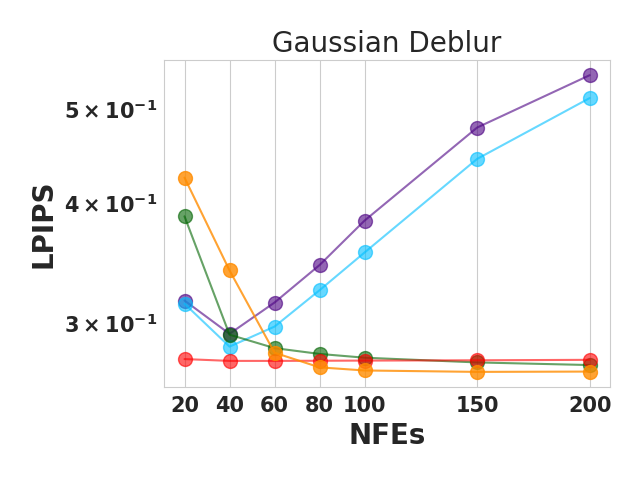}
        \includegraphics[width=0.44\textwidth]{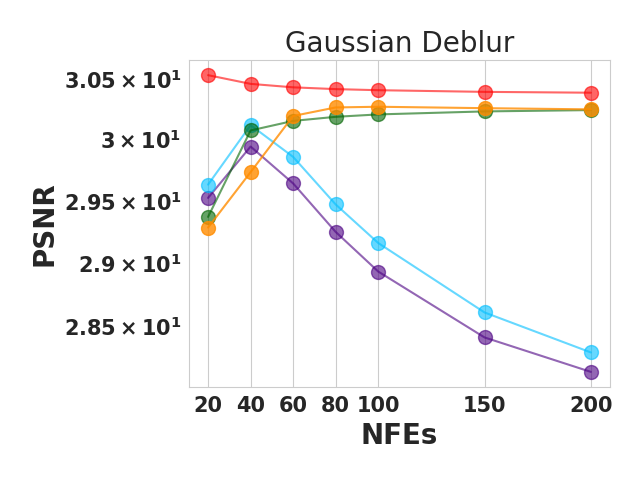}
        \includegraphics[width=0.44\textwidth]{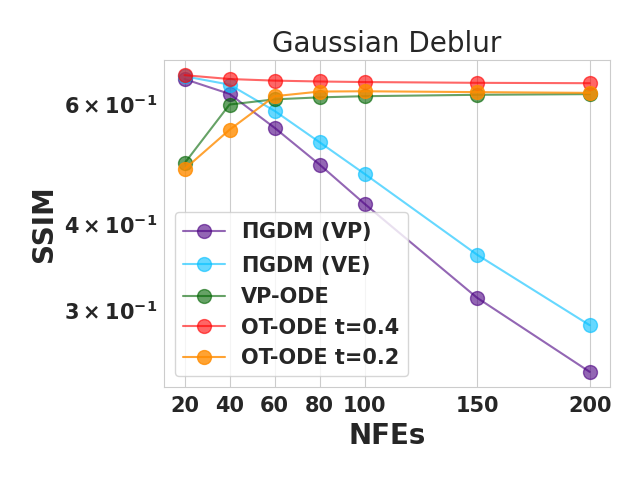}
    \end{minipage}
    \centering
    \begin{minipage}[b]{0.65\textwidth}
        \includegraphics[width=0.44\textwidth]{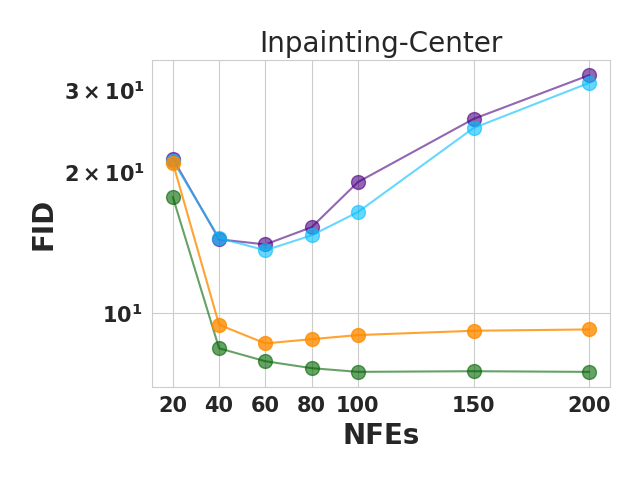}
        \includegraphics[width=0.44\textwidth]{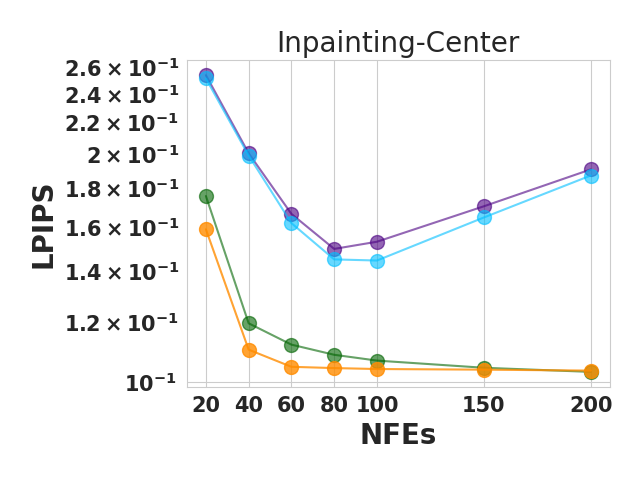}
        \includegraphics[width=0.44\textwidth]{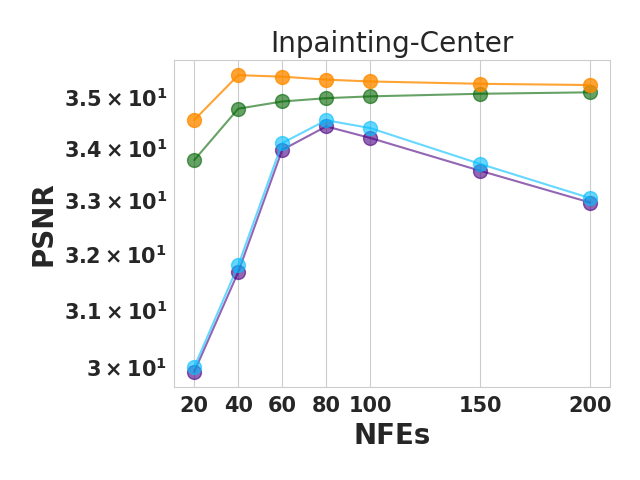}
        \includegraphics[width=0.44\textwidth]{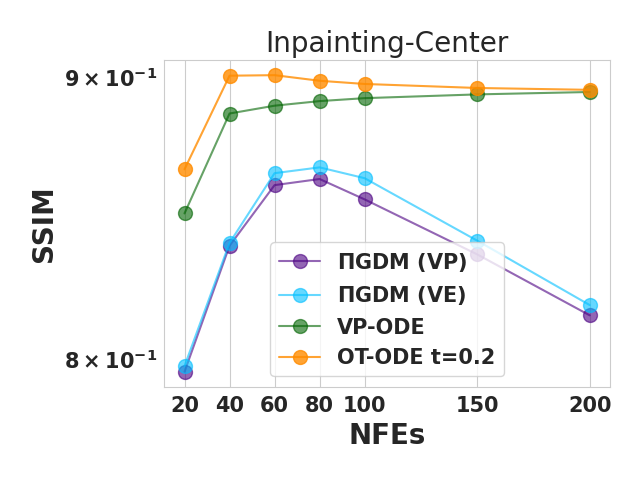}
    \end{minipage}
    \caption{Performance of different procedures for solving linear inverse  problems with variation in NFEs on AFHQ dataset. We use pretrained conditional OT model and set $\sigma_y=0.05$. The legends VP and VE indicate the choice of $r_t^2$ used in $\Pi$GDM (See \cref{sec:pigdm-impl-details}). Time $t=0.2$ and $0.4$ indicates the starting time of sampling with OT-ODE.}
    \label{fig:nfe-sr-gb-ot-noisy-0.05}
\end{figure}

\paragraph{Variation of performance with NFEs.} We analyze the variation in performance of OT-ODE, VP-ODE and $\Pi$GDM for solving linear inverse problems as NFEs are varied. 
The results have been summarized in \cref{fig:nfe-sr-gb-ot-noisy-0.05}. We observe that OT-ODE consistently outperforms VP-ODE and $\Pi$GDM across all measurements in terms of FID and LPIPS metrics, even for NFEs as small as $20$. 
We also note that the choice of starting time matters to achieve good performance with OT-ODE. For instance, starting at $t=0.4$ outperforms $t=0.2$ when NFEs are small, but eventually as NFEs is increased, $t=0.2$ performs better. 
We also note that $\Pi$GDM achieves higher values of PSNR and SSIM at smaller NFEs for super-resolution but has inferior FID and LPIPS compared to OT-ODE.

\begin{figure}[!htb]
    \centering
    \begin{minipage}[b]{\textwidth}
        \includegraphics[width=0.24\textwidth]{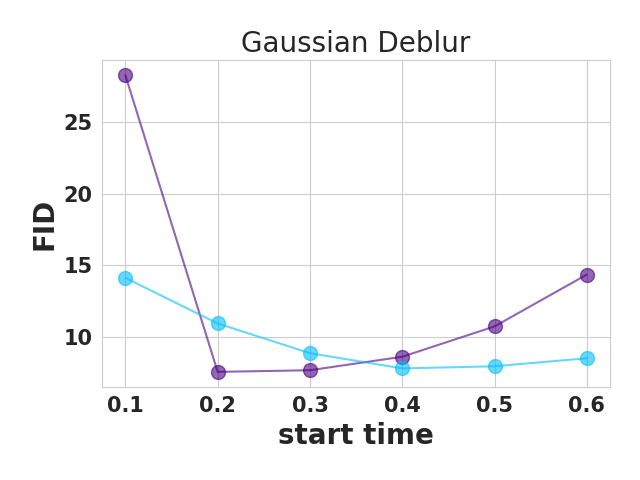}
        \includegraphics[width=0.24\textwidth]{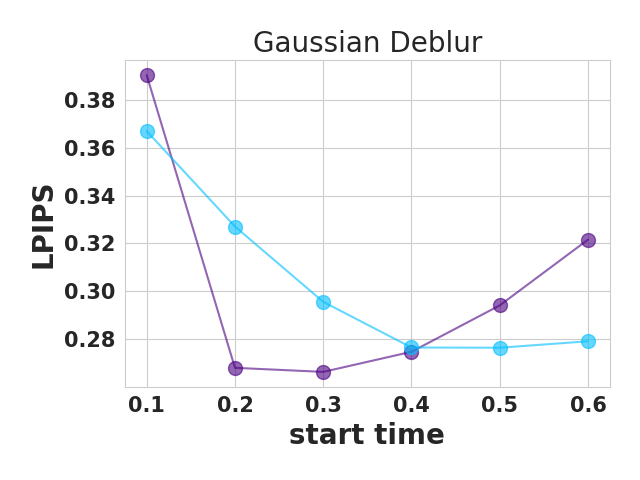}
        \includegraphics[width=0.24\textwidth]{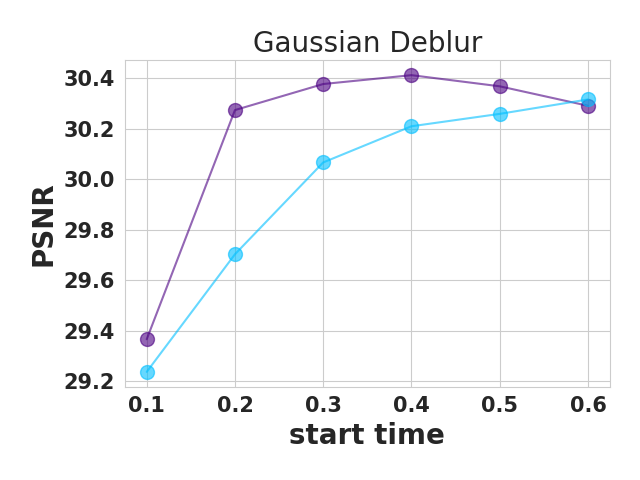}
        \includegraphics[width=0.24\textwidth]{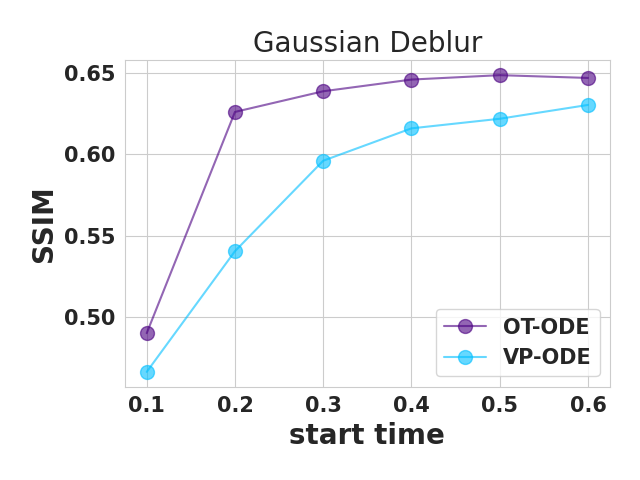}
    \end{minipage}
    \begin{minipage}[b]{\textwidth}
        \includegraphics[width=0.24\textwidth]{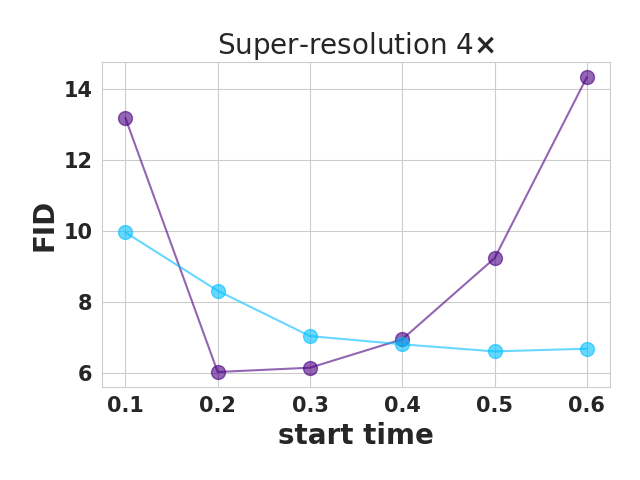}
        \includegraphics[width=0.24\textwidth]{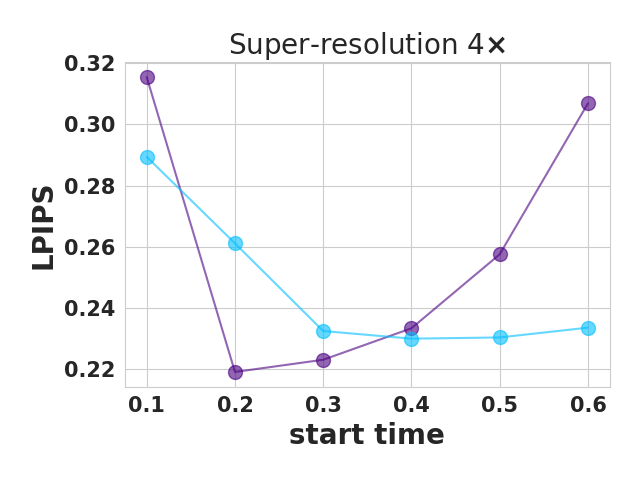}
        \includegraphics[width=0.24\textwidth]{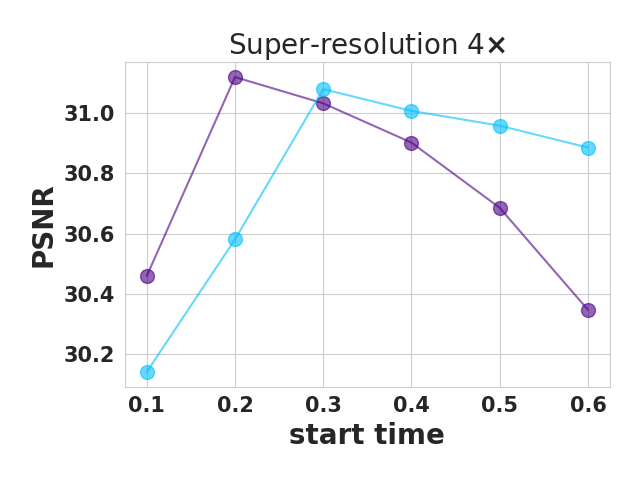}
        \includegraphics[width=0.24\textwidth]{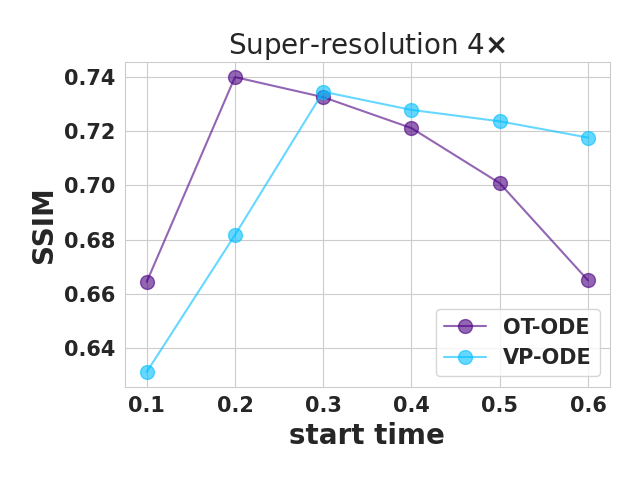}
    \end{minipage}
    \begin{minipage}[b]{\textwidth}
        \includegraphics[width=0.24\textwidth]{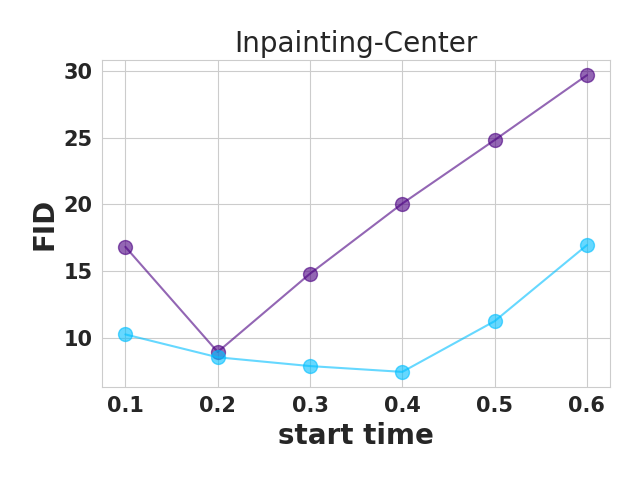}
        \includegraphics[width=0.24\textwidth]{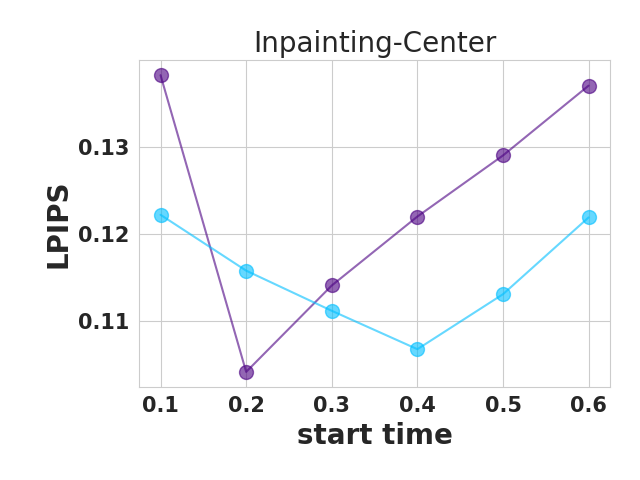}
        \includegraphics[width=0.24\textwidth]{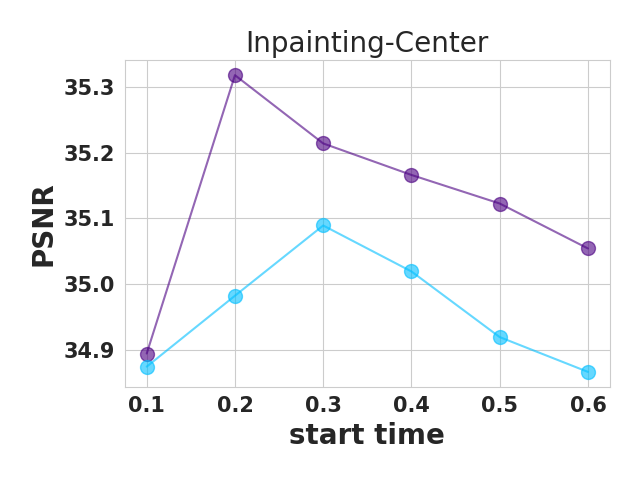}
        \includegraphics[width=0.24\textwidth]{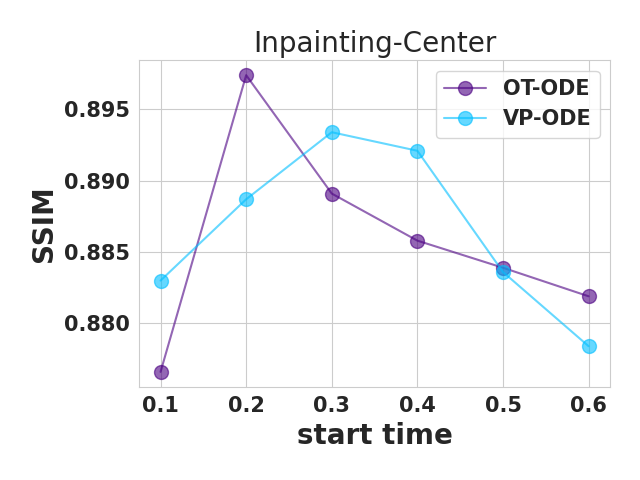}
    \end{minipage}
    \caption{Performance of OT-ODE and VP-ODE in solving linear inverse problems with varying start times on AFHQ dataset. We use pretrained cond-OT model and set $\sigma_y=0.05$.}
    \label{fig:st-time-ot-noisy-0.05}
\end{figure}

\begin{figure}[!htb]
    \centering
    \begin{minipage}[b]{\textwidth}
        \includegraphics[width=0.24\textwidth]{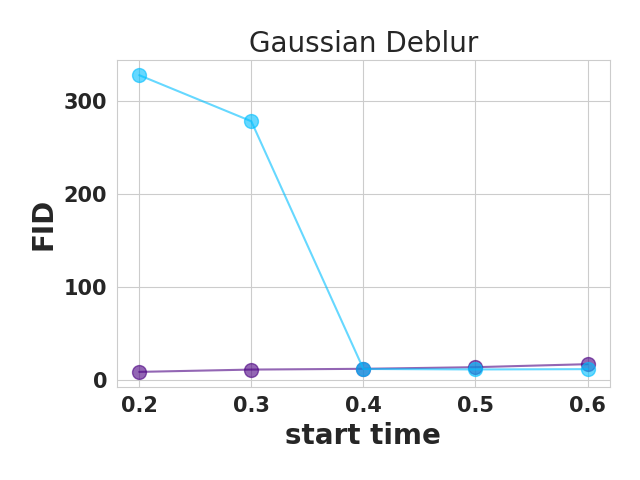}
        \includegraphics[width=0.24\textwidth]{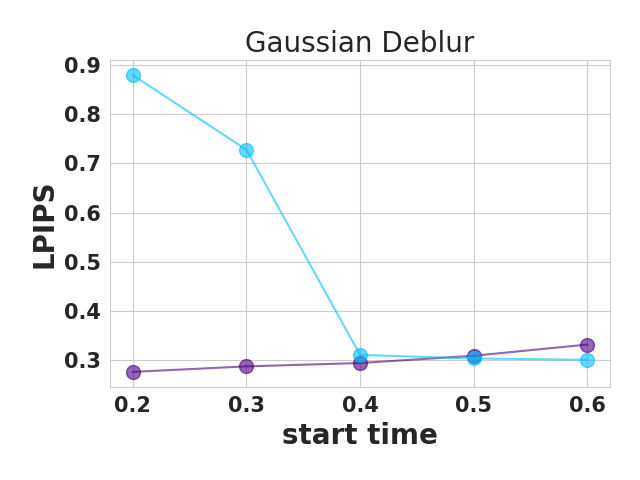}
        \includegraphics[width=0.24\textwidth]{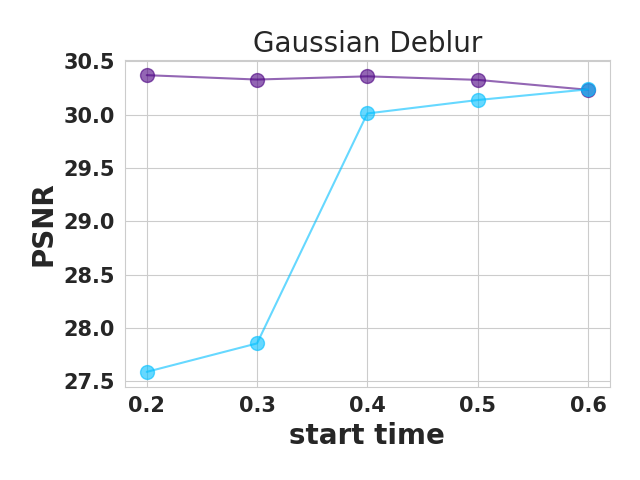}
        \includegraphics[width=0.24\textwidth]{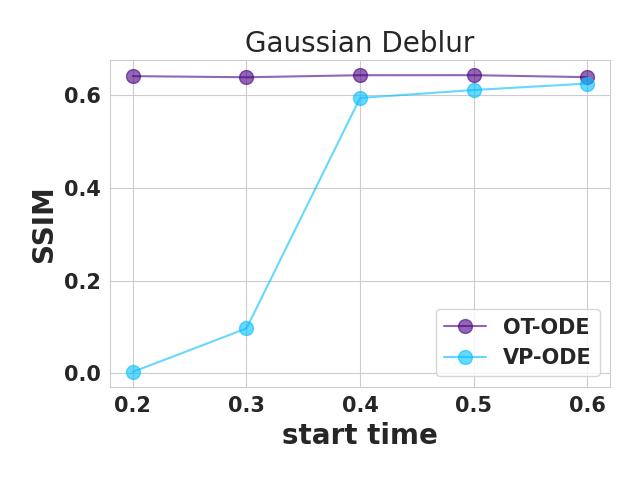}
    \end{minipage}
    \begin{minipage}[b]{\textwidth}
        \includegraphics[width=0.24\textwidth]{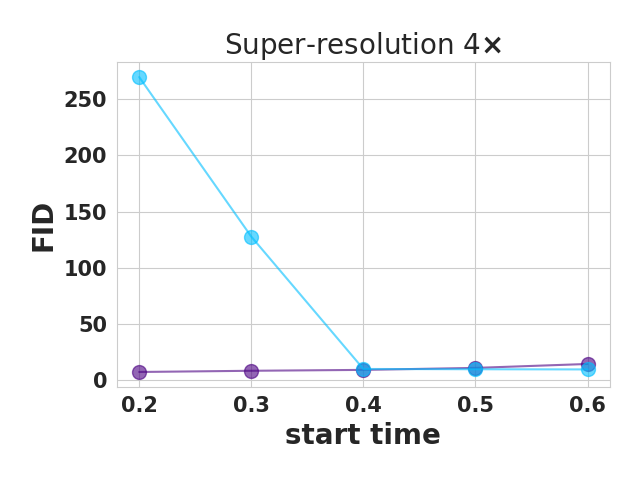}
        \includegraphics[width=0.24\textwidth]{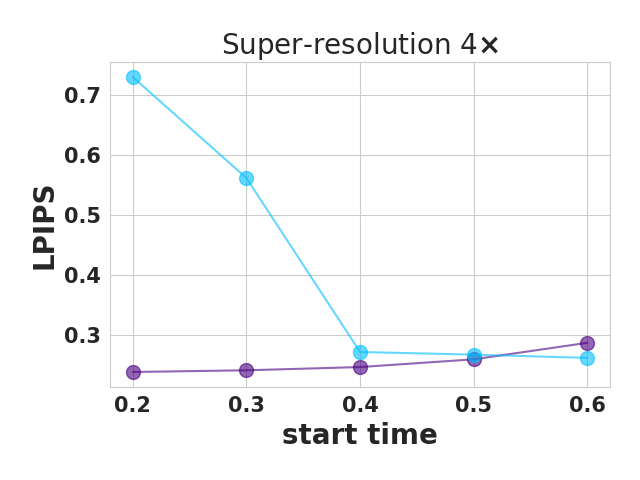}
        \includegraphics[width=0.24\textwidth]{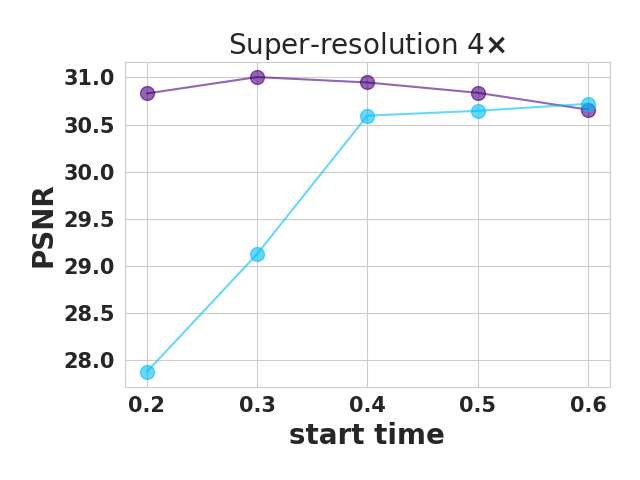}
        \includegraphics[width=0.24\textwidth]{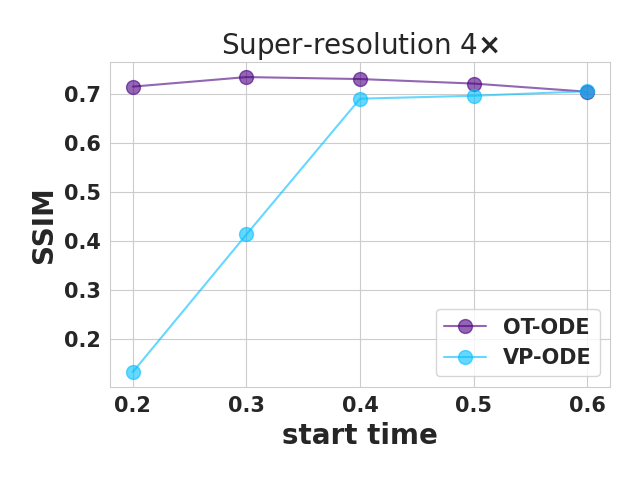}
    \end{minipage}
    \begin{minipage}[b]{\textwidth}
        \includegraphics[width=0.24\textwidth]{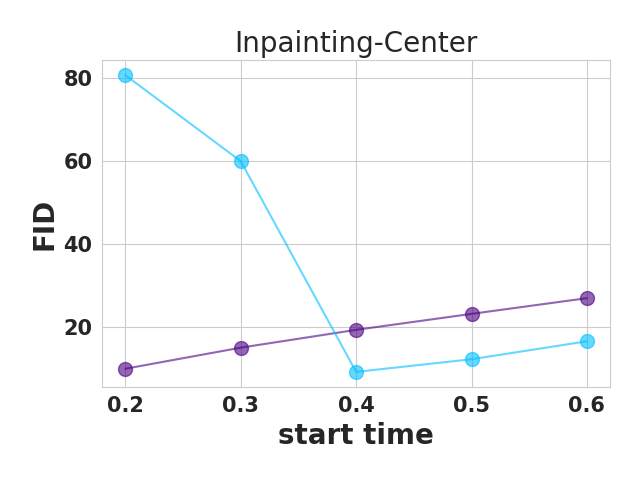}
        \includegraphics[width=0.24\textwidth]{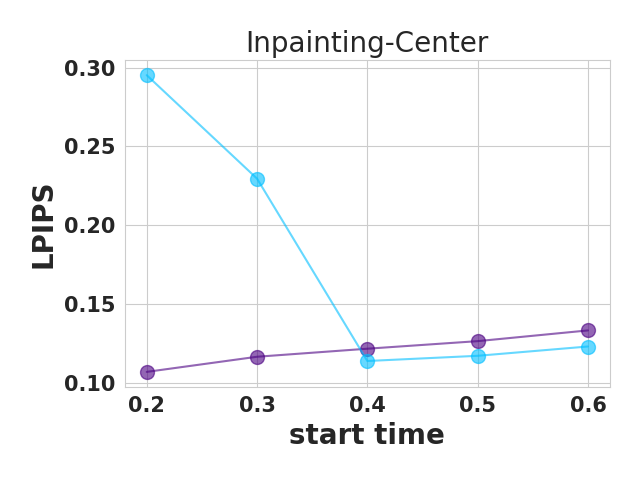}
        \includegraphics[width=0.24\textwidth]{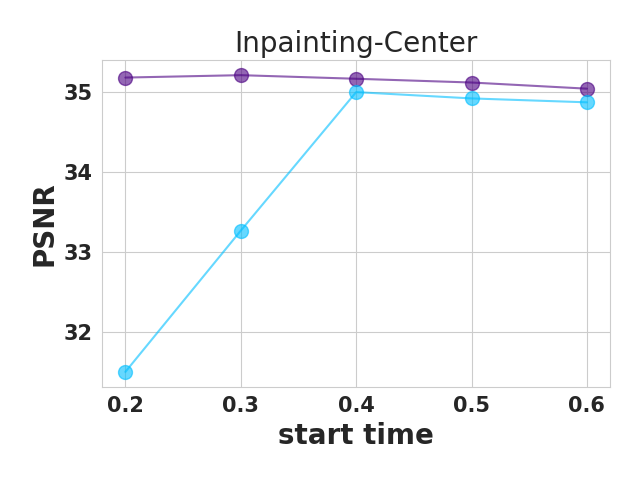}
        \includegraphics[width=0.24\textwidth]{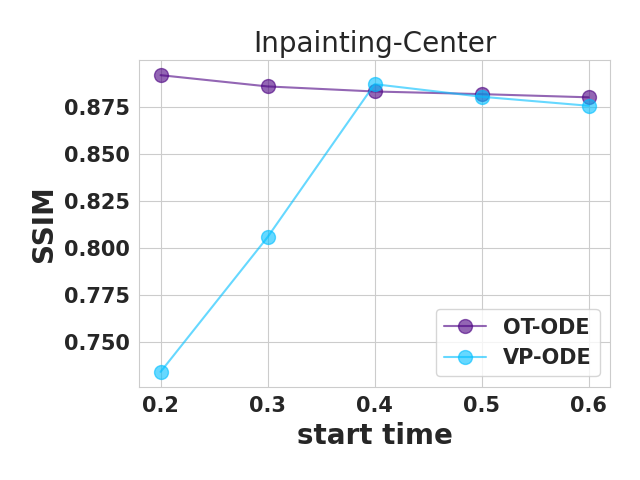}
    \end{minipage}
    \caption{Performance of OT-ODE and VP-ODE in solving linear inverse problems with varying start times on AFHQ dataset. We use pretrained VP-SDE model and set $\sigma_y=0.05$.}
    \label{fig:st-time-vpsde-noisy-0.05}
\end{figure}

\paragraph{Choice of starting time.} We plot the variation in performance of OT-ODE  and VP-ODE sampling with change in start times for conditional OT model and VP-SDE model on AFHQ dataset in \cref{fig:st-time-ot-noisy-0.05} and \cref{fig:st-time-vpsde-noisy-0.05}, respectively. We note that in general, OT-ODE sampling achieves optimal performance across all measurements and all metrics at $t=0.2$ while VP-ODE sampling achieves optimal performance between start times of $t=0.3$ and $0.4$. In this work, for all the experiments, we use $t=0.2$ for OT-ODE sampling and $t=0.4$ for VP-ODE sampling. 

\clearpage
\input{sections/empirical_results_table}

\input{sections/additional_qualitative_results}
\input{sections/failure_cases_rev}
\clearpage
\input{sections/nrsd}
\clearpage
\section{Baselines} \label{sec:baselines}
\input{sections/pigdm_impl}

\input{sections/reddiff_impl}
\input{sections/additional_background}

%% file: sections/empirical_results_table.tex
\section{Additional Empirical Results}\label{sec:empirical-results}

\begin{figure}[!htb]
    \begin{minipage}[b]{\textwidth}
    \centering
        \includegraphics[width=0.49\textwidth]{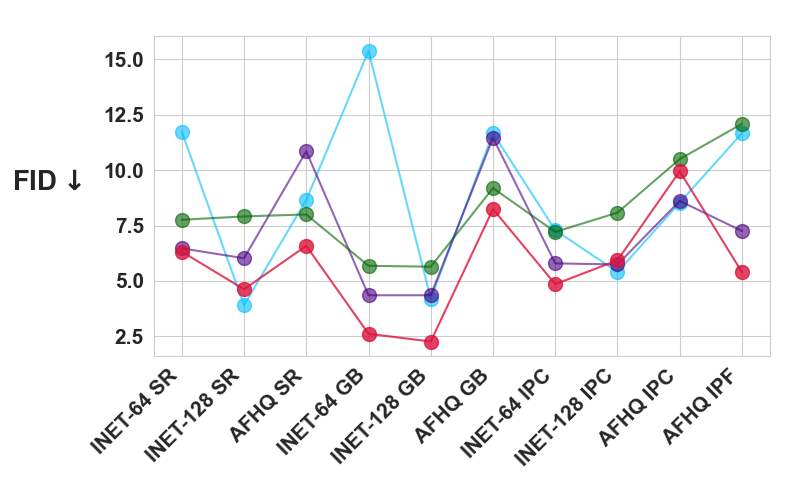}
        \includegraphics[width=0.49\textwidth]{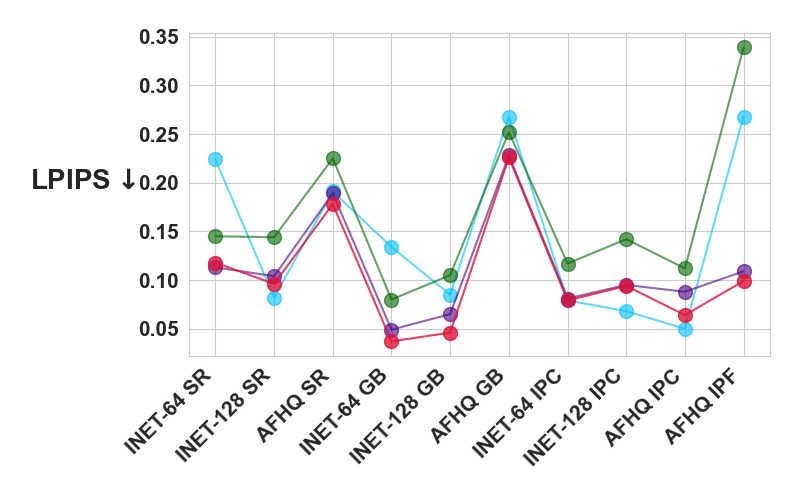}
        \includegraphics[width=0.49\textwidth]{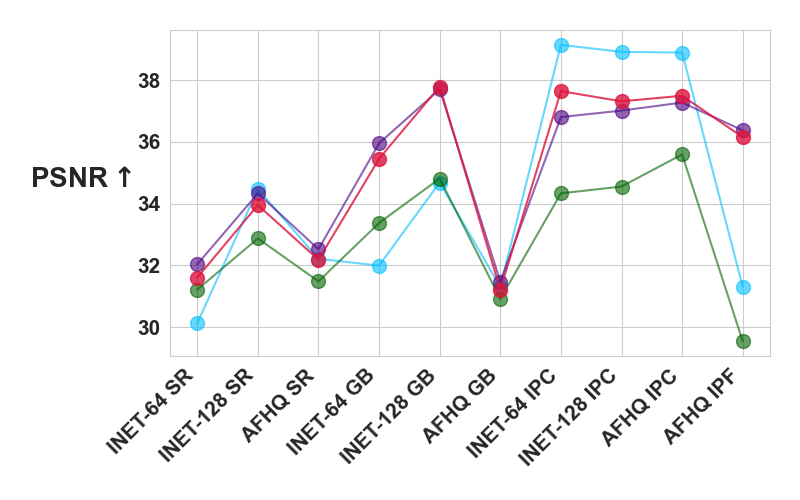}
        \includegraphics[width=0.49\textwidth]{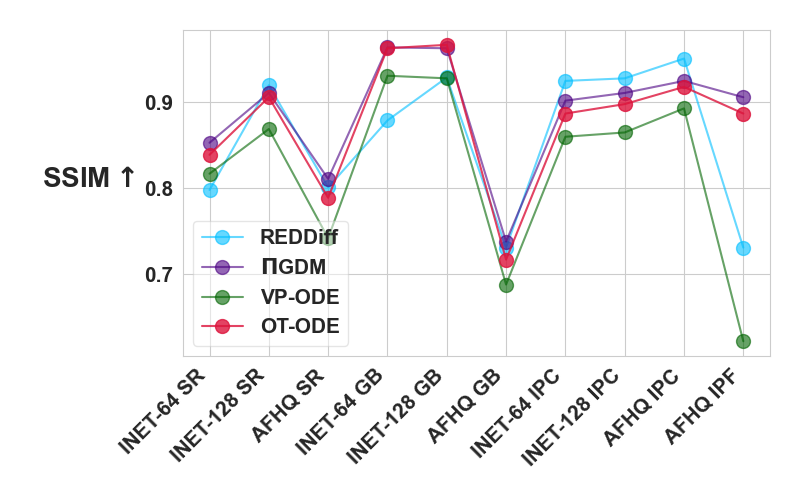}
    \end{minipage}
    \caption{Quantitative evaluation of pretrained VP-SDE model for linear inverse problems on super-resolution (SR), gaussian deblurring (GB), image inpainting - centered mask (IPC) and inpainting - free-form (IPF) with $\sigma_y=0$. We show results on face-blurred ImageNet-64 (INET-64), face-blurred ImageNet-128 (INET-128), and AFHQ-256 (AFHQ).}
    \label{fig:sigma-0-ddpm-ckpt}
\end{figure}

\begin{figure}[!thb]
    \begin{minipage}[b]{\textwidth}
    \centering
        \includegraphics[width=0.49\textwidth]{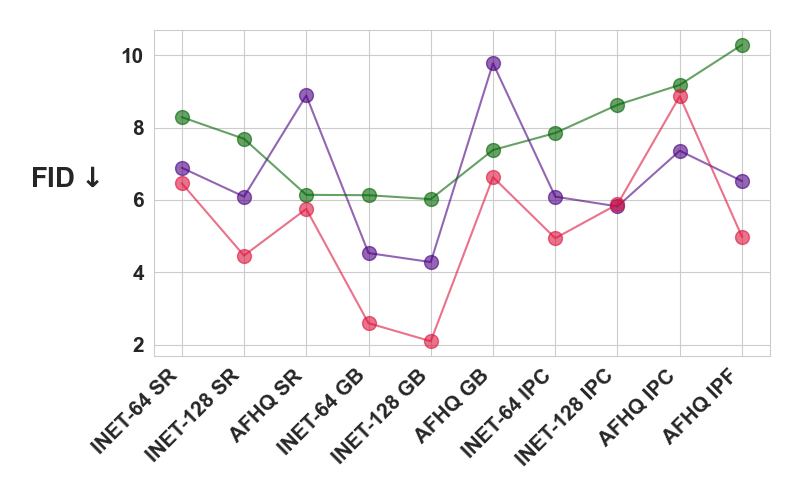}
        \includegraphics[width=0.49\textwidth]{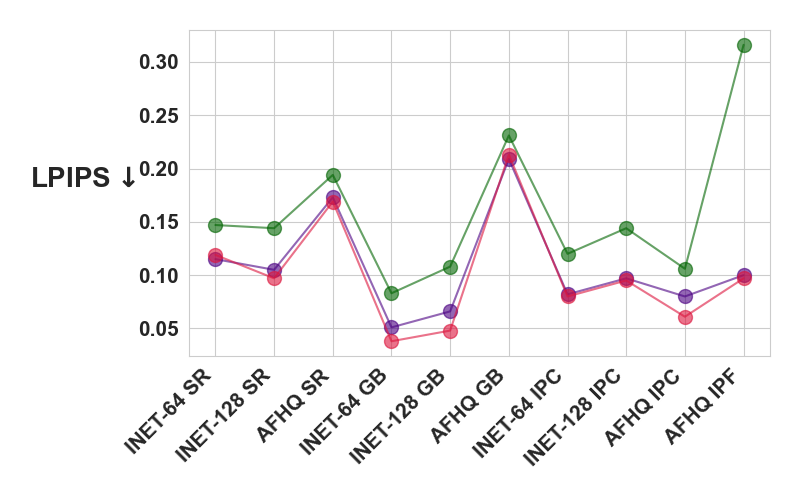}
        \includegraphics[width=0.49\textwidth]{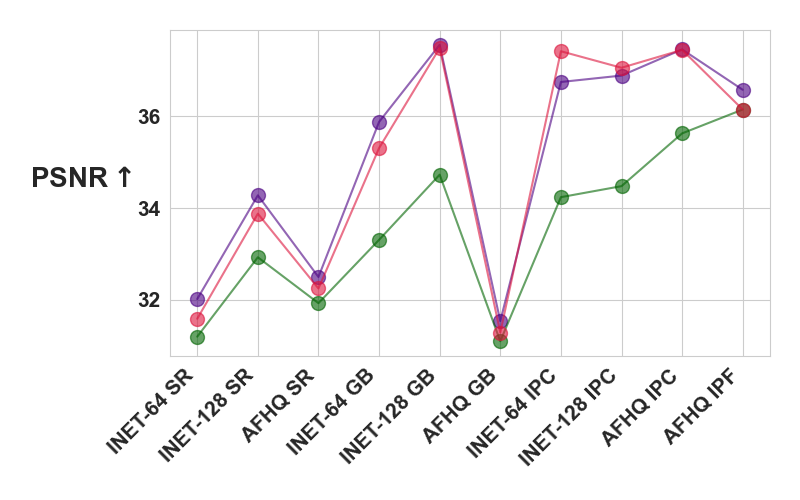}
        \includegraphics[width=0.49\textwidth]{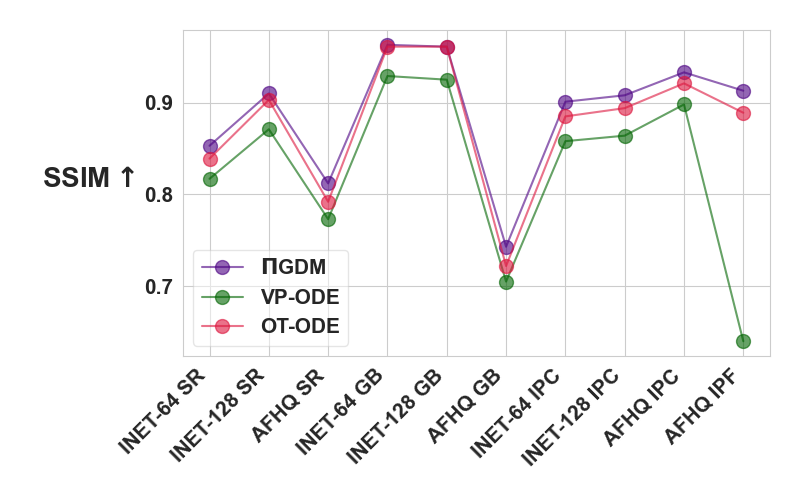}
    \end{minipage}
    \caption{Quantitative evaluation of pretrained conditional OT model for linear inverse problems on super-resolution (SR), gaussian deblurring (GB), image inpainting - centered mask (IPC) and inpainting - freeform (IPF) with $\sigma_y=0$. We show results on face-blurred ImageNet-64 (INET-64), face-blurred ImageNet-128 (INET-128), and AFHQ-256 (AFHQ).}
    \label{fig:sigma-0-ot-ckpt}
\end{figure}

\begin{figure}[!thb]
    \begin{minipage}[b]{\textwidth}
    \centering
        \includegraphics[width=0.49\textwidth]{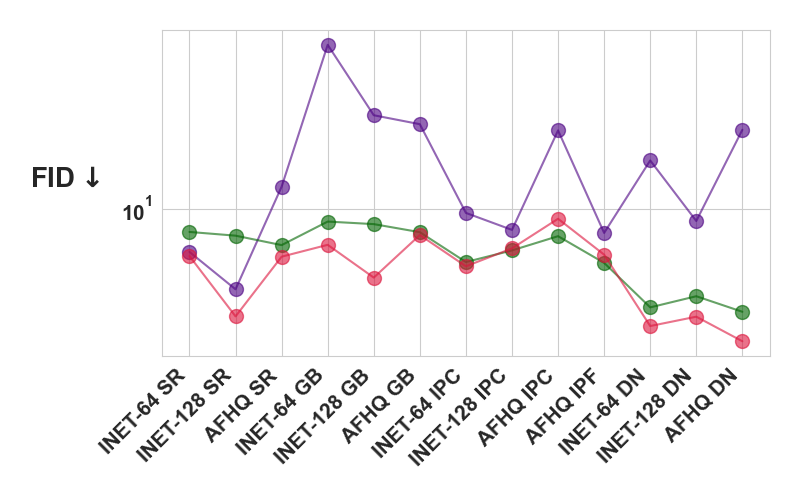}
        \includegraphics[width=0.49\textwidth]{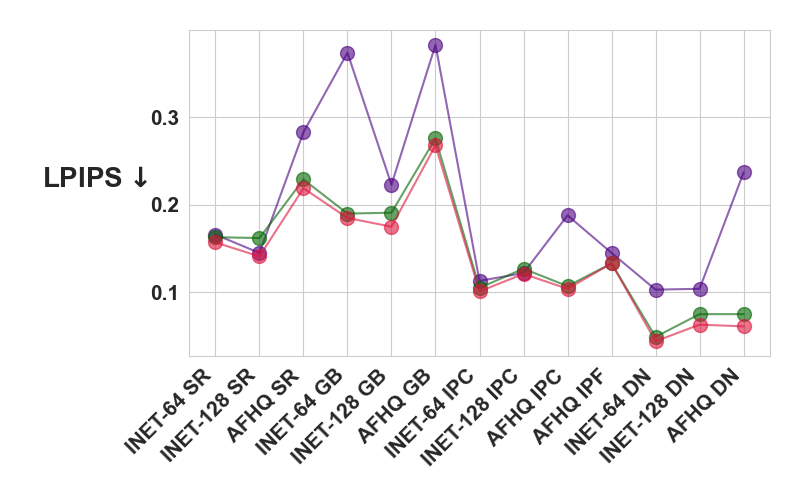}
        \includegraphics[width=0.49\textwidth]{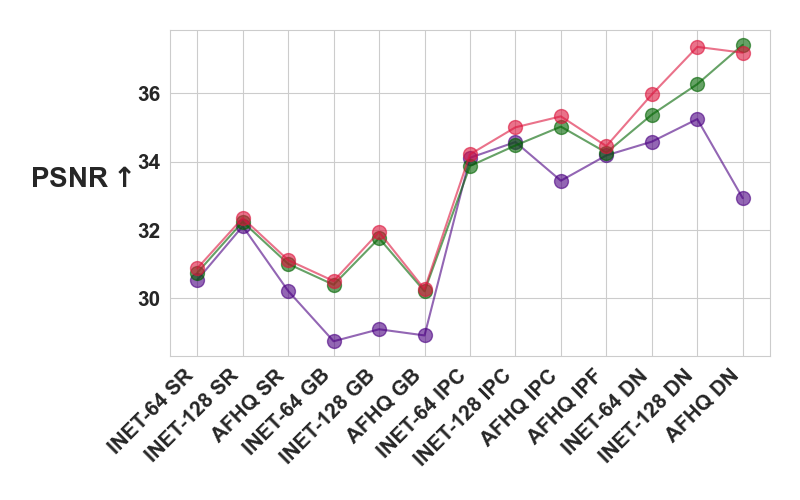}
        \includegraphics[width=0.49\textwidth]{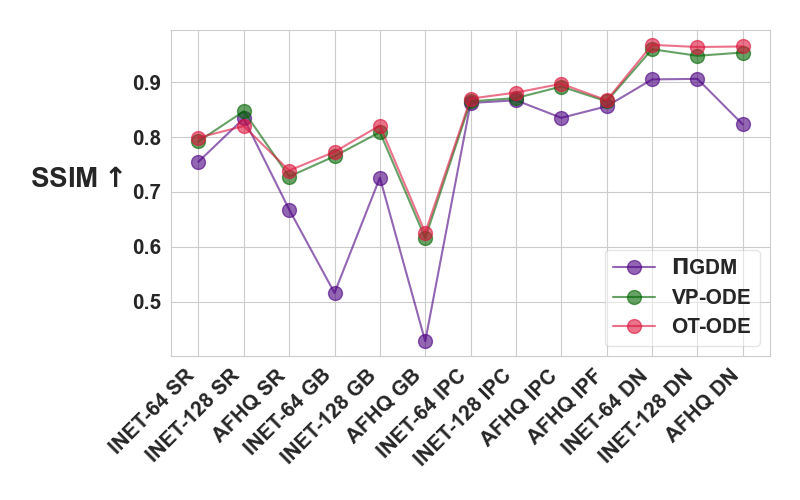}
    \end{minipage}
    \caption{Quantitative evaluation of pretrained conditional OT model for linear inverse problems on super-resolution (SR), gaussian deblurring (GB), image inpainting - centered mask (IPC) and denoising (DN) with $\sigma_y=0.05$. We show results on face-blurred ImageNet-64 (INET-64), face-blurred ImageNet-128 (INET-128), and AFHQ-256 (AFHQ).}
    \label{fig:sigma-0.05-cond-OT}
\end{figure}

The main text includes~\cref{fig:sigma-0.05-ddpm-ckpt} with $\sigma_y=0.05$ produced with the denoiser from the continuous-time VP-SDE diffusion model showing plots of various metrics across all datasets and tasks.  Here we provide the same for our pre-trained conditional OT flow matching model using Algorithm~\ref{alg:arb-flow-image-inversion} in~\cref{fig:sigma-0.05-cond-OT}, and noiseless figures for both models in~\cref{fig:sigma-0-ddpm-ckpt,fig:sigma-0-ot-ckpt}.  To save compute, for the flow model we only include our $\Pi$GDM baseline as RED-Diff required extensive hyperparameter tuning. The qualitative results using the flow model instead of diffusion model checkpoint are identical.  %

This section also includes tables containing the numerical values of metrics across all datasets and tasks.  The tables are hierarchically organized by noise, dataset, and task in consistent ordering.  Noisy results with $\sigma_y=0.05$ are in~\cref{tab:imagenet-64-sigma-0.05-sr-gb,tab:imagenet-64-sigma-0.05-ip-dn,tab:imagenet-128-sigma-0.05-sr-gb,tab:imagenet-128-sigma-0.05-ip-dn,tab:afhq-sigma-0.05-sr-gb,tab:afhq-sigma-0.05-ip-dn} and noiseless results with $\sigma_y=0$ are in~\cref{tab:imagenet-64-sigma-0-sr-gb,tab:imagenet-64-sigma-0-ip,tab:imagenet-128-sigma-0-sr-gb,tab:imagenet-128-sigma-0-ip,tab:afhq-sigma-0.05-sr-gb,tab:afhq-sigma-0.05-ip-dn}.

\begin{table*}[th!]
    \centering
    \caption{Quantitative evaluation of linear inverse problems on face-blurred ImageNet-$64\times64$}
    \label{tab:imagenet-64-sigma-0.05-sr-gb}
    \resizebox{\textwidth}{!}{
    \begin{tabular}{ccccccccccc}
    \toprule 
    \multirow{3}{*}{Model} & \multirow{3}{*}{Inference} & \multirow{3}{*}{NFEs $\downarrow$} &\multicolumn{4}{c}{SR 2$\times$, $\sigma_y=0.05$} &\multicolumn{4}{c}{Gaussian deblur, $\sigma_y=0.05$} \\
    \cmidrule(lr){4-7}\cmidrule(lr){8-11}
    & & & FID $\downarrow$ & LPIPS $\downarrow$ & PSNR $\uparrow$ &  SSIM $\uparrow$ & FID $\downarrow$ & LPIPS $\downarrow$ & PSNR $\uparrow$ &  SSIM $\uparrow$\\
    \midrule
    OT & OT-ODE & 80 &  \textbf{6.07} & \textbf{0.157} & \textbf{30.88} & \textbf{0.799} & \textbf{6.83} &  \textbf{0.185} & \textbf{30.51} & \textbf{0.773} \\
    OT & VP-ODE & 80 & 7.82 & 0.163 &  30.75 & 0.792 & 8.72 & 0.190 & 30.40 & 0.765 \\
    OT & $\Pi$GDM & 100 & 6.52 & 0.168 & 30.54 & 0.753 & 55.19 & 0.374 & 28.74 & 0.516 \\
    \midrule
    VP-SDE & OT-ODE & 80 & \textbf{5.57} & \textbf{0.155} & 30.88 & 0.799 & \textbf{6.33} & \textbf{0.181} & \textbf{30.52} & 0.773 \\
    VP-SDE & VP-ODE & 80 & 7.40 & 0.160 & 30.75 & 0.792 & 8.16 & 0.187 & 30.42 & 0.766 \\
    VP-SDE & $\Pi$GDM & 100 & 6.84 & 0.174 & 30.48 & 0.743 & 54.77 & 0.376 & 28.74 & 0.511 \\
    VP-SDE & RED-Diff & 1000 & 23.02 & 0.187 & \textbf{31.22} & \textbf{0.839} & 51.20 & 0.236 & 30.19 & \textbf{0.776} \\
    \bottomrule
    \end{tabular}
    }
\end{table*}

\begin{table*}[th!]
    \centering
    \caption{Quantitative evaluation of linear inverse problems on face-blurred ImageNet-$64\times64$}
    \label{tab:imagenet-64-sigma-0.05-ip-dn}
    \resizebox{\textwidth}{!}{
    \begin{tabular}{ccccccccccc}
    \toprule 
    \multirow{3}{*}{Model} & \multirow{3}{*}{Inference} & \multirow{3}{*}{NFEs $\downarrow$} &\multicolumn{4}{c}{Inpainting-\textit{Center}, $\sigma_y=0.05$} &\multicolumn{4}{c}{Denoising, $\sigma_y=0.05$} \\
    \cmidrule(lr){4-7}\cmidrule(lr){8-11}
    & & & FID $\downarrow$ & LPIPS $\downarrow$ & SSIM $\uparrow$ &  PSNR $\uparrow$ & FID $\downarrow$ & LPIPS $\downarrow$ & SSIM $\uparrow$ &  PSNR $\uparrow$\\
    \midrule
    OT & OT-ODE & 80 & \textbf{5.45} & \textbf{0.101} & \textbf{34.21} & \textbf{0.870} & \textbf{2.91} & \textbf{0.044} & \textbf{35.96} & \textbf{0.968} \\
    OT & VP-ODE & 80 & 5.70 & 0.105 & 33.87 & 0.865 & 3.54 & 0.049 & 35.37 & 0.960 \\
    OT & $\Pi$GDM & 100 &  9.25 & 0.111 & 34.13 & 0.863 & 16.59 & 0.102 & 34.60 & 0.906 \\
    \midrule
    VP-SDE & OT-ODE & 80 & \textbf{5.03} & \textbf{0.098} & \textbf{34.25} & \textbf{0.872} & \textbf{2.76} & \textbf{0.042} & \textbf{36.02} & \textbf{0.969} \\
    VP-SDE & VP-ODE & 80 &  5.26 & 0.103 & 33.93 & 0.866 & 3.29 & 0.048 & 35.45 & 0.961 \\
    VP-SDE & $\Pi$GDM & 100 &  9.75 & 0.113 & 34.03 & 0.860 & 17.19 & 0.107 & 34.25 & 0.901 \\
    VP-SDE & RED-Diff & 1000 & 12.18 & 0.119 & 33.97 & 0.881 & 6.02 & 0.041 & 35.64 & 0.964 \\
    \bottomrule
    \end{tabular}
    }
\end{table*}

\begin{table*}[th!]
    \centering
    \caption{Quantitative evaluation of linear inverse problems on face-blurred  ImageNet-$128\times128$}
    \label{tab:imagenet-128-sigma-0.05-sr-gb}
    \resizebox{\textwidth}{!}{
    \begin{tabular}{ccccccccccc}
    \toprule 
    \multirow{3}{*}{Model} & \multirow{3}{*}{Inference} & \multirow{3}{*}{NFEs $\downarrow$} &\multicolumn{4}{c}{SR 2$\times$, $\sigma_y=0.05$} &\multicolumn{4}{c}{Gaussian deblur, $\sigma_y=0.05$} \\
    \cmidrule(lr){4-7}\cmidrule(lr){8-11}
    & & & FID $\downarrow$ & LPIPS $\downarrow$ & PSNR $\uparrow$ &  SSIM $\uparrow$ & FID $\downarrow$ & LPIPS $\downarrow$ & PSNR $\uparrow$ &  SSIM $\uparrow$\\
    \midrule
    OT & OT-ODE & 70 &  \textbf{3.22} & \textbf{0.141} & \textbf{32.35} & 0.820 & \textbf{4.84} & \textbf{0.175} & \textbf{31.94} & \textbf{0.821} \\
    OT & VP-ODE & 70 & 7.52 & 0.162 & 32.24 & \textbf{0.847} & 8.49 & 0.191 & 31.76 & 0.809 \\
    OT & $\Pi$GDM & 100 &  4.38 & 0.148 & 32.07 & 0.831 & 30.30 & 0.328 & 29.96 & 0.606 \\
    \midrule
    VP-SDE & OT-ODE & 70 & \textbf{3.21} & \textbf{0.139} & \textbf{32.40} & \textbf{0.855} & \textbf{4.49} & \textbf{0.173} & \textbf{32.02} & \textbf{0.824} \\
    VP-SDE & VP-ODE & 70 & 9.14 & 0.166 & 32.06 & 0.838 & 9.35 & 0.193 & 31.66 & 0.804 \\
    VP-SDE & $\Pi$GDM & 100 & 7.55 & 0.183 & 31.61 & 0.785 & 55.61 & 0.463 & 28.57 & 0.414 \\
    VP-SDE & RED-Diff & 1000 & 10.54 & 0.182 & 31.82 & 0.852 & 21.43 & 0.229 & 31.41 & 0.807 \\
    \bottomrule
    \end{tabular}
    }
\end{table*}

\begin{table*}[th!]
    \centering
    \caption{Quantitative evaluation of linear inverse problems on face-blurred ImageNet-$128\times128$}
    \label{tab:imagenet-128-sigma-0.05-ip-dn}
    \resizebox{\textwidth}{!}{
    \begin{tabular}{ccccccccccc}
    \toprule 
    \multirow{3}{*}{Model} & \multirow{3}{*}{Inference} & \multirow{3}{*}{NFEs $\downarrow$} &\multicolumn{4}{c}{Inpainting-\textit{Center}, $\sigma_y=0.05$} &\multicolumn{4}{c}{Denoising, $\sigma_y=0.05$} \\
    \cmidrule(lr){4-7}\cmidrule(lr){8-11}
    & & & FID $\downarrow$ & LPIPS $\downarrow$ & PSNR $\uparrow$ &  SSIM $\uparrow$ & FID $\downarrow$ & LPIPS $\downarrow$ & PSNR $\uparrow$ &  SSIM $\uparrow$\\
    \midrule
    OT & OT-ODE & 70 & 6.58 & \textbf{0.121} & \textbf{35.00} & \textbf{0.881} & \textbf{3.21} & \textbf{0.063} & \textbf{37.35} & \textbf{0.964} \\
    OT & VP-ODE & 70 & \textbf{6.44} & 0.127 & 34.47 & 0.871 & 3.98 & 0.075 & 36.26 & 0.948 \\
    OT & $\Pi$GDM & 100 & 7.99 & 0.122 & 34.57 & 0.867 & 9.60 & 0.107 & 35.11 & 0.903 \\
    \midrule
    VP-SDE & OT-ODE & 70 & \textbf{6.39} & \textbf{0.120} & \textbf{35.04} & \textbf{0.882} & \textbf{3.25} & \textbf{0.062} & \textbf{37.41} & \textbf{0.965} \\
    VP-SDE & VP-ODE & 70 & 8.47 & 0.129 & 34.43 & 0.876 & 5.83 & 0.087 & 35.85 & 0.938 \\
    VP-SDE & $\Pi$GDM & 100 &  9.75 & 0.130 & 34.45 & 0.858 & 10.69 & 0.124 & 34.72 & 0.882 \\
    VP-SDE & RED-Diff & 1000 & 14.63 & 0.171 & 32.42 & 0.820 & 9.19 & 0.105 & 33.52 & 0.895 \\
    \bottomrule
    \end{tabular}
    }
\end{table*}

\begin{table*}[th!]
    \centering
    \caption{Quantitative evaluation of linear inverse problems on AFHQ-$256\times256$}
    \label{tab:afhq-sigma-0.05-sr-gb}
    \resizebox{\textwidth}{!}{
    \begin{tabular}{ccccccccccc}
    \toprule 
    \multirow{3}{*}{Model} & \multirow{3}{*}{Inference} & \multirow{3}{*}{NFEs $\downarrow$} &\multicolumn{4}{c}{SR 4$\times$, $\sigma_y=0.05$} &\multicolumn{4}{c}{Gaussian deblur, $\sigma_y=0.05$} \\
    \cmidrule(lr){4-7}\cmidrule(lr){8-11}
    & & & FID $\downarrow$ & LPIPS $\downarrow$ & PSNR $\uparrow$ &  SSIM $\uparrow$ & FID $\downarrow$ & LPIPS $\downarrow$ & PSNR $\uparrow$ &  SSIM $\uparrow$\\
    \midrule
    OT & OT-ODE & 100 & \textbf{6.03} & \textbf{0.219} & \textbf{31.12} & \textbf{0.739} & \textbf{7.57} & \textbf{0.268} & \textbf{30.27} & \textbf{0.626} \\
    OT & VP-ODE & 100 & 6.81 & 0.229 & 31.01 & 0.728 & 7.80 & 0.276 & 30.21 & 0.616 \\
    OT & $\Pi$GDM & 100 & 12.69 & 0.285 & 30.18 & 0.665 & 24.60 & 0.383 & 28.93 & 0.429 \\
    \midrule
    VP-SDE & OT-ODE & 100 & \textbf{7.28} & \textbf{0.238} & 30.83 & 0.714 & \textbf{8.53} & \textbf{0.276} & \textbf{30.37} & 0.641 \\
    VP-SDE & VP-ODE & 100 & 8.02 & 0.243 & \textbf{30.96} & \textbf{0.727} & 10.21 & 0.289 & 30.21 & 0.621 \\
    VP-SDE & $\Pi$GDM & 100 & 77.49 & 0.469 & 29.34 & 0.469 & 116.42 & 0.535 & 28.49 & 0.313 \\
    VP-SDE & RED-Diff & 1000 & 20.84 & 0.331 & 29.97 & 0.675 & 15.81 & 0.341 & 30.15 & \textbf{0.645} \\
    \bottomrule
    \end{tabular}
    }
\end{table*}

\begin{table*}[th!]
    \centering
    \caption{Quantitative evaluation of linear inverse problems on AFHQ-$256\times256$}
    \label{tab:afhq-sigma-0.05-ip-dn}
    \resizebox{\textwidth}{!}{
    \begin{tabular}{ccccccccccc}
    \toprule 
    \multirow{3}{*}{Model} & \multirow{3}{*}{Inference} & \multirow{3}{*}{NFEs $\downarrow$} &\multicolumn{4}{c}{Inpainting-\textit{Center}, $\sigma_y=0.05$} &\multicolumn{4}{c}{Denoising, $\sigma_y=0.05$} \\
    \cmidrule(lr){4-7}\cmidrule(lr){8-11}
    & & & FID $\downarrow$ & LPIPS $\downarrow$ & PSNR $\uparrow$ &  SSIM $\uparrow$ & FID $\downarrow$ & LPIPS $\downarrow$ & PSNR $\uparrow$ &  SSIM $\uparrow$\\
    \midrule
    OT & OT-ODE & 100 & 8.98 & \textbf{0.104} & \textbf{35.32} & \textbf{0.897} & \textbf{2.48} & \textbf{0.061} & 37.18 & \textbf{0.965} \\
    OT & VP-ODE & 100 & \textbf{7.48} & 0.107 & 35.02 & 0.892 & 3.38 & 0.075 & \textbf{37.41} & 0.954 \\
    OT & $\Pi$GDM & 100 & 19.09 & 0.153 & 34.20 & 0.855 & 22.87 & 0.237 & 32.93 & 0.823 \\
    \midrule
    VP-SDE & OT-ODE & 100 & 9.93 & \textbf{0.107} & \textbf{35.18} & \textbf{0.892} & \textbf{2.17} & \textbf{0.060} & \textbf{37.95} & \textbf{0.963} \\
    VP-SDE & VP-ODE & 100 & \textbf{8.78} & \textbf{0.107} & 35.12 & 0.891 & 3.08 & 0.071 & 37.68 & 0.959 \\
    VP-SDE & $\Pi$GDM & 100 & 57.46 & 0.239 & 32.40 & 0.773 & 81.15 & 0.451 & 29.62 & 0.639 \\
    VP-SDE & RED-Diff & 1000 & 11.02 & 0.124 & 34.97 & 0.893 & 4.93 & 0.112 & 34.18 & 0.899 \\
    \bottomrule
    \end{tabular}
    }
\end{table*}

\begin{table*}[th!]
    \centering
    \caption{Quantitative evaluation of linear inverse problems on face-blurred ImageNet-$64\times64$}
    \label{tab:imagenet-64-sigma-0-sr-gb}
    \resizebox{\textwidth}{!}{
    \begin{tabular}{ccccccccccc}
    \toprule 
    \multirow{3}{*}{Model} & \multirow{3}{*}{Inference} & \multirow{3}{*}{NFEs $\downarrow$} &\multicolumn{4}{c}{SR 2$\times$, $\sigma_y=0$} &\multicolumn{4}{c}{Gaussian deblur, $\sigma_y=0$} \\
    \cmidrule(lr){4-7}\cmidrule(lr){8-11}
    & & & FID $\downarrow$ & LPIPS $\downarrow$ & PSNR $\uparrow$ &  SSIM $\uparrow$ & FID $\downarrow$ & LPIPS $\downarrow$ & PSNR $\uparrow$ &  SSIM $\uparrow$\\
    \midrule
    OT & OT-ODE & 80 &  \textbf{6.46} & 0.119 & 31.59 & 0.839 & \textbf{2.59} & \textbf{0.038} & 35.31 & 0.961 \\
    OT & VP-ODE & 80 & 8.29 & 0.147 & 31.20 & 0.817 & 6.13 & 0.083 & 33.31 & 0.929 \\
    OT & $\Pi$GDM & 100 & 6.89 & \textbf{0.115} & \textbf{32.02} & \textbf{0.853} & 4.53 & 0.051 & \textbf{35.88} & \textbf{0.963} \\
    \midrule
    VP-SDE & OT-ODE  & 80 & \textbf{6.32} & 0.118 & 31.60 & 0.839 & \textbf{2.61} & \textbf{0.037} & 35.45 & 0.963 \\
    VP-SDE & VP-ODE  & 80 &  7.76 & 0.145 & 31.21 & 0.817 & 5.68 & 0.080 & 33.37 & 0.931 \\
    VP-SDE & $\Pi$GDM & 100 & 6.47 & \textbf{0.113} & \textbf{32.03} & \textbf{0.853} & 4.35 & 0.049 & \textbf{35.95} & \textbf{0.964} \\
    VP-SDE & RED-Diff & 1000 & 11.74 & 0.224 & 30.12 & 0.798 & 15.39 & 0.134 & 31.99 & 0.879\\
    \bottomrule
    \end{tabular}
    }
\end{table*}

\begin{table*}[th!]
    \centering
    \caption{Quantitative evaluation of linear inverse problems on face-blurred ImageNet-$64\times64$}
    \label{tab:imagenet-64-sigma-0-ip}
    \resizebox{0.65\textwidth}{!}{
    \begin{tabular}{ccccccc}
    \toprule 
    \multirow{3}{*}{Model} & \multirow{3}{*}{Inference} & \multirow{3}{*}{NFEs $\downarrow$} &\multicolumn{4}{c}{Inpainting-\textit{Center}, $\sigma_y=0$} \\
    \cmidrule(lr){4-7}
    & & & FID $\downarrow$ & LPIPS $\downarrow$ & PSNR $\uparrow$ &  SSIM $\uparrow$ \\
    \midrule
    OT & OT-ODE  & 80 & \textbf{4.94} & \textbf{0.080} & \textbf{37.42} & 0.885 \\
    OT & VP-ODE & 80 & 7.85 & 0.120 & 34.24 & 0.858 \\
    OT & $\Pi$GDM & 100 & 6.09 & 0.082 & 36.75 & \textbf{0.901} \\
    \midrule
    VP-SDE & OT-ODE  & 80 & \textbf{4.85} & \textbf{0.079} & 37.64 & 0.887 \\
    VP-SDE & VP-ODE & 80 & 7.21 & 0.117 & 34.33 & 0.860 \\
    VP-SDE & $\Pi$GDM & 100 & 5.79 & 0.081 & 36.81 & 0.902 \\
    VP-SDE & RED-Diff & 1000 & 7.29 & \textbf{0.079} & \textbf{39.14} & \textbf{0.925} \\
    \bottomrule
    \end{tabular}
    }
\end{table*}

\begin{table*}[th!]
    \centering
    \caption{Quantitative evaluation of linear inverse problems on face-blurred ImageNet-$128\times128$}
    \label{tab:imagenet-128-sigma-0-sr-gb}
    \resizebox{0.65\textwidth}{!}{
    \begin{tabular}{ccccccc}
    \toprule 
    \multirow{3}{*}{Model} & \multirow{3}{*}{Inference} & \multirow{3}{*}{NFEs $\downarrow$} &\multicolumn{4}{c}{Inpainting-\textit{Center}, $\sigma_y=0$} \\
    \cmidrule(lr){4-7}
    & & & FID $\downarrow$ & LPIPS $\downarrow$ & PSNR $\uparrow$ &  SSIM $\uparrow$ \\
    \midrule
    OT & OT-ODE & 70 & 5.88 & \textbf{0.095} & \textbf{37.06} & 0.894 \\
    OT & VP-ODE & 70 & 8.63 & 0.144 & 34.48 & 0.864 \\
    OT & $\Pi$GDM & 100 & \textbf{5.82} & 0.097 & 36.89 & \textbf{0.908} \\
    \midrule
    VP-SDE & OT-ODE & 70 & 5.93 & 0.094 & 37.31 & 0.898  \\
    VP-SDE & VP-ODE & 70 & 8.08 & 0.142 & 34.55 & 0.865 \\
    VP-SDE & $\Pi$GDM & 100 & 5.74 & 0.095 & 37.01 & 0.911 \\
    VP-SDE & RED-Diff & 1000 & \textbf{5.40} & \textbf{0.068} & \textbf{38.91} & \textbf{0.928} \\
    \bottomrule
    \end{tabular}
    }
\end{table*}

\begin{table*}[thbp!]
    \centering
    \caption{Quantitative evaluation of linear inverse problems on face-blurred ImageNet-$128\times128$}
    \label{tab:imagenet-128-sigma-0-ip}
    \resizebox{\textwidth}{!}{
    \begin{tabular}{ccccccccccc}
    \toprule 
    \multirow{3}{*}{Model} & \multirow{3}{*}{Inference} & \multirow{3}{*}{NFEs $\downarrow$} &\multicolumn{4}{c}{SR 2$\times$, $\sigma_y=0$} &\multicolumn{4}{c}{Gaussian deblur, $\sigma_y=0$} \\
    \cmidrule(lr){4-7}\cmidrule(lr){8-11}
    & & & FID $\downarrow$ & LPIPS $\downarrow$ & PSNR $\uparrow$ &  SSIM $\uparrow$ & FID $\downarrow$ & LPIPS $\downarrow$ & PSNR $\uparrow$ &  SSIM $\uparrow$\\
    \midrule
    OT & OT-ODE & 70 & \textbf{4.46} & \textbf{0.097} & 33.88 & 0.903 & \textbf{2.09} & \textbf{0.048} & 37.49 &\textbf{0.961} \\
    OT & VP-ODE & 70 & 7.69 & 0.144 & 32.93 & 0.871 & 6.02 & 0.108 & 34.73 & 0.925 \\
    OT & $\Pi$GDM & 100 & 6.09 & 0.105 & \textbf{34.28} & \textbf{0.910} & 4.28 & 0.066 & \textbf{37.56} & \textbf{0.961} \\
    \midrule
    VP-SDE & OT-ODE & 70 & 4.62 & 0.096 & 33.95 & 0.906 & \textbf{2.26} & \textbf{0.046} & \textbf{37.79} & \textbf{0.967} \\
    VP-SDE & VP-ODE & 70 & 7.91 & 0.144 & 32.87 & 0.869 & 5.64 & 0.105 & 34.81 & 0.928 \\
    VP-SDE & $\Pi$GDM & 100 & 6.02 & 0.104 & 34.33 & 0.911 & 4.35 & 0.065 & 37.70 & 0.963 \\
    VP-SDE & RED-Diff & 1000 &  \textbf{3.90} & \textbf{0.082} & \textbf{34.47} & \textbf{0.92} & 4.19 & 0.085 & 34.68 & 0.929 \\
    \bottomrule
    \end{tabular}
    }
\end{table*}

\begin{table*}[thbp!]
    \centering
    \caption{Quantitative evaluation of linear inverse problems on AFHQ-$256\times256$}
    \label{tab:afhq-sigma-0-gb-dn}
    \resizebox{\textwidth}{!}{
    \begin{tabular}{ccccccccccc}
    \toprule 
    \multirow{3}{*}{Model} & \multirow{3}{*}{Inference} & \multirow{3}{*}{NFEs $\downarrow$} &\multicolumn{4}{c}{SR 4$\times$, $\sigma_y=0$} &\multicolumn{4}{c}{Gaussian deblur, $\sigma_y=0$} \\
    \cmidrule(lr){4-7}\cmidrule(lr){8-11}
    & & & FID $\downarrow$ & LPIPS $\downarrow$ & PSNR $\uparrow$ &  SSIM $\uparrow$ & FID $\downarrow$ & LPIPS $\downarrow$ & PSNR $\uparrow$ &  SSIM $\uparrow$\\
    \midrule
    OT & OT-ODE & 100 & \textbf{5.75} & \textbf{0.169} & 32.25 & 0.792 & \textbf{6.63} & \textbf{0.213} & 31.29 & 0.722 \\
    OT & VP-ODE & 100 & 6.14 & 0.194 & 31.93 & 0.773 & 7.38 & 0.231 & 31.10 & 0.705 \\
    OT & $\Pi$GDM & 100 & 8.89 & 0.173 & \textbf{32.57} & \textbf{0.812} & 9.78 & 0.209 & \textbf{31.54} & \textbf{0.743} \\
    \midrule
    VP-SDE & OT-ODE & 100 & \textbf{6.58} & \textbf{0.178} & 32.18 & 0.789 & \textbf{8.24} & \textbf{0.226} & 31.21 & 0.717 \\
    VP-SDE & VP-ODE & 100 & 8.00 & 0.225 & 31.48 & 0.742 & 9.19 & 0.252 & 30.91 & 0.688 \\
    VP-SDE & $\Pi$GDM & 100 & 10.85 & 0.189 & \textbf{32.52} & \textbf{0.811} & 11.46 & 0.228 & \textbf{31.47} & \textbf{0.738} \\
    VP-SDE & RED-Diff & 1000 & 8.65 & 0.191 & 32.21 & 0.801 & 11.67 & 0.268 & 31.30 & 0.731 \\
    \bottomrule
    \end{tabular}
    }
\end{table*}

\begin{table*}[thbp!]
    \centering
    \caption{Quantitative evaluation of linear inverse problems on AFHQ-$256\times256$}
    \label{tab:afhq-sigma-0-ip}
    \resizebox{\textwidth}{!}{
    \begin{tabular}{ccccccccccc}
    \toprule 
    \multirow{3}{*}{Model} & \multirow{3}{*}{Inference} & \multirow{3}{*}{NFEs $\downarrow$} &\multicolumn{4}{c}{Inpainting-\textit{Center}, $\sigma_y=0$} &\multicolumn{4}{c}{Inpainting-\textit{Free-form}, $\sigma_y=0$} \\
    \cmidrule(lr){4-7}\cmidrule(lr){8-11}
    & & & FID $\downarrow$ & LPIPS $\downarrow$ & PSNR $\uparrow$ &  SSIM $\uparrow$ & FID $\downarrow$ & LPIPS $\downarrow$ & PSNR $\uparrow$ &  SSIM $\uparrow$\\
    \midrule
    OT & OT-ODE & 100 & 8.87 & \textbf{0.061} & \textbf{37.45} & 0.921 & \textbf{4.98} & \textbf{0.097} & 36.15 & 0.889 \\
    OT & VP-ODE & 100 & 9.18 & 0.106 & 35.63 & 0.898 & 6.92 & 0.135 & 34.72 & 0.869 \\
    OT & $\Pi$GDM & 100 & \textbf{7.36} & 0.080 & \textbf{37.45} & \textbf{0.933} & 6.52 & 0.100 & \textbf{36.58} & \textbf{0.913} \\
    \midrule
    VP-SDE & OT-ODE & 100 & 9.95 & 0.064 & 37.49 & 0.918 & \textbf{5.39} & 0.099 & 36.15 & 0.887 \\
    VP-SDE & VP-ODE & 100 & 10.50 & 0.112 & 35.59 & 0.893 & 7.36 & 0.139 & 34.65 & 0.865 \\
    VP-SDE & $\Pi$GDM & 100 & 8.61 & 0.088 & 37.27 & 0.925 & 7.25 & 0.109 & 36.37 & \textbf{0.906} \\
    VP-SDE & RED-Diff & 1000 & \textbf{8.53} & \textbf{0.050} & \textbf{38.89} & \textbf{0.951} & 7.27 & \textbf{0.090} & \textbf{36.88} & 0.892 \\
    \bottomrule
    \end{tabular}
    }
\end{table*}

%% file: sections/additional_qualitative_results.tex
\clearpage
\subsection{Additional qualitative results}
\begin{figure}[!htb]
    \begin{subfigure}[t]{0.19\textwidth}
        \includegraphics[width=\textwidth]{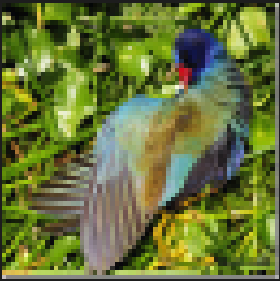}
    \end{subfigure}
    \begin{subfigure}[t]{0.19\textwidth}
        \includegraphics[width=\textwidth]{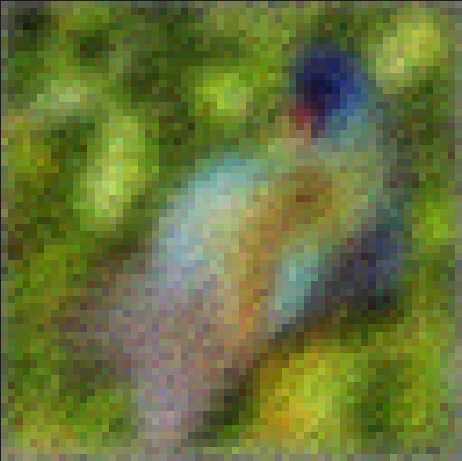}
    \end{subfigure}
    \begin{subfigure}[t]{0.19\textwidth}
        \includegraphics[width=\textwidth]{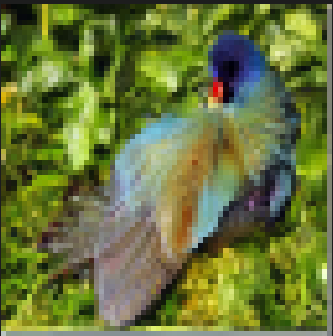}
    \end{subfigure}
    \begin{subfigure}[t]{0.19\textwidth}
        \includegraphics[width=\textwidth]{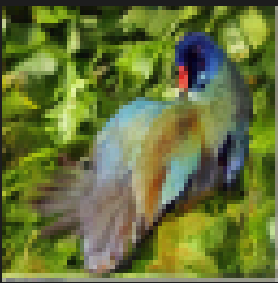}
    \end{subfigure}
    \begin{subfigure}[t]{0.19\textwidth}    
        \includegraphics[width=\textwidth]{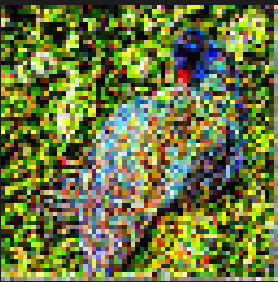}
    \end{subfigure} \\
    \begin{subfigure}[t]{0.19\textwidth}
        \includegraphics[width=\textwidth]{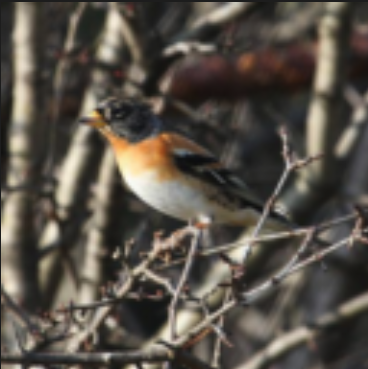}
    \end{subfigure}
    \begin{subfigure}[t]{0.19\textwidth}
        \includegraphics[width=\textwidth]{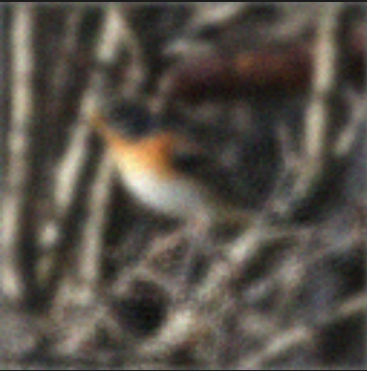}
    \end{subfigure}
    \begin{subfigure}[t]{0.19\textwidth}
        \includegraphics[width=\textwidth]{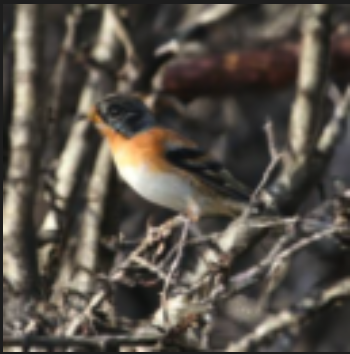}
    \end{subfigure}
    \begin{subfigure}[t]{0.19\textwidth}
        \includegraphics[width=\textwidth]{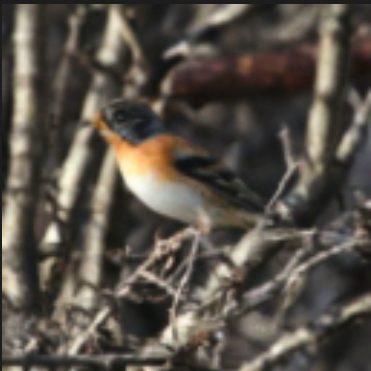}
    \end{subfigure}
    \begin{subfigure}[t]{0.19\textwidth}    
        \includegraphics[width=\textwidth]{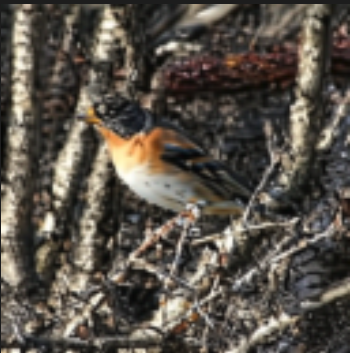}
    \end{subfigure}\\
        \begin{subfigure}[t]{0.19\textwidth}
        \includegraphics[width=\textwidth]{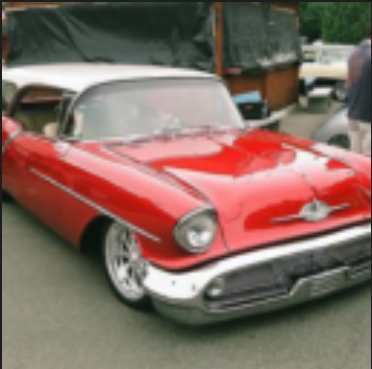}
    \end{subfigure}
    \begin{subfigure}[t]{0.19\textwidth}
        \includegraphics[width=\textwidth]{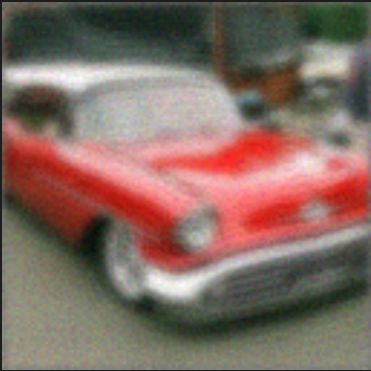}
    \end{subfigure}
    \begin{subfigure}[t]{0.19\textwidth}
        \includegraphics[width=\textwidth]{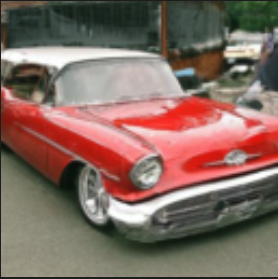}
    \end{subfigure}
    \begin{subfigure}[t]{0.19\textwidth}
        \includegraphics[width=\textwidth]{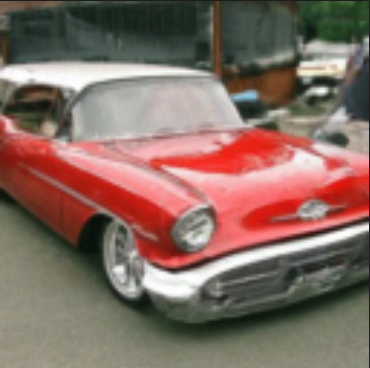}
    \end{subfigure}
    \begin{subfigure}[t]{0.19\textwidth}    
        \includegraphics[width=\textwidth]{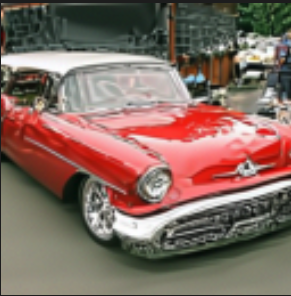}
    \end{subfigure}\\
        \begin{subfigure}[t]{0.19\textwidth}
        \includegraphics[width=\textwidth]{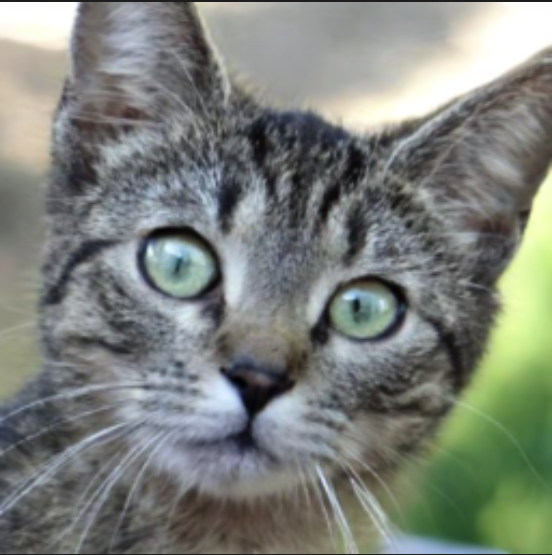}
    \end{subfigure}
    \begin{subfigure}[t]{0.19\textwidth}
        \includegraphics[width=\textwidth]{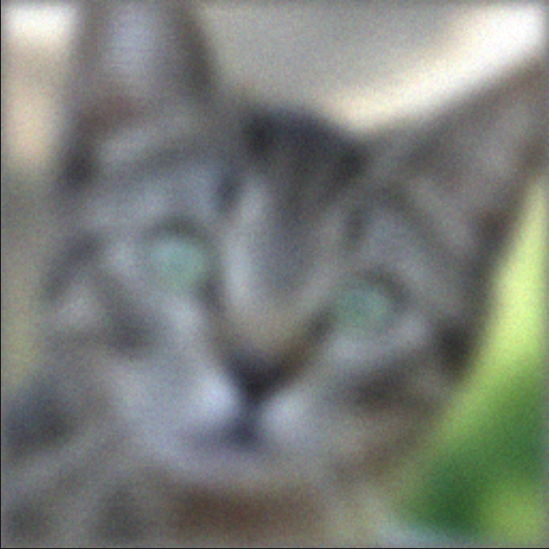}
    \end{subfigure}
    \begin{subfigure}[t]{0.19\textwidth}
        \includegraphics[width=\textwidth]{images/gaussian_blur/afhq/196_6_ot_ot.png}
    \end{subfigure}
    \begin{subfigure}[t]{0.19\textwidth}
        \includegraphics[width=\textwidth]{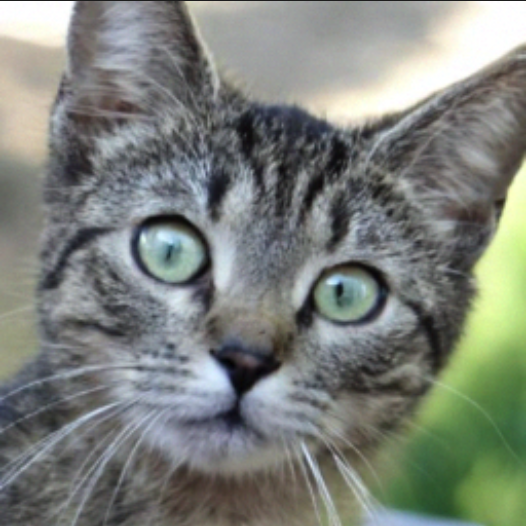}
    \end{subfigure}
    \begin{subfigure}[t]{0.19\textwidth}    
        \includegraphics[width=\textwidth]{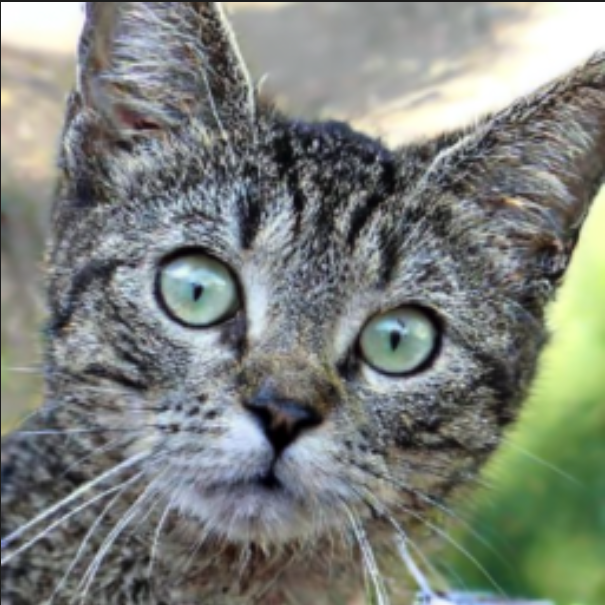}
    \end{subfigure}\\
    \begin{subfigure}[t]{0.19\textwidth}
        \includegraphics[width=\textwidth]{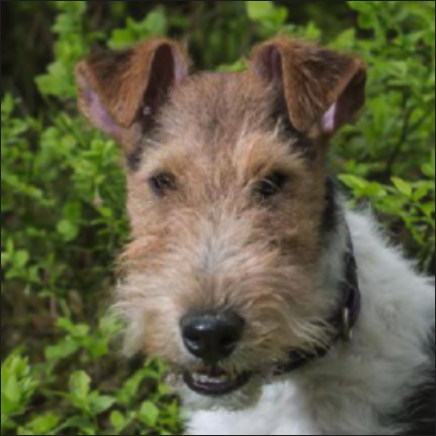}
        \caption{Reference}
    \end{subfigure}
    \begin{subfigure}[t]{0.19\textwidth}
        \includegraphics[width=\textwidth]{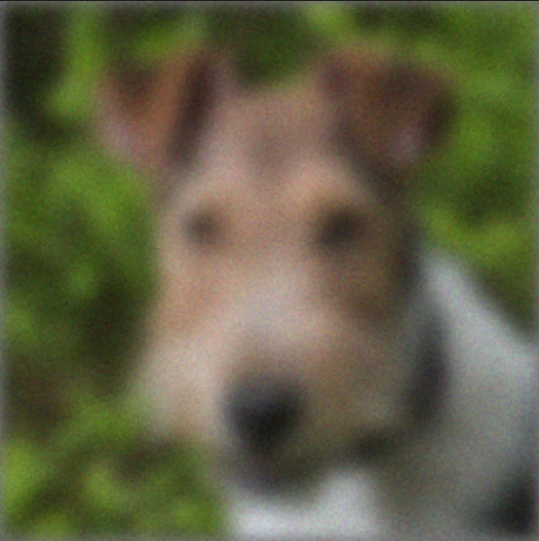}
        \caption{Distorted}
    \end{subfigure}
    \begin{subfigure}[t]{0.19\textwidth}
        \includegraphics[width=\textwidth]{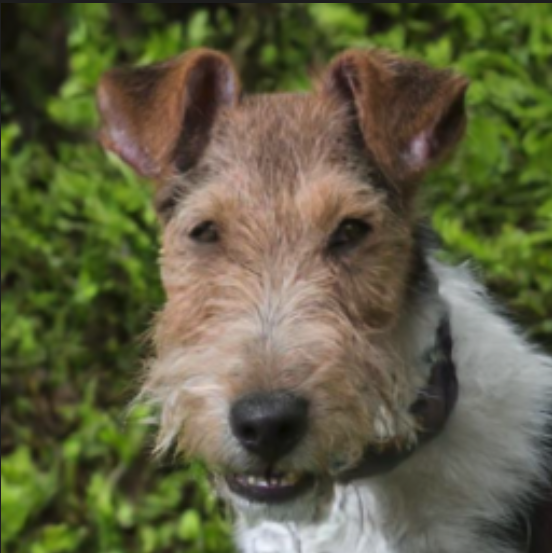}
        \caption{OT-ODE}
    \end{subfigure}
    \begin{subfigure}[t]{0.19\textwidth}
        \includegraphics[width=\textwidth]{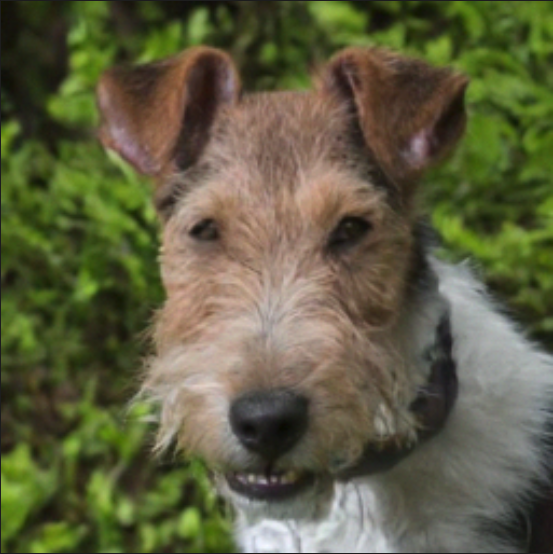}
        \caption{VP-ODE}
    \end{subfigure}
    \begin{subfigure}[t]{0.19\textwidth}    
        \includegraphics[width=\textwidth]{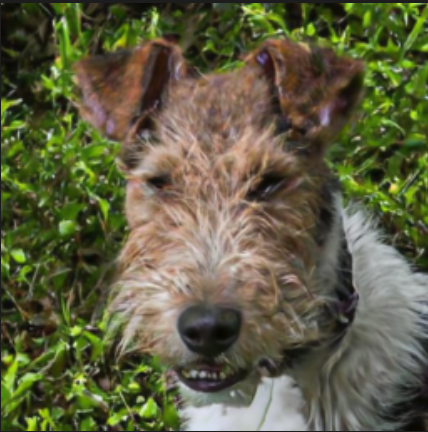}
        \caption{$\Pi$GDM}
    \end{subfigure}
    \caption{Gaussian-deblur with conditional OT model and $\sigma_y=0.05$ for (\textbf{first row}) face-blurred ImageNet-64, (\textbf{second and third row}) face-blurred ImageNet-128, and (\textbf{ fourth and fifth row}) AFHQ.}
    \label{fig:gb-noisy-ot-vis}
\end{figure}

\begin{figure}[!htb]
    \begin{subfigure}[t]{0.19\textwidth}
        \includegraphics[width=\textwidth]{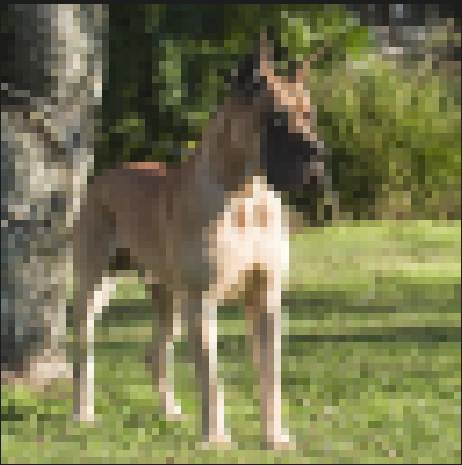}
    \end{subfigure}
    \begin{subfigure}[t]{0.19\textwidth}
        \includegraphics[width=\textwidth]{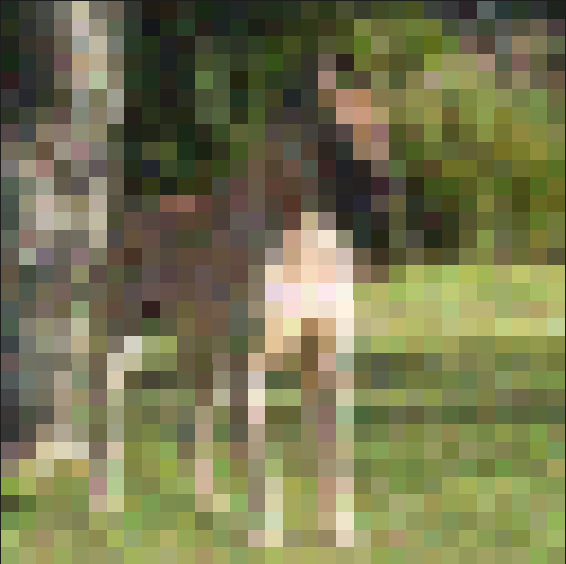}
    \end{subfigure}
    \begin{subfigure}[t]{0.19\textwidth}
        \includegraphics[width=\textwidth]{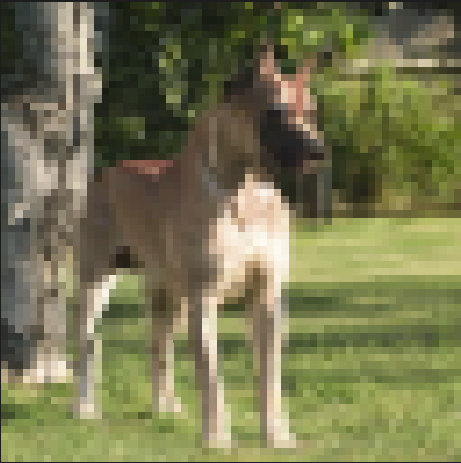}
    \end{subfigure}
    \begin{subfigure}[t]{0.19\textwidth}
        \includegraphics[width=\textwidth]{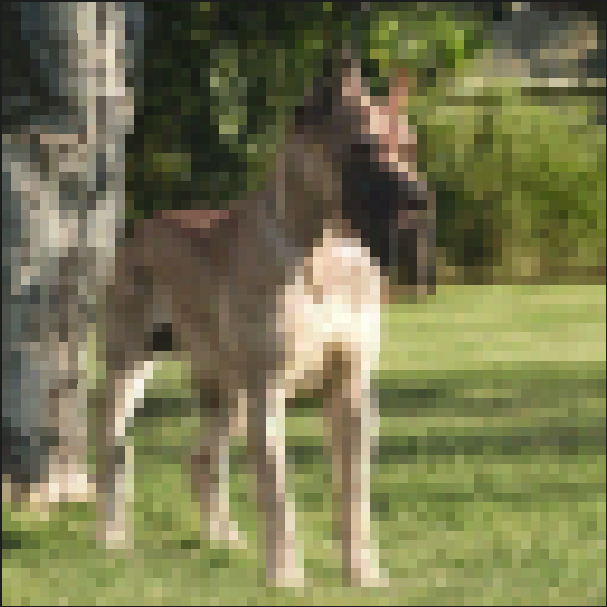}
    \end{subfigure}
    \begin{subfigure}[t]{0.19\textwidth}    
        \includegraphics[width=\textwidth]{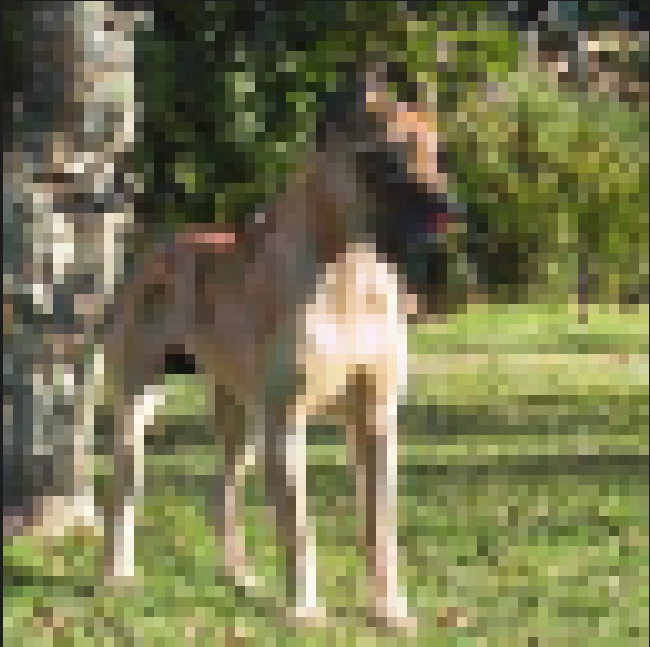}
    \end{subfigure} \\
        \begin{subfigure}[t]{0.19\textwidth}
        \includegraphics[width=\textwidth]{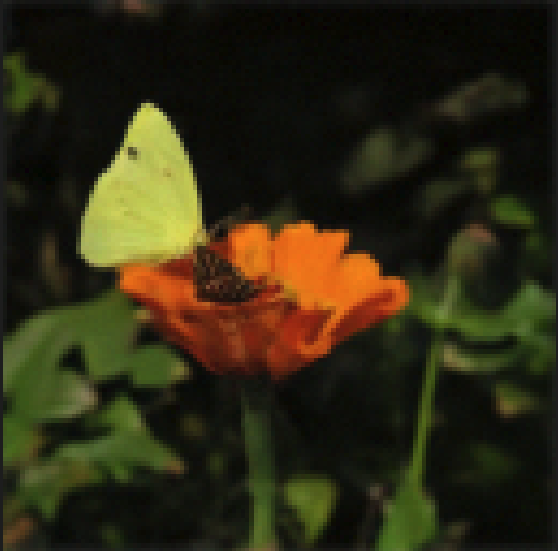}
    \end{subfigure}
    \begin{subfigure}[t]{0.19\textwidth}
        \includegraphics[width=\textwidth]{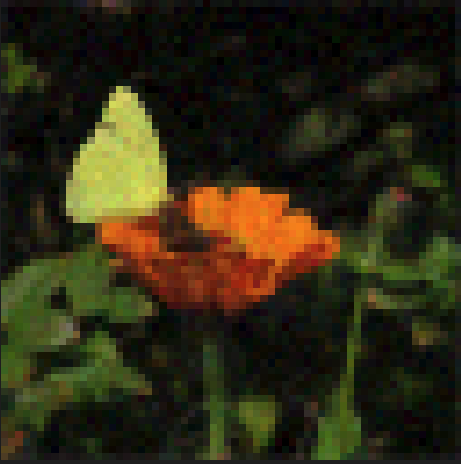}
    \end{subfigure}
    \begin{subfigure}[t]{0.19\textwidth}
        \includegraphics[width=\textwidth]{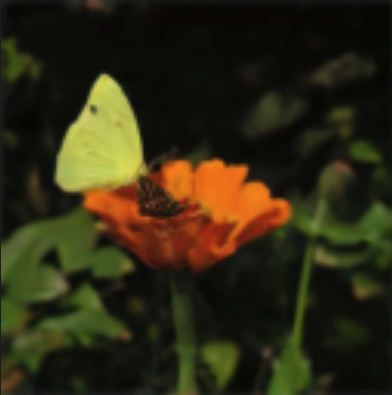}
    \end{subfigure}
    \begin{subfigure}[t]{0.19\textwidth}
        \includegraphics[width=\textwidth]{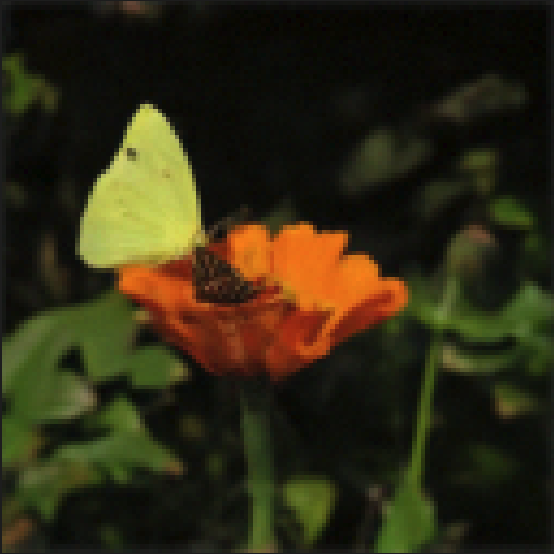}
    \end{subfigure}
    \begin{subfigure}[t]{0.19\textwidth}    
        \includegraphics[width=\textwidth]{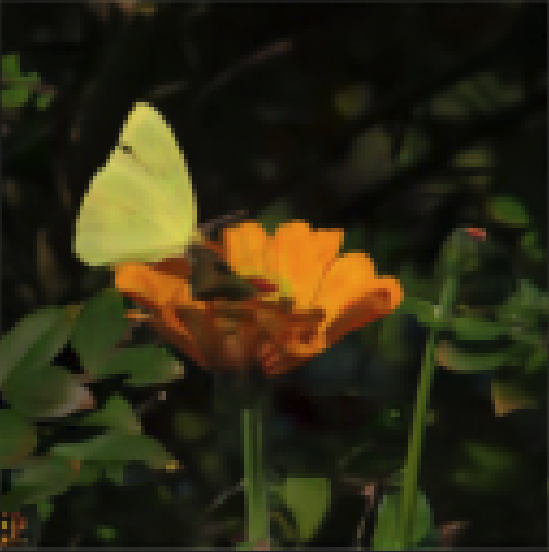}
    \end{subfigure}\\
    \begin{subfigure}[t]{0.19\textwidth}
        \includegraphics[width=\textwidth]{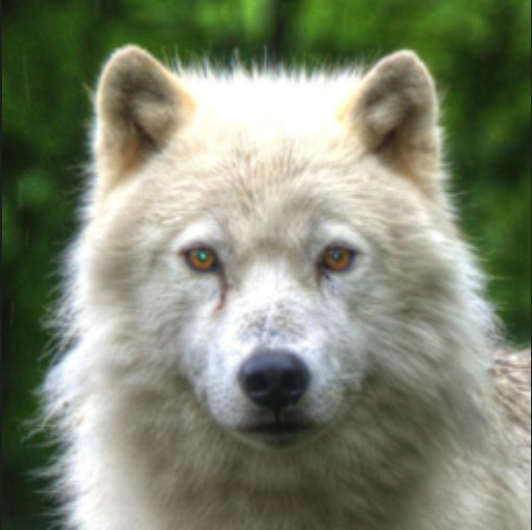}
        \caption{Reference}
    \end{subfigure}
    \begin{subfigure}[t]{0.19\textwidth}
        \includegraphics[width=\textwidth]{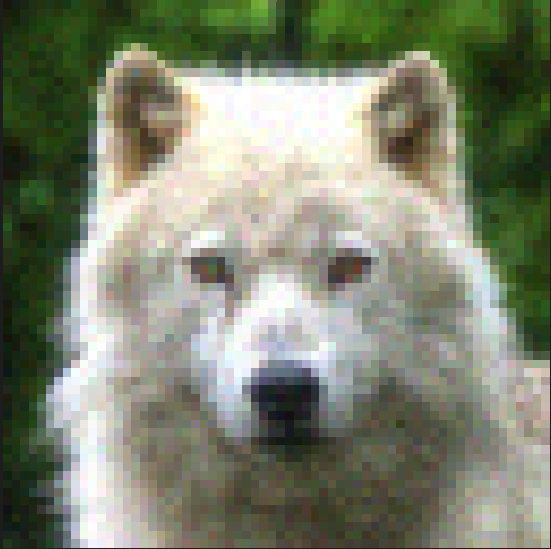}
        \caption{Distorted}
    \end{subfigure}
    \begin{subfigure}[t]{0.19\textwidth}
        \includegraphics[width=\textwidth]{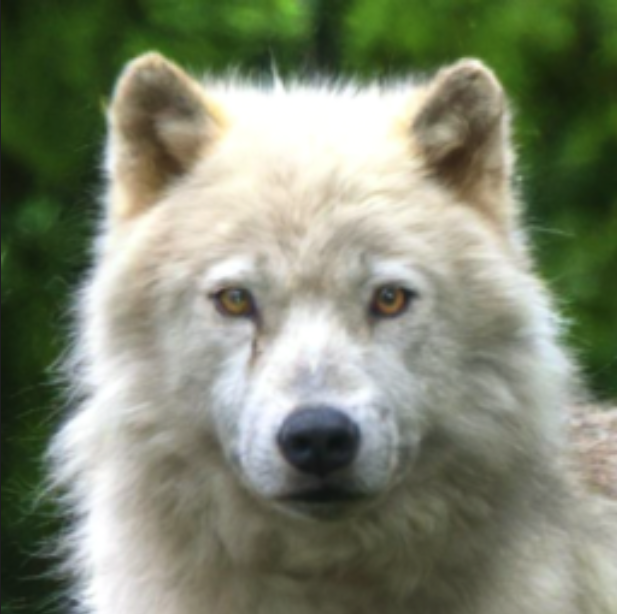}
        \caption{OT-ODE}
    \end{subfigure}
    \begin{subfigure}[t]{0.19\textwidth}
        \includegraphics[width=\textwidth]{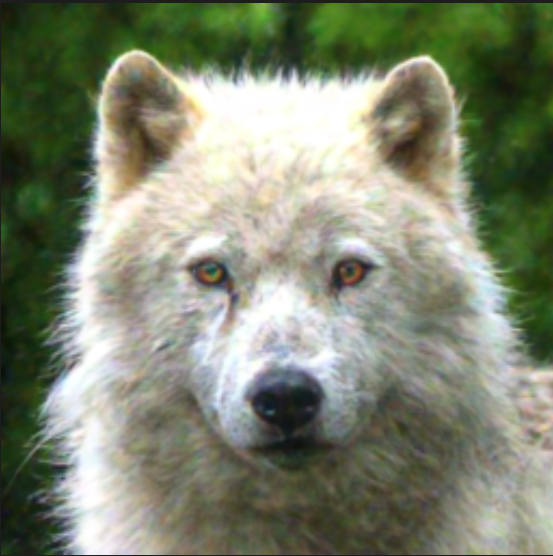}
        \caption{VP-ODE}
    \end{subfigure}
    \begin{subfigure}[t]{0.19\textwidth}    
        \includegraphics[width=\textwidth]{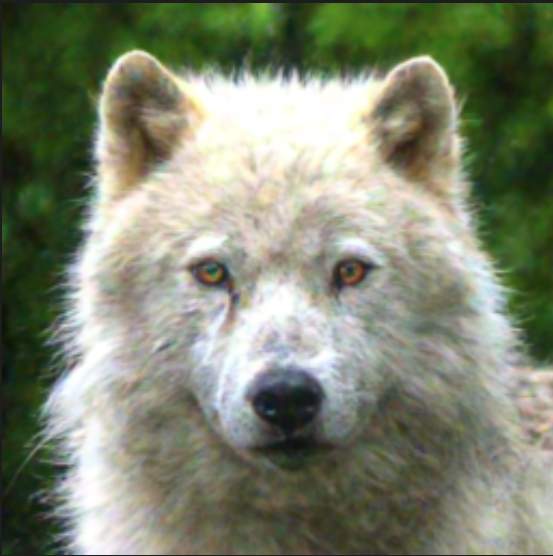}
        \caption{$\Pi$GDM}
    \end{subfigure}
    \caption{Super-resolution with conditional OT model and $\sigma_y=0.05$ for (\textbf{first row}) face-blurred ImageNet-64 $2\times$, (\textbf{second row}) face-blurred ImageNet-128 $2\times$, and (\textbf{third row}) AFHQ $4\times$.}
    \label{fig:super-resolution-noisy-ot-vis}
\end{figure}

\begin{figure}[!htb]
    \begin{subfigure}[t]{0.19\textwidth}
        \includegraphics[width=\textwidth]{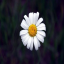}
    \end{subfigure}
    \begin{subfigure}[t]{0.19\textwidth}
        \includegraphics[width=\textwidth]{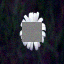}
    \end{subfigure}
    \begin{subfigure}[t]{0.19\textwidth}
        \includegraphics[width=\textwidth]{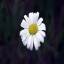}
    \end{subfigure}
    \begin{subfigure}[t]{0.19\textwidth}
        \includegraphics[width=\textwidth]{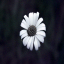}
    \end{subfigure}
    \begin{subfigure}[t]{0.19\textwidth}    
        \includegraphics[width=\textwidth]{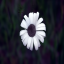}
    \end{subfigure} \\
        \begin{subfigure}[t]{0.19\textwidth}
        \includegraphics[width=\textwidth]{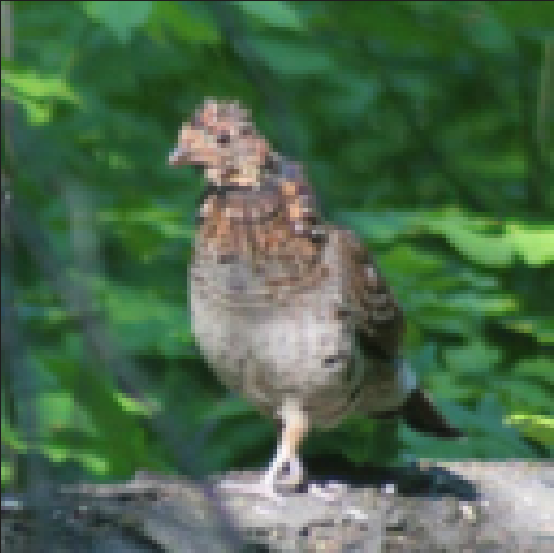}
    \end{subfigure}
    \begin{subfigure}[t]{0.19\textwidth}
        \includegraphics[width=\textwidth]{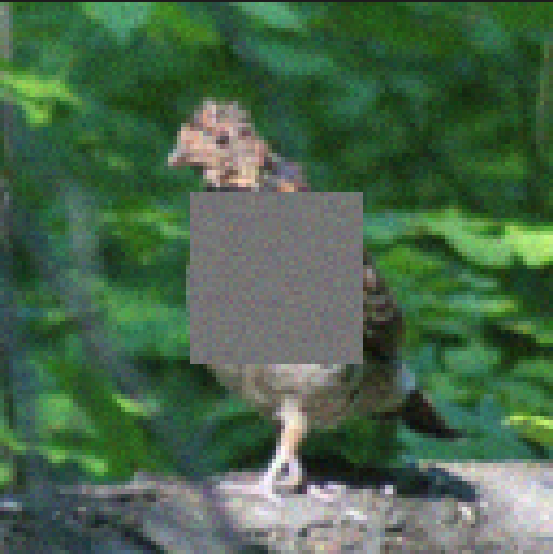}
    \end{subfigure}
    \begin{subfigure}[t]{0.19\textwidth}
        \includegraphics[width=\textwidth]{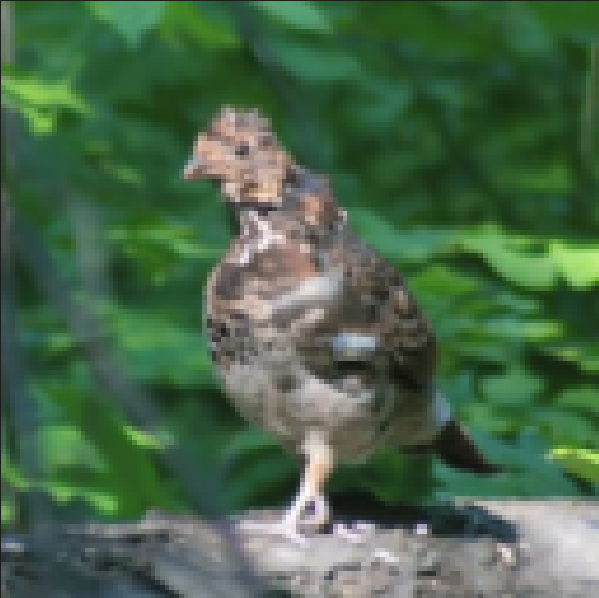}
    \end{subfigure}
    \begin{subfigure}[t]{0.19\textwidth}
        \includegraphics[width=\textwidth]{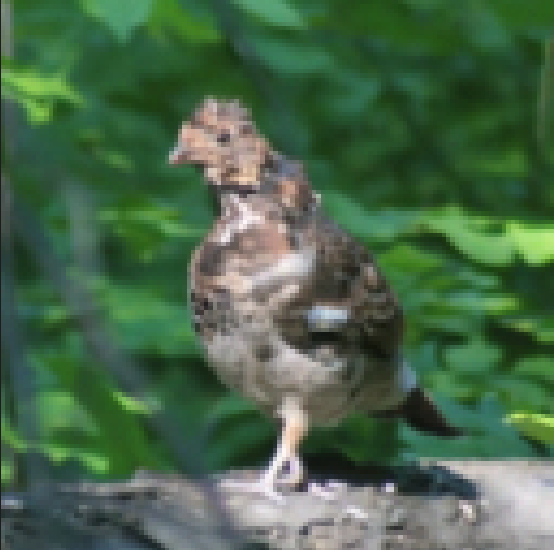}
    \end{subfigure}
    \begin{subfigure}[t]{0.19\textwidth}    
        \includegraphics[width=\textwidth]{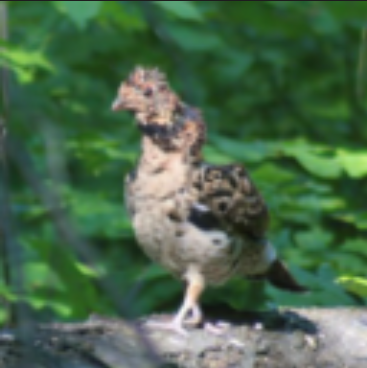}
    \end{subfigure}\\
    \begin{subfigure}[t]{0.19\textwidth}
        \includegraphics[width=\textwidth]{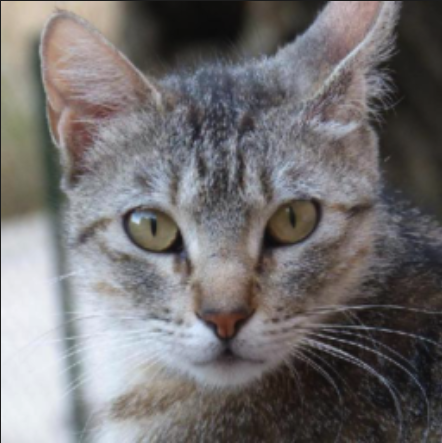}
        \caption{Reference}
    \end{subfigure}
    \begin{subfigure}[t]{0.19\textwidth}
        \includegraphics[width=\textwidth]{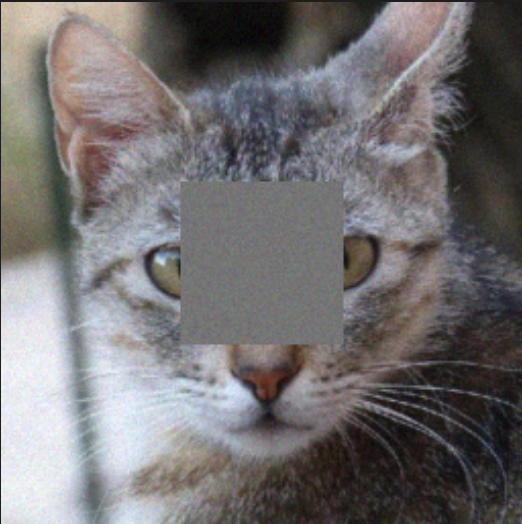}
        \caption{Distorted}
    \end{subfigure}
    \begin{subfigure}[t]{0.19\textwidth}
        \includegraphics[width=\textwidth]{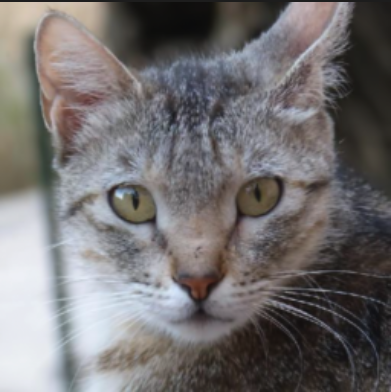}
        \caption{OT-ODE}
    \end{subfigure}
    \begin{subfigure}[t]{0.19\textwidth}
        \includegraphics[width=\textwidth]{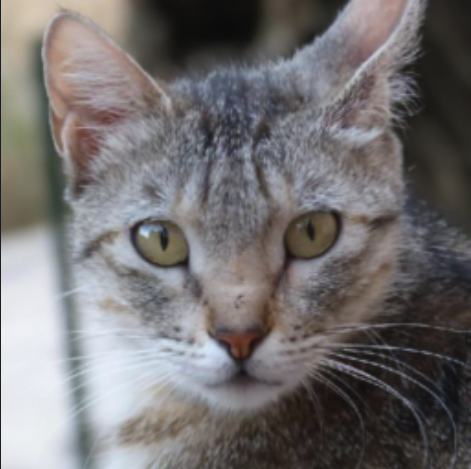}
        \caption{VP-ODE}
    \end{subfigure}
    \begin{subfigure}[t]{0.19\textwidth}    
        \includegraphics[width=\textwidth]{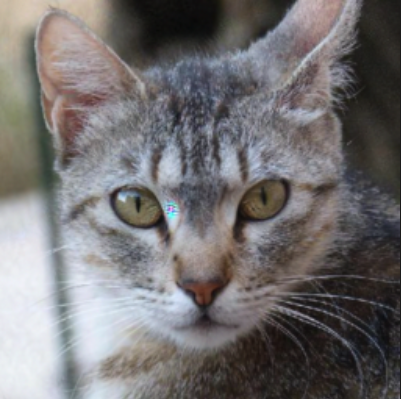}
        \caption{$\Pi$GDM}
    \end{subfigure}
    \caption{Inpainting (Center mask) with conditional OT model and $\sigma_y=0.05$ for (\textbf{first row}) face-blurred ImageNet-64, (\textbf{second row}) face-blurred ImageNet-128, and (\textbf{third row}) AFHQ.}
    \label{fig:inpainting-noisy-ot-vis}
\end{figure}

\begin{figure}[!htb]
    \begin{subfigure}[t]{0.19\textwidth}
        \includegraphics[width=\textwidth]{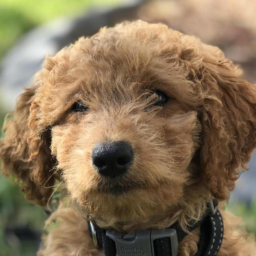}
    \end{subfigure}
    \begin{subfigure}[t]{0.19\textwidth}
        \includegraphics[width=\textwidth]{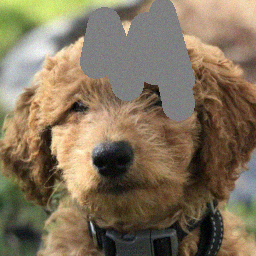}
    \end{subfigure}
    \begin{subfigure}[t]{0.19\textwidth}
        \includegraphics[width=\textwidth]{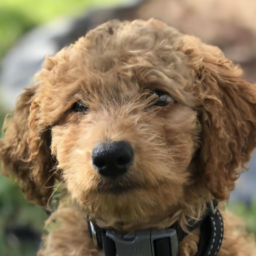}
    \end{subfigure}
    \begin{subfigure}[t]{0.19\textwidth}
        \includegraphics[width=\textwidth]{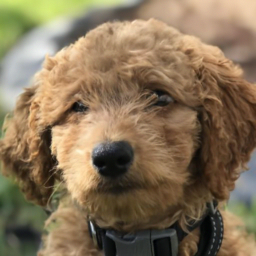}
    \end{subfigure}
    \begin{subfigure}[t]{0.19\textwidth}    
        \includegraphics[width=\textwidth]{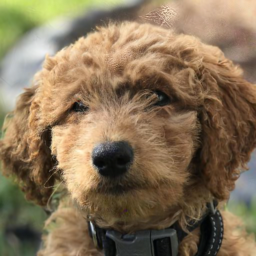}
    \end{subfigure}\\
    \begin{subfigure}[t]{0.19\textwidth}
        \includegraphics[width=\textwidth]{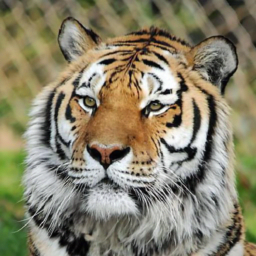}
    \end{subfigure}
    \begin{subfigure}[t]{0.19\textwidth}
        \includegraphics[width=\textwidth]{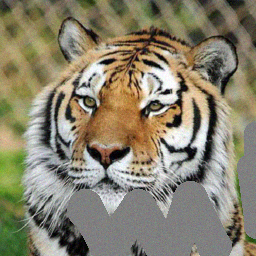}
    \end{subfigure}
    \begin{subfigure}[t]{0.19\textwidth}
        \includegraphics[width=\textwidth]{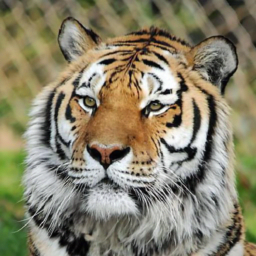}
    \end{subfigure}
    \begin{subfigure}[t]{0.19\textwidth}
        \includegraphics[width=\textwidth]{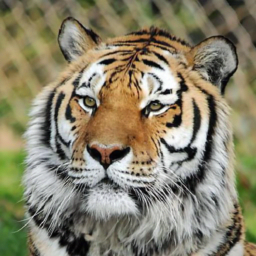}
    \end{subfigure}
    \begin{subfigure}[t]{0.19\textwidth}    
        \includegraphics[width=\textwidth]{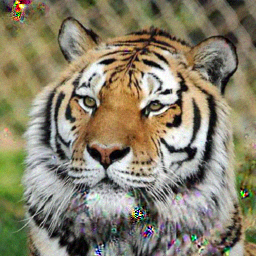}
    \end{subfigure} \\
    \begin{subfigure}[t]{0.19\textwidth}
        \includegraphics[width=\textwidth]{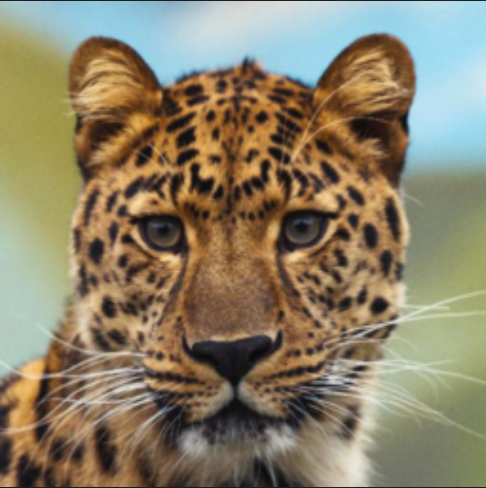}
    \end{subfigure}
    \begin{subfigure}[t]{0.19\textwidth}
        \includegraphics[width=\textwidth]{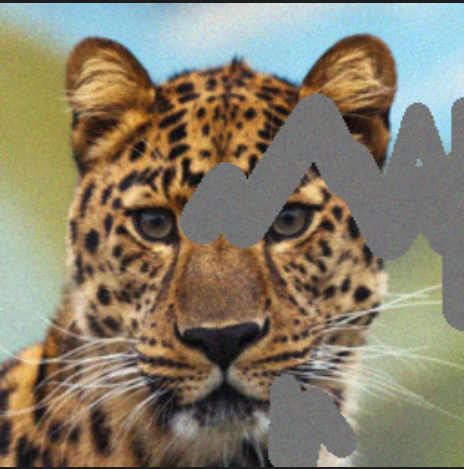}
    \end{subfigure}
    \begin{subfigure}[t]{0.19\textwidth}
        \includegraphics[width=\textwidth]{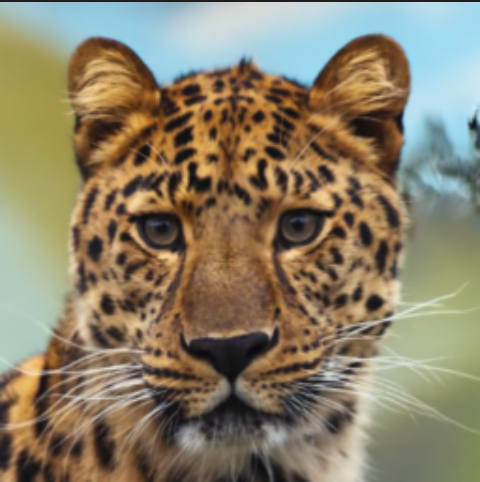}
    \end{subfigure}
    \begin{subfigure}[t]{0.19\textwidth}
        \includegraphics[width=\textwidth]{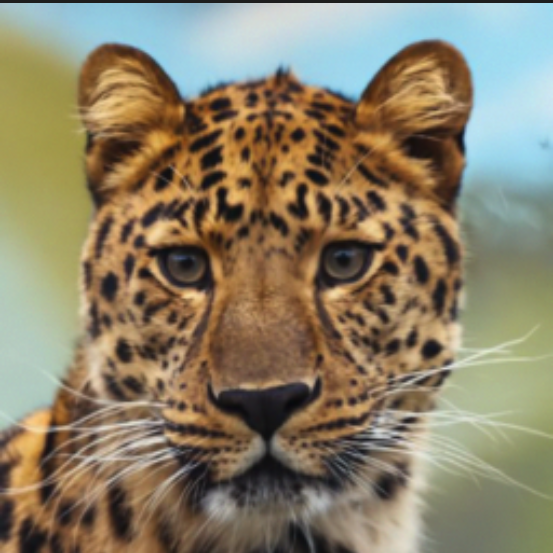}
    \end{subfigure}
    \begin{subfigure}[t]{0.19\textwidth}    
        \includegraphics[width=\textwidth]{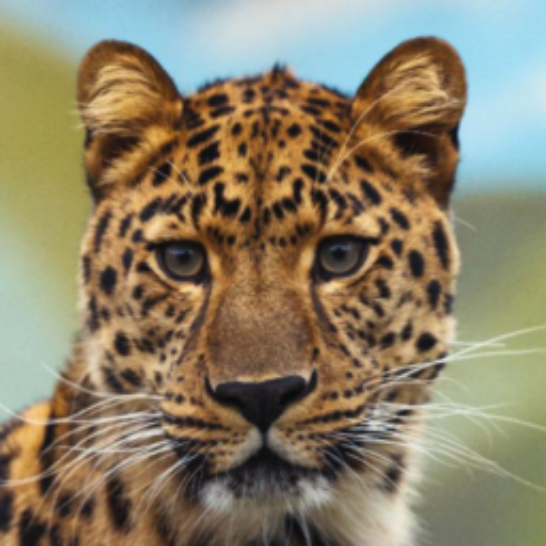}
    \end{subfigure}\\
    \begin{subfigure}[t]{0.19\textwidth}
        \includegraphics[width=\textwidth]{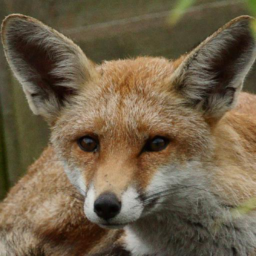}
        \caption{Reference}
    \end{subfigure}
    \begin{subfigure}[t]{0.19\textwidth}
        \includegraphics[width=\textwidth]{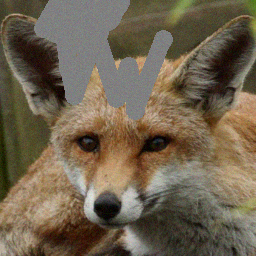}
        \caption{Distorted}
    \end{subfigure}
    \begin{subfigure}[t]{0.19\textwidth}
        \includegraphics[width=\textwidth]{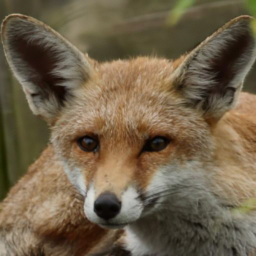}
        \caption{OT-ODE}
    \end{subfigure}
    \begin{subfigure}[t]{0.19\textwidth}
        \includegraphics[width=\textwidth]{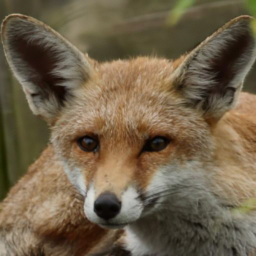}
        \caption{VP-ODE}
    \end{subfigure}
    \begin{subfigure}[t]{0.19\textwidth}    
        \includegraphics[width=\textwidth]{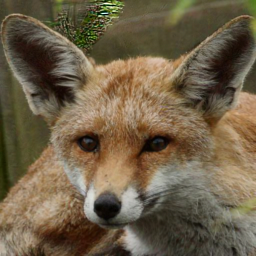}
        \caption{$\Pi$GDM}
    \end{subfigure}
    \caption{Inpainting (Free-form mask) with conditional OT model and $\sigma_y=0.05$ for AFHQ.}
    \label{fig:inpainting-ff-noisy-ot-vis}
\end{figure}

\begin{figure}[!htb]
    \begin{subfigure}[t]{0.19\textwidth}
        \includegraphics[width=\textwidth]{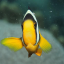}
    \end{subfigure}
    \begin{subfigure}[t]{0.19\textwidth}
        \includegraphics[width=\textwidth]{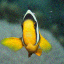}
    \end{subfigure}
    \begin{subfigure}[t]{0.19\textwidth}
        \includegraphics[width=\textwidth]{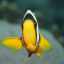}
    \end{subfigure}
    \begin{subfigure}[t]{0.19\textwidth}
        \includegraphics[width=\textwidth]{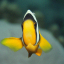}
    \end{subfigure}
    \begin{subfigure}[t]{0.19\textwidth}    
        \includegraphics[width=\textwidth]{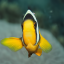}
    \end{subfigure} \\
        \begin{subfigure}[t]{0.19\textwidth}
        \includegraphics[width=\textwidth]{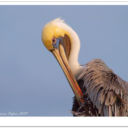}
    \end{subfigure}
    \begin{subfigure}[t]{0.19\textwidth}
        \includegraphics[width=\textwidth]{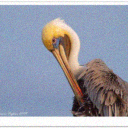}
    \end{subfigure}
    \begin{subfigure}[t]{0.19\textwidth}
        \includegraphics[width=\textwidth]{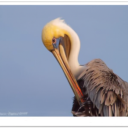}
    \end{subfigure}
    \begin{subfigure}[t]{0.19\textwidth}
        \includegraphics[width=\textwidth]{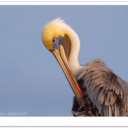}
    \end{subfigure}
    \begin{subfigure}[t]{0.19\textwidth}    
        \includegraphics[width=\textwidth]{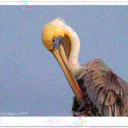}
    \end{subfigure}\\
    \begin{subfigure}[t]{0.19\textwidth}
        \includegraphics[width=\textwidth]{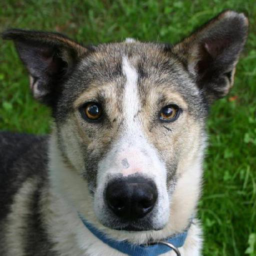}
        \caption{Reference}
    \end{subfigure}
    \begin{subfigure}[t]{0.19\textwidth}
        \includegraphics[width=\textwidth]{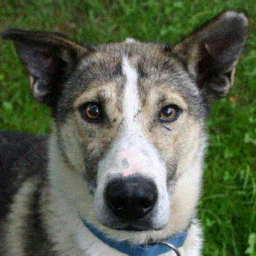}
        \caption{Distorted}
    \end{subfigure}
    \begin{subfigure}[t]{0.19\textwidth}
        \includegraphics[width=\textwidth]{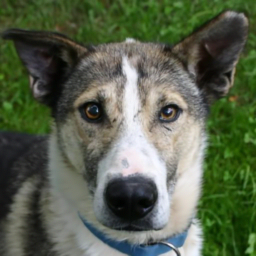}
        \caption{OT-ODE}
    \end{subfigure}
    \begin{subfigure}[t]{0.19\textwidth}
        \includegraphics[width=\textwidth]{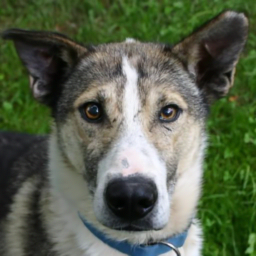}
        \caption{VP-ODE}
    \end{subfigure}
    \begin{subfigure}[t]{0.19\textwidth}    
        \includegraphics[width=\textwidth]{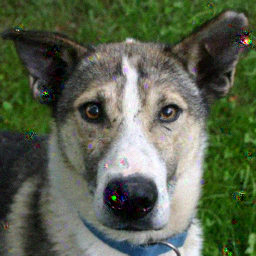}
        \caption{$\Pi$GDM}
    \end{subfigure}
    \caption{Denoising with conditional OT model and $\sigma_y=0.05$ for (\textbf{first row}) face-blurred ImageNet-64, (\textbf{second row}) face-blurred ImageNet-128, and (\textbf{third row}) AFHQ.}
    \label{fig:denoising-ot-vis}
\end{figure}

\begin{figure}[!htb]
    \begin{subfigure}[t]{0.16\textwidth}
        \includegraphics[width=\textwidth]{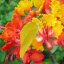}
    \end{subfigure}
    \begin{subfigure}[t]{0.16\textwidth}
        \includegraphics[width=\textwidth]{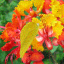}
    \end{subfigure}
    \begin{subfigure}[t]{0.16\textwidth}
        \includegraphics[width=\textwidth]{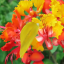}
    \end{subfigure}
    \begin{subfigure}[t]{0.16\textwidth}
        \includegraphics[width=\textwidth]{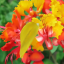}
    \end{subfigure}
    \begin{subfigure}[t]{0.16\textwidth}    
        \includegraphics[width=\textwidth]{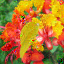}
    \end{subfigure} 
    \begin{subfigure}[t]{0.16\textwidth}    
        \includegraphics[width=\textwidth]{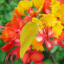}
    \end{subfigure} 
    \\
        \begin{subfigure}[t]{0.16\textwidth}
        \includegraphics[width=\textwidth]{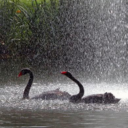}
    \end{subfigure}
    \begin{subfigure}[t]{0.16\textwidth}
        \includegraphics[width=\textwidth]{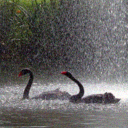}
    \end{subfigure}
    \begin{subfigure}[t]{0.16\textwidth}
        \includegraphics[width=\textwidth]{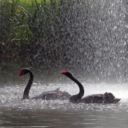}
    \end{subfigure}
    \begin{subfigure}[t]{0.16\textwidth}
        \includegraphics[width=\textwidth]{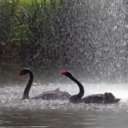}
    \end{subfigure}
    \begin{subfigure}[t]{0.16\textwidth}    
        \includegraphics[width=\textwidth]{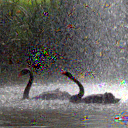}
    \end{subfigure}
    \begin{subfigure}[t]{0.16\textwidth}    
        \includegraphics[width=\textwidth]{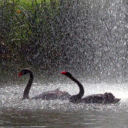}
    \end{subfigure}
    \\
    \begin{subfigure}[t]{0.16\textwidth}
        \includegraphics[width=\textwidth]{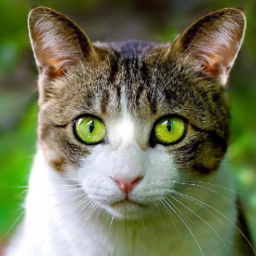}
        \caption{Reference}
    \end{subfigure}
    \begin{subfigure}[t]{0.16\textwidth}
        \includegraphics[width=\textwidth]{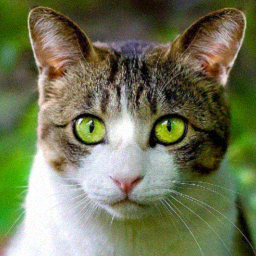}
        \caption{Distorted}
    \end{subfigure}
    \begin{subfigure}[t]{0.16\textwidth}
        \includegraphics[width=\textwidth]{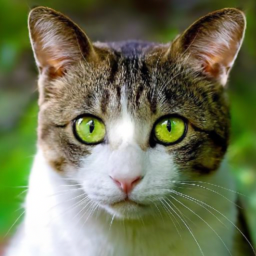}
        \caption{OT-ODE}
    \end{subfigure}
    \begin{subfigure}[t]{0.16\textwidth}
        \includegraphics[width=\textwidth]{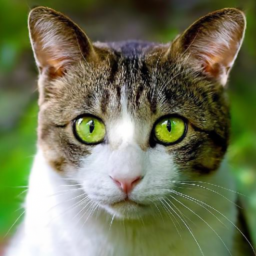}
        \caption{VP-ODE}
    \end{subfigure}
    \begin{subfigure}[t]{0.16\textwidth}    
        \includegraphics[width=\textwidth]{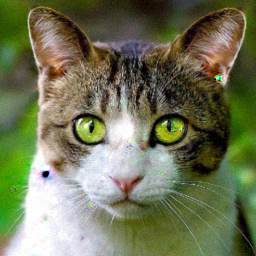}
        \caption{$\Pi$GDM}
    \end{subfigure}
    \begin{subfigure}[t]{0.16\textwidth}    
        \includegraphics[width=\textwidth]{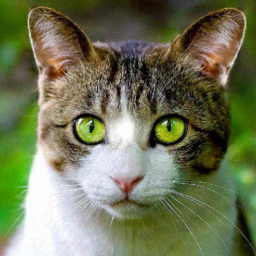}
        \caption{RED-Diff}
    \end{subfigure}
    \caption{Denoising with pretrained VP-SDE model and $\sigma_y=0.05$ for (\textbf{first row}) face-blurred ImageNet-64, (\textbf{second row}) face-blurred ImageNet-128, and (\textbf{third row}) AFHQ.}
    \label{fig:denoising-ddpm-vis}
\end{figure}

\begin{figure}[!htb]
    \begin{subfigure}[t]{0.19\textwidth}
        \includegraphics[width=\textwidth]{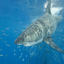}
    \end{subfigure}
    \begin{subfigure}[t]{0.19\textwidth}
        \includegraphics[width=\textwidth]{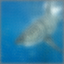}
    \end{subfigure}
    \begin{subfigure}[t]{0.19\textwidth}
        \includegraphics[width=\textwidth]{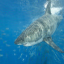}
    \end{subfigure}
    \begin{subfigure}[t]{0.19\textwidth}
        \includegraphics[width=\textwidth]{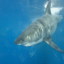}
    \end{subfigure}
    \begin{subfigure}[t]{0.19\textwidth}    
        \includegraphics[width=\textwidth]{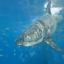}
    \end{subfigure}\\
    \begin{subfigure}[t]{0.19\textwidth}
        \includegraphics[width=\textwidth]{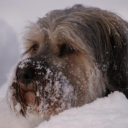}
    \end{subfigure}
    \begin{subfigure}[t]{0.19\textwidth}
        \includegraphics[width=\textwidth]{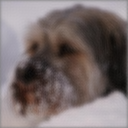}
    \end{subfigure}
    \begin{subfigure}[t]{0.19\textwidth}
        \includegraphics[width=\textwidth]{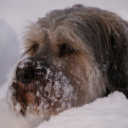}
    \end{subfigure}
    \begin{subfigure}[t]{0.19\textwidth}
        \includegraphics[width=\textwidth]{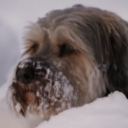}
    \end{subfigure}
    \begin{subfigure}[t]{0.19\textwidth}    
        \includegraphics[width=\textwidth]{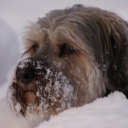}
    \end{subfigure} \\
    \begin{subfigure}[t]{0.19\textwidth}
        \includegraphics[width=\textwidth]{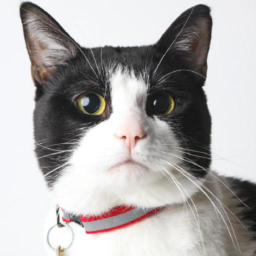}
        \caption{Reference}
    \end{subfigure}
    \begin{subfigure}[t]{0.19\textwidth}
        \includegraphics[width=\textwidth]{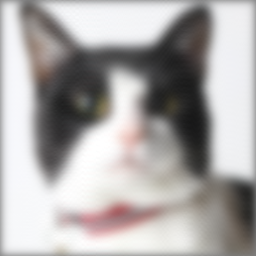}
        \caption{Distorted}
    \end{subfigure}
    \begin{subfigure}[t]{0.19\textwidth}
        \includegraphics[width=\textwidth]{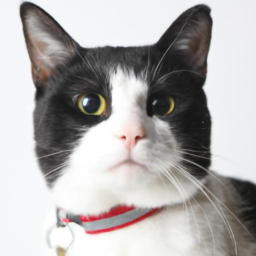}
        \caption{OT-ODE}
    \end{subfigure}
    \begin{subfigure}[t]{0.19\textwidth}
        \includegraphics[width=\textwidth]{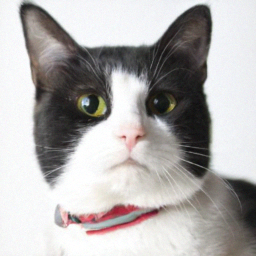}
        \caption{VP-ODE}
    \end{subfigure}
    \begin{subfigure}[t]{0.19\textwidth}    
        \includegraphics[width=\textwidth]{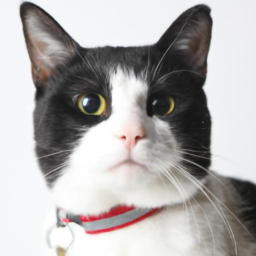}
        \caption{$\Pi$GDM}
    \end{subfigure}
    \caption{Gaussian deblurring with conditional OT model and $\sigma_y=0$ for (\textbf{first row}) face-blurred ImageNet-64, (\textbf{second row}) face-blurred ImageNet-128 and (\textbf{third row}) AFHQ.}
    \label{fig:gb-noiseless-ot-vis}
\end{figure}

\begin{figure}[!htb]
    \begin{subfigure}[t]{0.19\textwidth}
        \includegraphics[width=\textwidth]{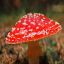}
    \end{subfigure}
    \begin{subfigure}[t]{0.19\textwidth}
        \includegraphics[width=\textwidth]{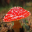}
    \end{subfigure}
    \begin{subfigure}[t]{0.19\textwidth}
        \includegraphics[width=\textwidth]{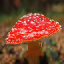}
    \end{subfigure}
    \begin{subfigure}[t]{0.19\textwidth}
        \includegraphics[width=\textwidth]{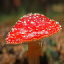}
    \end{subfigure}
    \begin{subfigure}[t]{0.19\textwidth}    
        \includegraphics[width=\textwidth]{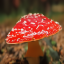}
    \end{subfigure}\\
    \begin{subfigure}[t]{0.19\textwidth}
        \includegraphics[width=\textwidth]{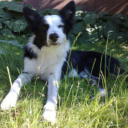}
    \end{subfigure}
    \begin{subfigure}[t]{0.19\textwidth}
        \includegraphics[width=\textwidth]{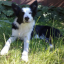}
    \end{subfigure}
    \begin{subfigure}[t]{0.19\textwidth}
        \includegraphics[width=\textwidth]{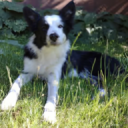}
    \end{subfigure}
    \begin{subfigure}[t]{0.19\textwidth}
        \includegraphics[width=\textwidth]{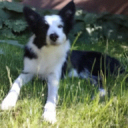}
    \end{subfigure}
    \begin{subfigure}[t]{0.19\textwidth}    
        \includegraphics[width=\textwidth]{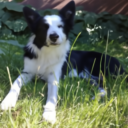}
    \end{subfigure} \\
    \begin{subfigure}[t]{0.19\textwidth}
        \includegraphics[width=\textwidth]{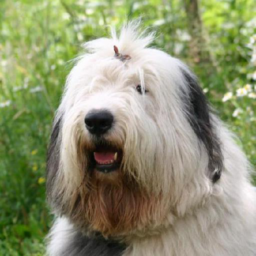}
        \caption{Reference}
    \end{subfigure}
    \begin{subfigure}[t]{0.19\textwidth}
        \includegraphics[width=\textwidth]{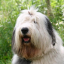}
        \caption{Distorted}
    \end{subfigure}
    \begin{subfigure}[t]{0.19\textwidth}
        \includegraphics[width=\textwidth]{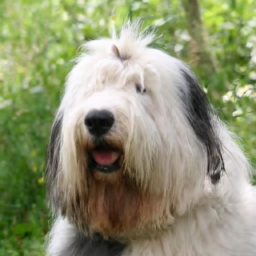}
        \caption{OT-ODE}
    \end{subfigure}
    \begin{subfigure}[t]{0.19\textwidth}
        \includegraphics[width=\textwidth]{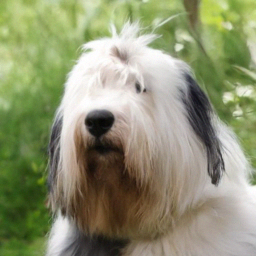}
        \caption{VP-ODE}
    \end{subfigure}
    \begin{subfigure}[t]{0.19\textwidth}    
        \includegraphics[width=\textwidth]{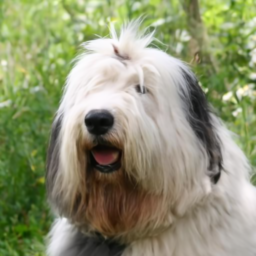}
        \caption{$\Pi$GDM}
    \end{subfigure}
    \caption{Super-resolution with conditional OT model and $\sigma_y=0$ for (\textbf{first row}) face-blurred ImageNet-64, (\textbf{second row}) face-blurred ImageNet-128 and (\textbf{third row}) AFHQ.}
    \label{fig:sr-noiseless-ot-vis}
\end{figure}

\begin{figure}[!htb]
    \begin{subfigure}[t]{0.19\textwidth}
        \includegraphics[width=\textwidth]{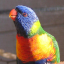}
    \end{subfigure}
    \begin{subfigure}[t]{0.19\textwidth}
        \includegraphics[width=\textwidth]{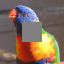}
    \end{subfigure}
    \begin{subfigure}[t]{0.19\textwidth}
        \includegraphics[width=\textwidth]{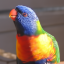}
    \end{subfigure}
    \begin{subfigure}[t]{0.19\textwidth}
        \includegraphics[width=\textwidth]{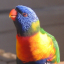}
    \end{subfigure}
    \begin{subfigure}[t]{0.19\textwidth}    
        \includegraphics[width=\textwidth]{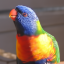}
    \end{subfigure}\\
    \begin{subfigure}[t]{0.19\textwidth}
        \includegraphics[width=\textwidth]{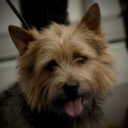}
    \end{subfigure}
    \begin{subfigure}[t]{0.19\textwidth}
        \includegraphics[width=\textwidth]{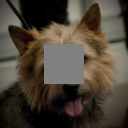}
    \end{subfigure}
    \begin{subfigure}[t]{0.19\textwidth}
        \includegraphics[width=\textwidth]{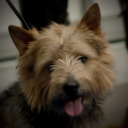}
    \end{subfigure}
    \begin{subfigure}[t]{0.19\textwidth}
        \includegraphics[width=\textwidth]{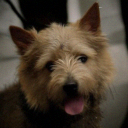}
    \end{subfigure}
    \begin{subfigure}[t]{0.19\textwidth}    
        \includegraphics[width=\textwidth]{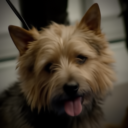}
    \end{subfigure} \\
    \begin{subfigure}[t]{0.19\textwidth}
        \includegraphics[width=\textwidth]{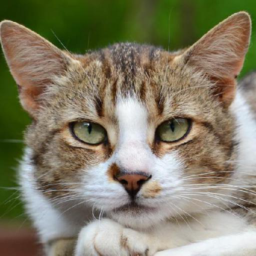}
        \caption{Reference}
    \end{subfigure}
    \begin{subfigure}[t]{0.19\textwidth}
        \includegraphics[width=\textwidth]{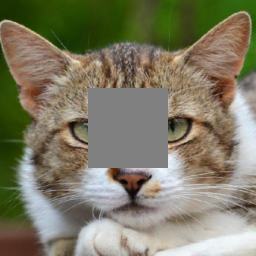}
        \caption{Distorted}
    \end{subfigure}
    \begin{subfigure}[t]{0.19\textwidth}
        \includegraphics[width=\textwidth]{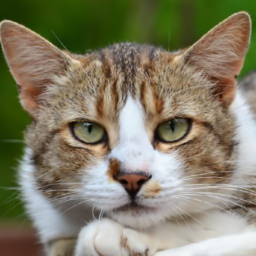}
        \caption{OT-ODE}
    \end{subfigure}
    \begin{subfigure}[t]{0.19\textwidth}
        \includegraphics[width=\textwidth]{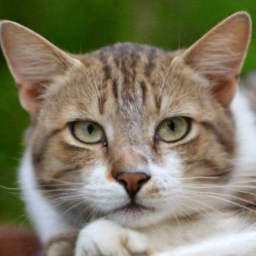}
        \caption{VP-ODE}
    \end{subfigure}
    \begin{subfigure}[t]{0.19\textwidth}    
        \includegraphics[width=\textwidth]{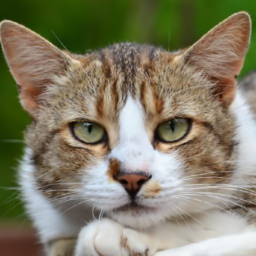}
        \caption{$\Pi$GDM}
    \end{subfigure}
    \caption{Inpainting (centered mask) with conditional OT model and $\sigma_y=0$ for (\textbf{first row}) face-blurred ImageNet-64, (\textbf{second row}) face-blurred ImageNet-128 and (\textbf{third row}) AFHQ.}
    \label{fig:ipc-noiseless-ot-vis}
\end{figure}

\begin{figure}[!htb]
    \begin{subfigure}[t]{0.16\textwidth}
        \includegraphics[width=\textwidth]{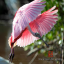}
    \end{subfigure}
    \begin{subfigure}[t]{0.16\textwidth}
        \includegraphics[width=\textwidth]{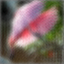}
    \end{subfigure}
    \begin{subfigure}[t]{0.16\textwidth}
        \includegraphics[width=\textwidth]{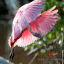}
    \end{subfigure}
    \begin{subfigure}[t]{0.16\textwidth}
        \includegraphics[width=\textwidth]{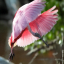}
    \end{subfigure}
    \begin{subfigure}[t]{0.16\textwidth}    
        \includegraphics[width=\textwidth]{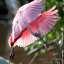}
    \end{subfigure}
    \begin{subfigure}[t]{0.16\textwidth}    
        \includegraphics[width=\textwidth]{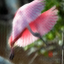}
    \end{subfigure}
    \\
    \begin{subfigure}[t]{0.16\textwidth}
        \includegraphics[width=\textwidth]{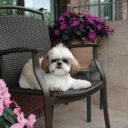}
    \end{subfigure}
    \begin{subfigure}[t]{0.16\textwidth}
        \includegraphics[width=\textwidth]{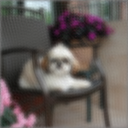}
    \end{subfigure}
    \begin{subfigure}[t]{0.16\textwidth}
        \includegraphics[width=\textwidth]{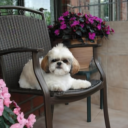}
    \end{subfigure}
    \begin{subfigure}[t]{0.16\textwidth}
        \includegraphics[width=\textwidth]{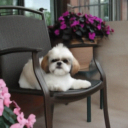}
    \end{subfigure}
    \begin{subfigure}[t]{0.16\textwidth}    
        \includegraphics[width=\textwidth]{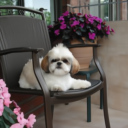}
    \end{subfigure} 
    \begin{subfigure}[t]{0.16\textwidth}    
        \includegraphics[width=\textwidth]{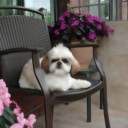}
    \end{subfigure}
    \\
    \begin{subfigure}[t]{0.16\textwidth}
        \includegraphics[width=\textwidth]{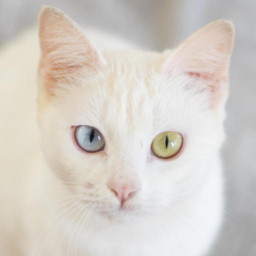}

    \end{subfigure}
    \begin{subfigure}[t]{0.16\textwidth}
        \includegraphics[width=\textwidth]{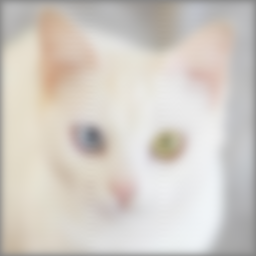}

    \end{subfigure}
    \begin{subfigure}[t]{0.16\textwidth}
        \includegraphics[width=\textwidth]{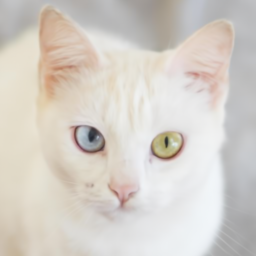}
    \end{subfigure}
    \begin{subfigure}[t]{0.16\textwidth}
        \includegraphics[width=\textwidth]{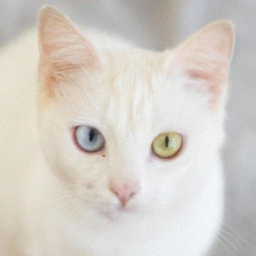}
    \end{subfigure}
    \begin{subfigure}[t]{0.16\textwidth}    
        \includegraphics[width=\textwidth]{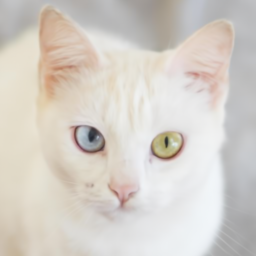}
    \end{subfigure}
    \begin{subfigure}[t]{0.16\textwidth}    
        \includegraphics[width=\textwidth]{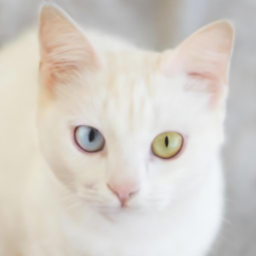}
    \end{subfigure} \\
    \begin{subfigure}[t]{0.16\textwidth}
        \includegraphics[width=\textwidth]{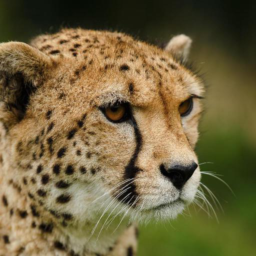}
        \caption{Reference}
    \end{subfigure}
    \begin{subfigure}[t]{0.16\textwidth}
        \includegraphics[width=\textwidth]{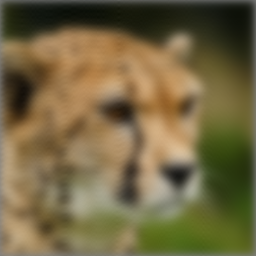}
        \caption{Distorted}
    \end{subfigure}
    \begin{subfigure}[t]{0.16\textwidth}
        \includegraphics[width=\textwidth]{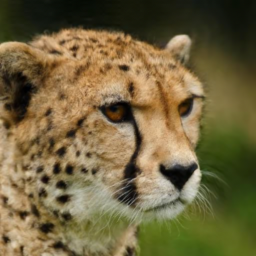}
        \caption{OT-ODE}
    \end{subfigure}
    \begin{subfigure}[t]{0.16\textwidth}
        \includegraphics[width=\textwidth]{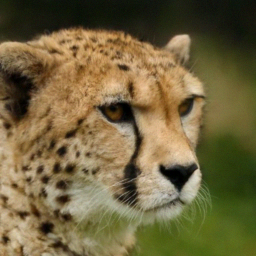}
        \caption{VP-ODE}
    \end{subfigure}
    \begin{subfigure}[t]{0.16\textwidth}    
        \includegraphics[width=\textwidth]{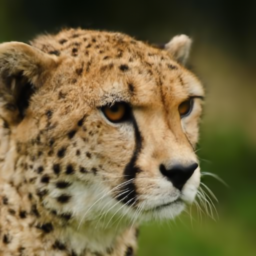}
        \caption{$\Pi$GDM}
    \end{subfigure}
    \begin{subfigure}[t]{0.16\textwidth}    
        \includegraphics[width=\textwidth]{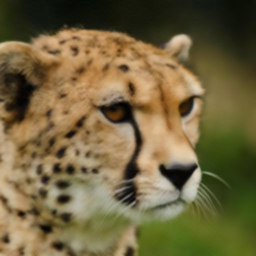}
        \caption{RED-Diff}
    \end{subfigure}
    \caption{Gaussian deblurring with VP-SDE model and $\sigma_y=0$ for (\textbf{first row}) face-blurred ImageNet-64, (\textbf{second row}) face-blurred ImageNet-128 and (\textbf{third and fourth row}) AFHQ.}
    \label{fig:gb-noiseless-ddpm-vis}
\end{figure}

\begin{figure}[!htb]
    \begin{subfigure}[t]{0.16\textwidth}
        \includegraphics[width=\textwidth]{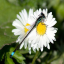}
    \end{subfigure}
    \begin{subfigure}[t]{0.16\textwidth}
        \includegraphics[width=\textwidth]{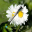}
    \end{subfigure}
    \begin{subfigure}[t]{0.16\textwidth}
        \includegraphics[width=\textwidth]{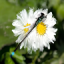}
    \end{subfigure}
    \begin{subfigure}[t]{0.16\textwidth}
        \includegraphics[width=\textwidth]{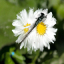}
    \end{subfigure}
    \begin{subfigure}[t]{0.16\textwidth}    
        \includegraphics[width=\textwidth]{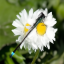}
    \end{subfigure}
    \begin{subfigure}[t]{0.16\textwidth}    
        \includegraphics[width=\textwidth]{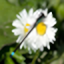}
    \end{subfigure}
    \\
    \begin{subfigure}[t]{0.16\textwidth}
        \includegraphics[width=\textwidth]{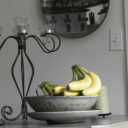}
    \end{subfigure}
    \begin{subfigure}[t]{0.16\textwidth}
        \includegraphics[width=\textwidth]{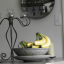}
    \end{subfigure}
    \begin{subfigure}[t]{0.16\textwidth}
        \includegraphics[width=\textwidth]{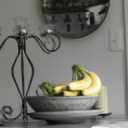}
    \end{subfigure}
    \begin{subfigure}[t]{0.16\textwidth}
        \includegraphics[width=\textwidth]{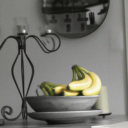}
    \end{subfigure}
    \begin{subfigure}[t]{0.16\textwidth}    
        \includegraphics[width=\textwidth]{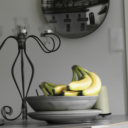}
    \end{subfigure} 
    \begin{subfigure}[t]{0.16\textwidth}    
        \includegraphics[width=\textwidth]{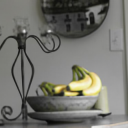}
    \end{subfigure} 
    \\
    \begin{subfigure}[t]{0.16\textwidth}
        \includegraphics[width=\textwidth]{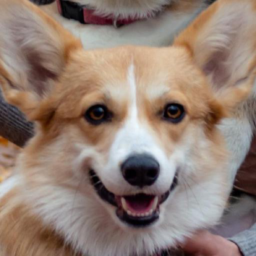}
        \caption{Reference}
    \end{subfigure}
    \begin{subfigure}[t]{0.16\textwidth}
        \includegraphics[width=\textwidth]{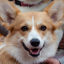}
        \caption{Distorted}
    \end{subfigure}
    \begin{subfigure}[t]{0.16\textwidth}
        \includegraphics[width=\textwidth]{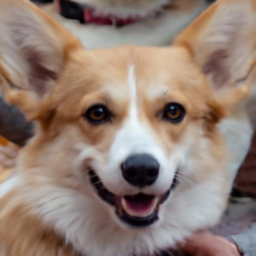}
        \caption{OT-ODE}
    \end{subfigure}
    \begin{subfigure}[t]{0.16\textwidth}
        \includegraphics[width=\textwidth]{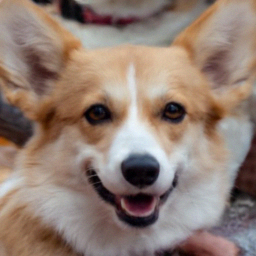}
        \caption{VP-ODE}
    \end{subfigure}
    \begin{subfigure}[t]{0.16\textwidth}    
        \includegraphics[width=\textwidth]{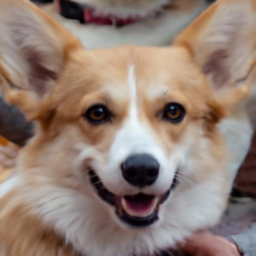}
        \caption{$\Pi$GDM}
    \end{subfigure}
    \begin{subfigure}[t]{0.16\textwidth}    
        \includegraphics[width=\textwidth]{images_noiseless/afhq/sr/723_1_ddpm_pigdm.png}
        \caption{RED-Diff}
    \end{subfigure}
    \caption{Super-resolution with VP-SDE model and $\sigma_y=0$ for (\textbf{first row}) face-blurred ImageNet-64, (\textbf{second row}) face-blurred ImageNet-128 and (\textbf{third row}) AFHQ.}
    \label{fig:sr-noiseless-ddpm-vis}
\end{figure}

\begin{figure}[!htb]
    \begin{subfigure}[t]{0.16\textwidth}
        \includegraphics[width=\textwidth]{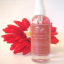}
    \end{subfigure}
    \begin{subfigure}[t]{0.16\textwidth}
        \includegraphics[width=\textwidth]{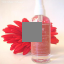}
    \end{subfigure}
    \begin{subfigure}[t]{0.16\textwidth}
        \includegraphics[width=\textwidth]{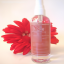}
    \end{subfigure}
    \begin{subfigure}[t]{0.16\textwidth}
        \includegraphics[width=\textwidth]{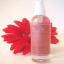}
    \end{subfigure}
    \begin{subfigure}[t]{0.16\textwidth}    
        \includegraphics[width=\textwidth]{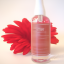}
    \end{subfigure}
        \begin{subfigure}[t]{0.16\textwidth}    
        \includegraphics[width=\textwidth]{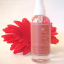}
    \end{subfigure}
    \\
    \begin{subfigure}[t]{0.16\textwidth}    
        \includegraphics[width=\textwidth]{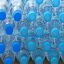}
    \end{subfigure}
        \begin{subfigure}[t]{0.16\textwidth}
        \includegraphics[width=\textwidth]{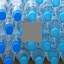}
    \end{subfigure}
    \begin{subfigure}[t]{0.16\textwidth}
        \includegraphics[width=\textwidth]{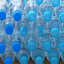}
    \end{subfigure}
    \begin{subfigure}[t]{0.16\textwidth}
        \includegraphics[width=\textwidth]{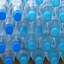}
    \end{subfigure}
    \begin{subfigure}[t]{0.16\textwidth}
        \includegraphics[width=\textwidth]{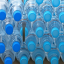}
    \end{subfigure}
    \begin{subfigure}[t]{0.16\textwidth}
        \includegraphics[width=\textwidth]{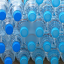}
    \end{subfigure}
    \\
    \begin{subfigure}[t]{0.16\textwidth}
        \includegraphics[width=\textwidth]{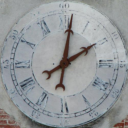}
    \end{subfigure}
    \begin{subfigure}[t]{0.16\textwidth}
        \includegraphics[width=\textwidth]{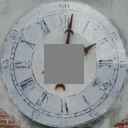}
    \end{subfigure}
    \begin{subfigure}[t]{0.16\textwidth}
        \includegraphics[width=\textwidth]{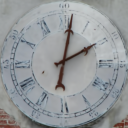}
    \end{subfigure}
    \begin{subfigure}[t]{0.16\textwidth}
        \includegraphics[width=\textwidth]{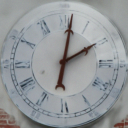}
    \end{subfigure}
    \begin{subfigure}[t]{0.16\textwidth}    
        \includegraphics[width=\textwidth]{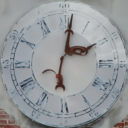}
    \end{subfigure} 
    \begin{subfigure}[t]{0.16\textwidth}    
        \includegraphics[width=\textwidth]{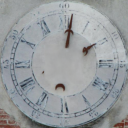}
    \end{subfigure}
    \\
    \begin{subfigure}[t]{0.16\textwidth}
        \includegraphics[width=\textwidth]{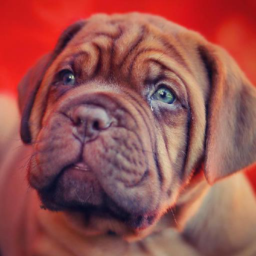}
        \caption{Reference}
    \end{subfigure}
    \begin{subfigure}[t]{0.16\textwidth}
        \includegraphics[width=\textwidth]{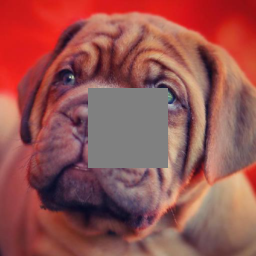}
        \caption{Distorted}
    \end{subfigure}
    \begin{subfigure}[t]{0.16\textwidth}
        \includegraphics[width=\textwidth]{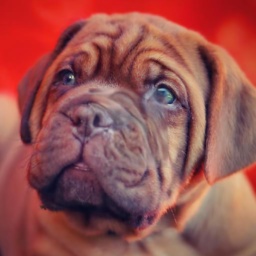}
        \caption{OT-ODE}
    \end{subfigure}
    \begin{subfigure}[t]{0.16\textwidth}
        \includegraphics[width=\textwidth]{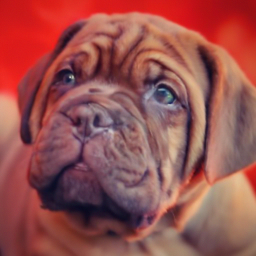}
        \caption{VP-ODE}
    \end{subfigure}
    \begin{subfigure}[t]{0.16\textwidth}    
        \includegraphics[width=\textwidth]{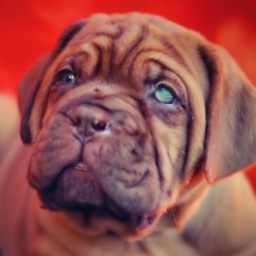}
        \caption{$\Pi$GDM}
    \end{subfigure}
    \begin{subfigure}[t]{0.16\textwidth}    
        \includegraphics[width=\textwidth]{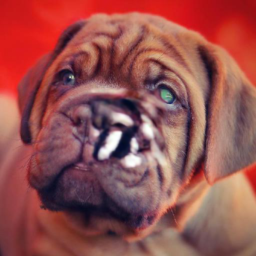}
        \caption{RED-Diff}
    \end{subfigure}
    \caption{Inpainting (centered mask) with VP-SDE model and $\sigma_y=0$ for (\textbf{first and second row}) face-blurred ImageNet-64, (\textbf{third  row}) face-blurred ImageNet-128 and (\textbf{fourth row}) AFHQ.}
    \label{fig:ipc-noiseless-ddpm-vis}
\end{figure}

\begin{figure}[!htb]
    \begin{subfigure}[t]{0.16\textwidth}
        \includegraphics[width=\textwidth]{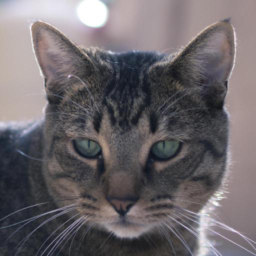}
    \end{subfigure}
    \begin{subfigure}[t]{0.16\textwidth}
        \includegraphics[width=\textwidth]{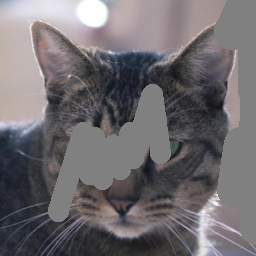}
    \end{subfigure}
    \begin{subfigure}[t]{0.16\textwidth}
        \includegraphics[width=\textwidth]{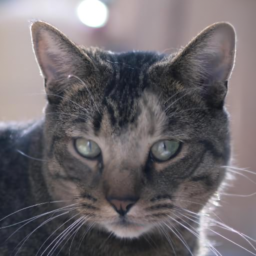}
    \end{subfigure}
    \begin{subfigure}[t]{0.16\textwidth}
        \includegraphics[width=\textwidth]{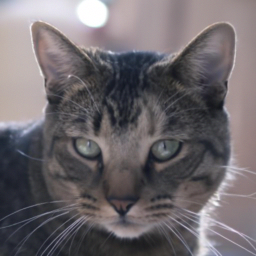}
    \end{subfigure}
    \begin{subfigure}[t]{0.16\textwidth}    
        \includegraphics[width=\textwidth]{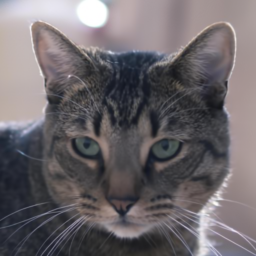}
    \end{subfigure}
    \begin{subfigure}[t]{0.16\textwidth}    
        \includegraphics[width=\textwidth]{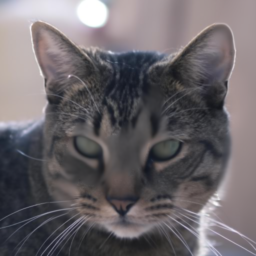}
    \end{subfigure}
    \\
    \begin{subfigure}[t]{0.16\textwidth}
        \includegraphics[width=\textwidth]{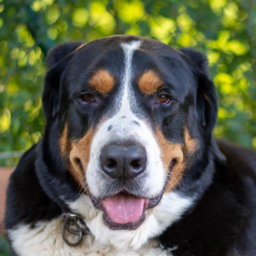}
    \end{subfigure}
    \begin{subfigure}[t]{0.16\textwidth}
        \includegraphics[width=\textwidth]{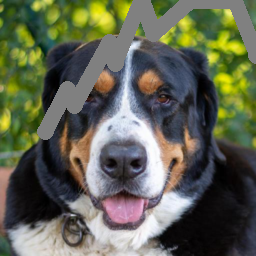}
    \end{subfigure}
    \begin{subfigure}[t]{0.16\textwidth}
        \includegraphics[width=\textwidth]{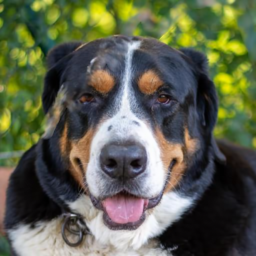}
    \end{subfigure}
    \begin{subfigure}[t]{0.16\textwidth}
        \includegraphics[width=\textwidth]{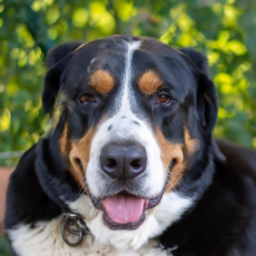}
    \end{subfigure}
    \begin{subfigure}[t]{0.16\textwidth}    
        \includegraphics[width=\textwidth]{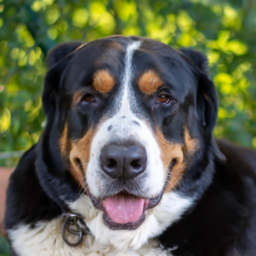}
    \end{subfigure} 
    \begin{subfigure}[t]{0.16\textwidth}    
        \includegraphics[width=\textwidth]{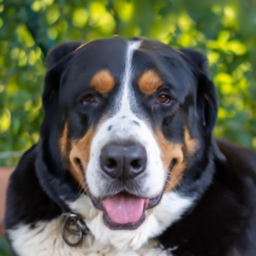}
    \end{subfigure}
    \\
    \begin{subfigure}[t]{0.16\textwidth}
        \includegraphics[width=\textwidth]{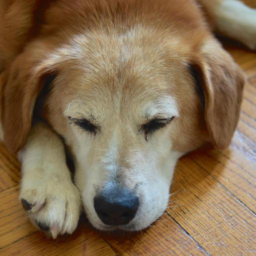}
        \caption{Reference}
    \end{subfigure}
    \begin{subfigure}[t]{0.16\textwidth}
        \includegraphics[width=\textwidth]{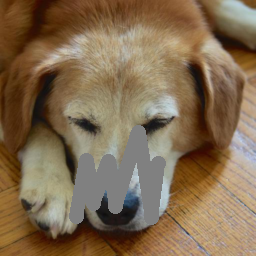}
        \caption{Distorted}
    \end{subfigure}
    \begin{subfigure}[t]{0.16\textwidth}
        \includegraphics[width=\textwidth]{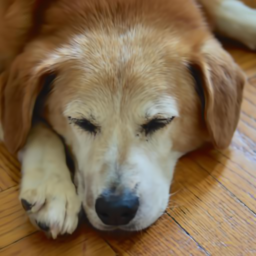}
        \caption{OT-ODE}
    \end{subfigure}
    \begin{subfigure}[t]{0.16\textwidth}
        \includegraphics[width=\textwidth]{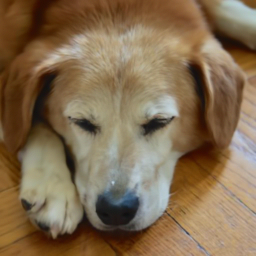}
        \caption{VP-ODE}
    \end{subfigure}
    \begin{subfigure}[t]{0.16\textwidth}    
        \includegraphics[width=\textwidth]{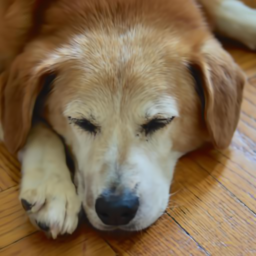}
        \caption{$\Pi$GDM}
    \end{subfigure}
    \begin{subfigure}[t]{0.16\textwidth}    
        \includegraphics[width=\textwidth]{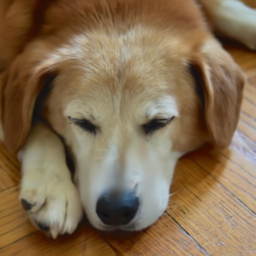}
        \caption{RED-Diff}
    \end{subfigure}
    \caption{Inpainting (freeform mask) with VP-SDE model and $\sigma_y=0$ for AFHQ.}
    \label{fig:ipf-noiseless-ot-ddpm-vis}
\end{figure}

%% file: sections/failure_cases_rev.tex
\clearpage
\subsection{Negative results from Inpainting} \label{sec:failure-modes-inpainting}
\begin{figure}[!htb]
     \begin{subfigure}[t]{0.19\textwidth}
         \includegraphics[width=\textwidth]{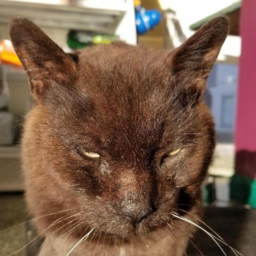}
     \end{subfigure}
     \begin{subfigure}[t]{0.19\textwidth}
         \includegraphics[width=\textwidth]{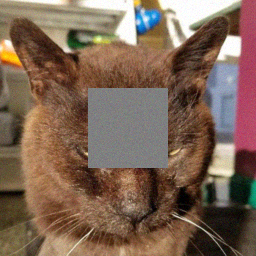}
     \end{subfigure}
     \begin{subfigure}[t]{0.19\textwidth}
         \includegraphics[width=\textwidth]{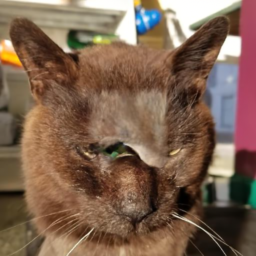}
     \end{subfigure}
         \begin{subfigure}[t]{0.19\textwidth}
         \includegraphics[width=\textwidth]{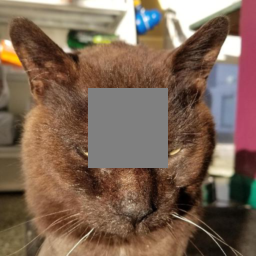}
     \end{subfigure}
     \begin{subfigure}[t]{0.19\textwidth}
         \includegraphics[width=\textwidth]{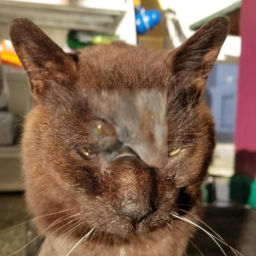}
     \end{subfigure}
    \\
    \begin{subfigure}[t]{0.19\textwidth}
        \includegraphics[width=\textwidth]{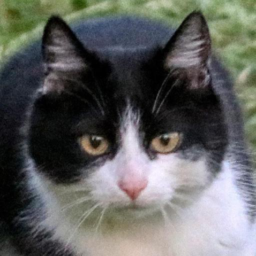}
    \end{subfigure}
    \begin{subfigure}[t]{0.19\textwidth}
        \includegraphics[width=\textwidth]{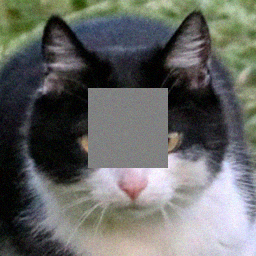}
    \end{subfigure}
    \begin{subfigure}[t]{0.19\textwidth}
        \includegraphics[width=\textwidth]{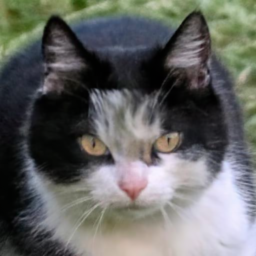}
    \end{subfigure}
        \begin{subfigure}[t]{0.19\textwidth}
        \includegraphics[width=\textwidth]{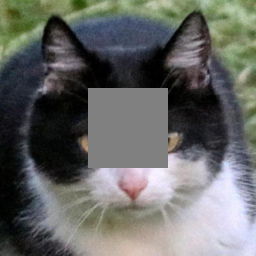}
    \end{subfigure}
    \begin{subfigure}[t]{0.19\textwidth}
        \includegraphics[width=\textwidth]{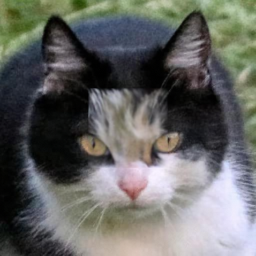}
    \end{subfigure}
    \\
    \begin{subfigure}[t]{0.19\textwidth}
        \includegraphics[width=\textwidth]{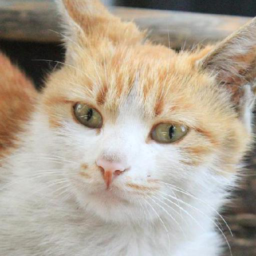}
    \end{subfigure}
    \begin{subfigure}[t]{0.19\textwidth}
        \includegraphics[width=\textwidth]{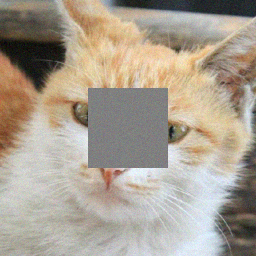}
    \end{subfigure}
    \begin{subfigure}[t]{0.19\textwidth}
        \includegraphics[width=\textwidth]{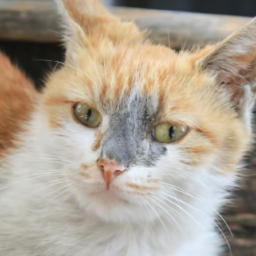}
    \end{subfigure}
        \begin{subfigure}[t]{0.19\textwidth}
        \includegraphics[width=\textwidth]{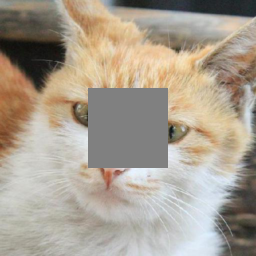}
    \end{subfigure}
    \begin{subfigure}[t]{0.19\textwidth}
        \includegraphics[width=\textwidth]{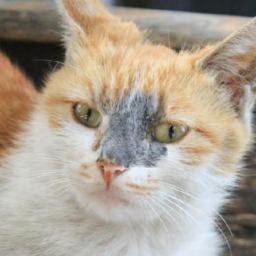}
    \end{subfigure}
    \\
    \begin{subfigure}[t]{0.19\textwidth}
        \includegraphics[width=\textwidth]{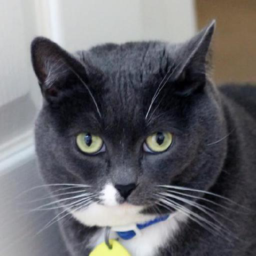}
    \end{subfigure}
    \begin{subfigure}[t]{0.19\textwidth}
        \includegraphics[width=\textwidth]{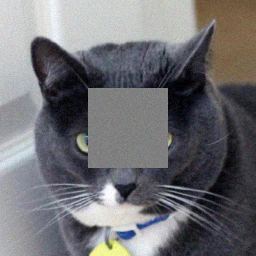}
    \end{subfigure}
    \begin{subfigure}[t]{0.19\textwidth}
        \includegraphics[width=\textwidth]{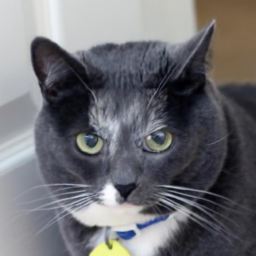}
    \end{subfigure}
        \begin{subfigure}[t]{0.19\textwidth}
        \includegraphics[width=\textwidth]{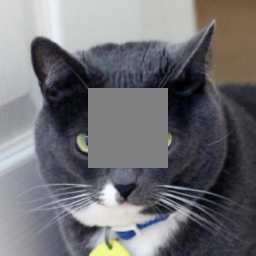}
    \end{subfigure}
    \begin{subfigure}[t]{0.19\textwidth}
        \includegraphics[width=\textwidth]{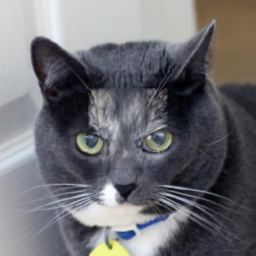}
    \end{subfigure}
    \\
    \begin{subfigure}[t]{0.19\textwidth}
        \includegraphics[width=\textwidth]{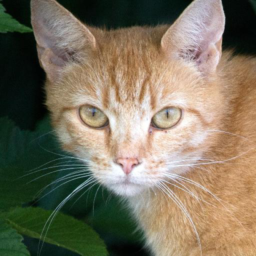}
    \end{subfigure}
    \begin{subfigure}[t]{0.19\textwidth}
        \includegraphics[width=\textwidth]{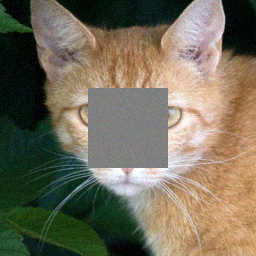}
    \end{subfigure}
    \begin{subfigure}[t]{0.19\textwidth}
        \includegraphics[width=\textwidth]{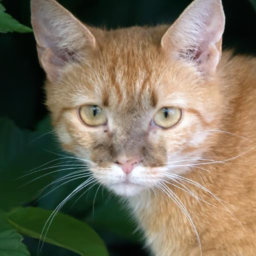}
    \end{subfigure}
        \begin{subfigure}[t]{0.19\textwidth}
        \includegraphics[width=\textwidth]{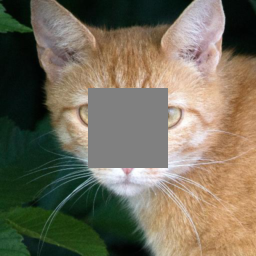}
    \end{subfigure}
    \begin{subfigure}[t]{0.19\textwidth}
        \includegraphics[width=\textwidth]{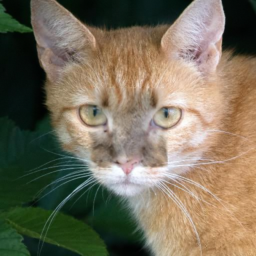}
    \end{subfigure}
    \\
    \begin{subfigure}[t]{0.19\textwidth}
        \includegraphics[width=\textwidth]{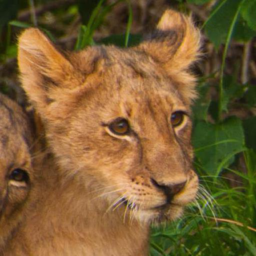}
        \caption{Reference}
    \end{subfigure}
    \begin{subfigure}[t]{0.19\textwidth}
        \includegraphics[width=\textwidth]{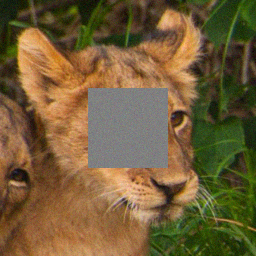}
        \caption{Masked-noisy}
    \end{subfigure}
    \begin{subfigure}[t]{0.19\textwidth}
        \includegraphics[width=\textwidth]{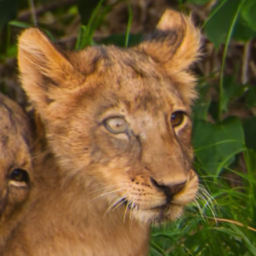}
        \caption{Corrected-noisy}
    \end{subfigure}
        \begin{subfigure}[t]{0.19\textwidth}
        \includegraphics[width=\textwidth]{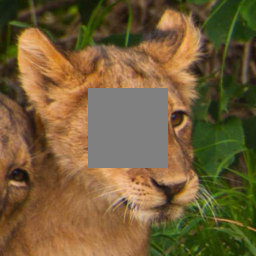}
        \caption{Masked $\sigma_y=0$}
    \end{subfigure}
    \begin{subfigure}[t]{0.19\textwidth}
        \includegraphics[width=\textwidth]{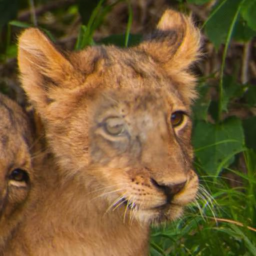}
        \caption{Corrected $\sigma_y=0$}
    \end{subfigure}
    \\
    \caption{Negative results for inpainting with OT-ODE on AFHQ. We can observe artifacts in high-resolution images where the masked region is not inpainted correctly and there are patches in the inpainted region that are semantically incorrect.  The observed artifacts are present in both the noiseless (e) and noisy (c) columns.}
    \label{fig:inpainting-failure-modes-afhq}
\end{figure}

\begin{figure}[!htb]
    \begin{subfigure}[t]{0.19\textwidth}
        \includegraphics[width=\textwidth]{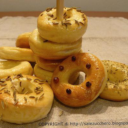}
    \end{subfigure}
    \begin{subfigure}[t]{0.19\textwidth}
        \includegraphics[width=\textwidth]{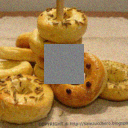}
    \end{subfigure}
    \begin{subfigure}[t]{0.19\textwidth}
        \includegraphics[width=\textwidth]{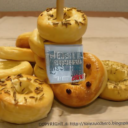}
    \end{subfigure}
        \begin{subfigure}[t]{0.19\textwidth}
        \includegraphics[width=\textwidth]{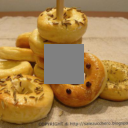}
    \end{subfigure}
    \begin{subfigure}[t]{0.19\textwidth}
        \includegraphics[width=\textwidth]{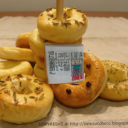}
    \end{subfigure}
    \\
    \begin{subfigure}[t]{0.19\textwidth}
        \includegraphics[width=\textwidth]{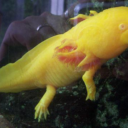}
    \end{subfigure}
    \begin{subfigure}[t]{0.19\textwidth}
        \includegraphics[width=\textwidth]{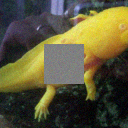}
    \end{subfigure}
    \begin{subfigure}[t]{0.19\textwidth}
        \includegraphics[width=\textwidth]{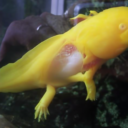}
    \end{subfigure}
        \begin{subfigure}[t]{0.19\textwidth}
        \includegraphics[width=\textwidth]{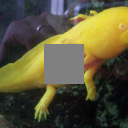}
    \end{subfigure}
    \begin{subfigure}[t]{0.19\textwidth}
        \includegraphics[width=\textwidth]{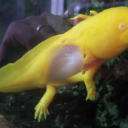}
    \end{subfigure}
    \\
    \begin{subfigure}[t]{0.19\textwidth}
        \includegraphics[width=\textwidth]{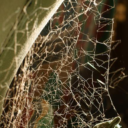}
    \end{subfigure}
    \begin{subfigure}[t]{0.19\textwidth}
        \includegraphics[width=\textwidth]{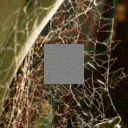}
    \end{subfigure}
    \begin{subfigure}[t]{0.19\textwidth}
        \includegraphics[width=\textwidth]{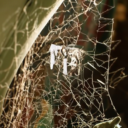}
    \end{subfigure}
        \begin{subfigure}[t]{0.19\textwidth}
        \includegraphics[width=\textwidth]{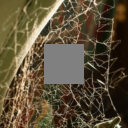}
    \end{subfigure}
    \begin{subfigure}[t]{0.19\textwidth}
        \includegraphics[width=\textwidth]{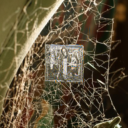}
    \end{subfigure}
    \\
    \begin{subfigure}[t]{0.19\textwidth}
        \includegraphics[width=\textwidth]{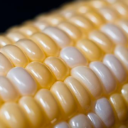}
    \end{subfigure}
    \begin{subfigure}[t]{0.19\textwidth}
        \includegraphics[width=\textwidth]{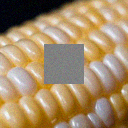}
    \end{subfigure}
    \begin{subfigure}[t]{0.19\textwidth}
        \includegraphics[width=\textwidth]{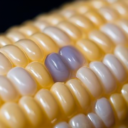}
    \end{subfigure}
        \begin{subfigure}[t]{0.19\textwidth}
        \includegraphics[width=\textwidth]{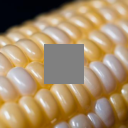}
    \end{subfigure}
    \begin{subfigure}[t]{0.19\textwidth}
        \includegraphics[width=\textwidth]{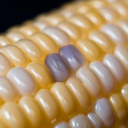}
    \end{subfigure}
    \\
    \begin{subfigure}[t]{0.19\textwidth}
        \includegraphics[width=\textwidth]{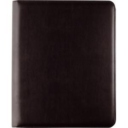}
    \end{subfigure}
    \begin{subfigure}[t]{0.19\textwidth}
        \includegraphics[width=\textwidth]{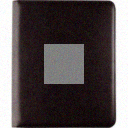}
    \end{subfigure}
    \begin{subfigure}[t]{0.19\textwidth}
        \includegraphics[width=\textwidth]{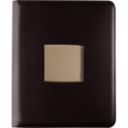}
    \end{subfigure}
        \begin{subfigure}[t]{0.19\textwidth}
        \includegraphics[width=\textwidth]{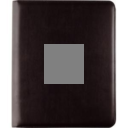}
    \end{subfigure}
    \begin{subfigure}[t]{0.19\textwidth}
        \includegraphics[width=\textwidth]{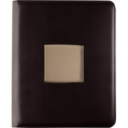}
    \end{subfigure}
    \\
    \begin{subfigure}[t]{0.19\textwidth}
        \includegraphics[width=\textwidth]{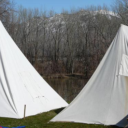}
        \caption{Reference}
    \end{subfigure}
    \begin{subfigure}[t]{0.19\textwidth}
        \includegraphics[width=\textwidth]{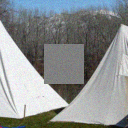}
        \caption{Masked-noisy}
    \end{subfigure}
    \begin{subfigure}[t]{0.19\textwidth}
        \includegraphics[width=\textwidth]{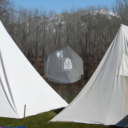}
        \caption{Corrected-noisy}
    \end{subfigure}
        \begin{subfigure}[t]{0.19\textwidth}
        \includegraphics[width=\textwidth]{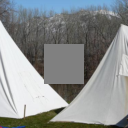}
        \caption{Masked $\sigma_y=0$}
    \end{subfigure}
    \begin{subfigure}[t]{0.19\textwidth}
        \includegraphics[width=\textwidth]{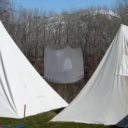}
        \caption{Corrected $\sigma_y=0$}
    \end{subfigure}
    \\
    \caption{Negative results for inpainting with OT-ODE on face-blurred ImageNet-128. We can observe artifacts in high-resolution images where the masked region is not inpainted correctly and there are patches in the inpainted region that are semantically incorrect.  The observed artifacts are present in both the noiseless (e) and noisy (c) columns.}
    \label{fig:inpainting-failure-modes-imagenet-128}
\end{figure}

%% file: sections/nrsd.tex
\section{Noiseless null and range space decomposition}
\label{app:null_space}
When $\sigma_y^2 = 0$, we can produce a vector field approximation with even lower Conditional Flow Matching loss by applying a null-space and range-space decomposition motivated by DDNM~\citep{ddnm}.  In particular, when $\vy = \mA\vx_1$, we have that $\mA^{\dagger}\vy = \mA^{\dagger}\mA\vx_1$ (where $\mA^{\dagger}$ is the pseudo-inverse of $\mA$) and so
\begin{align}
\E_q[\vx_1 | \vx_t, \vy] = \E_q[\mA^{\dagger}\mA \vx_1 + (\mI - \mA^{\dagger}\mA)\vx_1 | \vx_t, \vy] 
= \mA^{\dagger}\vy + (\mI-\mA^{\dagger}\mA)\E_q[\vx_1 | \vx_t, \vy].
\end{align}
So when $\sigma_\vy^2 = 0$, it is only necessary to approximate the second term, as the first term is known through $\vy$.  The regression loss is minimized for the first term automatically and $\widehat{\vx_1}(\vx_t, \vy)$ is only responsible for predicting the second term.

\begin{figure}[!htb]
    \begin{subfigure}[t]{0.19\textwidth}
        \includegraphics[width=\textwidth]{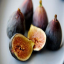}
    \end{subfigure}
    \begin{subfigure}[t]{0.19\textwidth}
        \includegraphics[width=\textwidth]{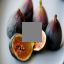}
    \end{subfigure}
    \begin{subfigure}[t]{0.19\textwidth}
        \includegraphics[width=\textwidth]{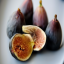}
    \end{subfigure}
    \begin{subfigure}[t]{0.19\textwidth}
        \includegraphics[width=\textwidth]{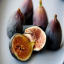}
    \end{subfigure}
    \begin{subfigure}[t]{0.19\textwidth}    
        \includegraphics[width=\textwidth]{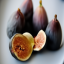}
    \end{subfigure} \\
    \begin{subfigure}[t]{0.19\textwidth}
        \includegraphics[width=\textwidth]{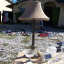}
    \end{subfigure}
    \begin{subfigure}[t]{0.19\textwidth}
        \includegraphics[width=\textwidth]{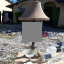}
    \end{subfigure}
    \begin{subfigure}[t]{0.19\textwidth}
        \includegraphics[width=\textwidth]{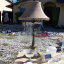}
    \end{subfigure}
    \begin{subfigure}[t]{0.19\textwidth}
        \includegraphics[width=\textwidth]{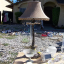}
    \end{subfigure}
    \begin{subfigure}[t]{0.19\textwidth}    
        \includegraphics[width=\textwidth]{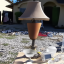}
    \end{subfigure} \\
    \begin{subfigure}[t]{0.19\textwidth}
        \includegraphics[width=\textwidth]{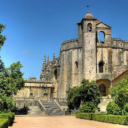}
    \end{subfigure}
    \begin{subfigure}[t]{0.19\textwidth}
        \includegraphics[width=\textwidth]{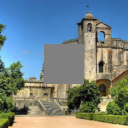}
    \end{subfigure}
    \begin{subfigure}[t]{0.19\textwidth}
        \includegraphics[width=\textwidth]{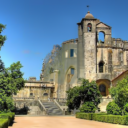}
    \end{subfigure}
    \begin{subfigure}[t]{0.19\textwidth}
        \includegraphics[width=\textwidth]{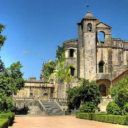}
    \end{subfigure}
    \begin{subfigure}[t]{0.19\textwidth}    
        \includegraphics[width=\textwidth]{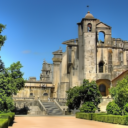}
    \end{subfigure}\\
    \begin{subfigure}[t]{0.19\textwidth}
        \includegraphics[width=\textwidth]{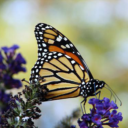}
    \end{subfigure}
    \begin{subfigure}[t]{0.19\textwidth}
        \includegraphics[width=\textwidth]{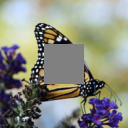}
    \end{subfigure}
    \begin{subfigure}[t]{0.19\textwidth}
        \includegraphics[width=\textwidth]{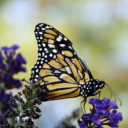}
    \end{subfigure}
    \begin{subfigure}[t]{0.19\textwidth}
        \includegraphics[width=\textwidth]{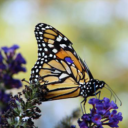}
    \end{subfigure}
    \begin{subfigure}[t]{0.19\textwidth}    
        \includegraphics[width=\textwidth]{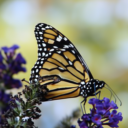}
    \end{subfigure}\\
    \begin{subfigure}[t]{0.19\textwidth}
        \includegraphics[width=\textwidth]{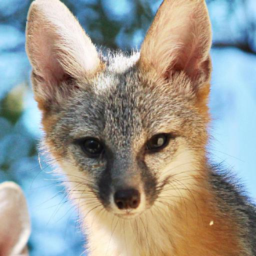}
        \caption{Reference}
    \end{subfigure}
    \begin{subfigure}[t]{0.19\textwidth}
        \includegraphics[width=\textwidth]{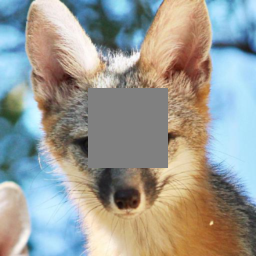}
        \caption{Distorted}
    \end{subfigure}
    \begin{subfigure}[t]{0.19\textwidth}
        \includegraphics[width=\textwidth]{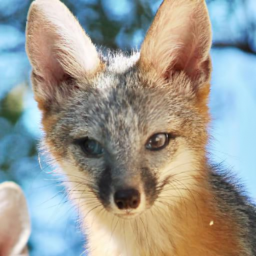}
        \caption{OT-ODE}
    \end{subfigure}
    \begin{subfigure}[t]{0.19\textwidth}
        \includegraphics[width=\textwidth]{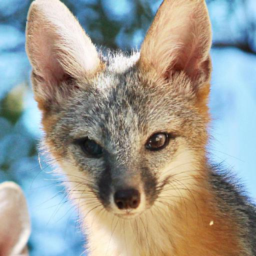}
        \caption{OT-ODE-NRSD}
    \end{subfigure}
    \begin{subfigure}[t]{0.19\textwidth}    
        \includegraphics[width=\textwidth]{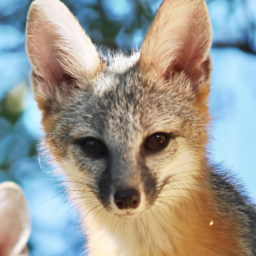}
        \caption{$\Pi$GDM}
    \end{subfigure}
    \caption{Comparison of inpainting (center mask) via OT-ODE sampling with and without null and range space decomposition (NRSD). We use conditional OT model and $\sigma_y=0$ for (\textbf{first and second row}) face-blurred ImageNet-64, (\textbf{third row}) face-blurred ImageNet-128, and (\textbf{fourth row}) AFHQ.}
    \label{fig:ddnm-inpainting-ot-vis}
\end{figure}

\begin{figure}[!htb]
    \begin{subfigure}[t]{0.19\textwidth}
        \includegraphics[width=\textwidth]{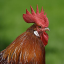}
    \end{subfigure}
    \begin{subfigure}[t]{0.19\textwidth}
        \includegraphics[width=\textwidth]{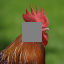}
    \end{subfigure}
    \begin{subfigure}[t]{0.19\textwidth}
        \includegraphics[width=\textwidth]{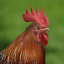}
    \end{subfigure}
    \begin{subfigure}[t]{0.19\textwidth}
        \includegraphics[width=\textwidth]{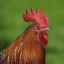}
    \end{subfigure}
    \begin{subfigure}[t]{0.19\textwidth}    
        \includegraphics[width=\textwidth]{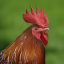}
    \end{subfigure} \\
        \begin{subfigure}[t]{0.19\textwidth}
        \includegraphics[width=\textwidth]{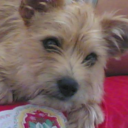}
    \end{subfigure}
    \begin{subfigure}[t]{0.19\textwidth}
        \includegraphics[width=\textwidth]{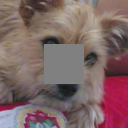}
    \end{subfigure}
    \begin{subfigure}[t]{0.19\textwidth}
        \includegraphics[width=\textwidth]{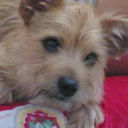}
    \end{subfigure}
    \begin{subfigure}[t]{0.19\textwidth}
        \includegraphics[width=\textwidth]{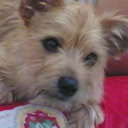}
    \end{subfigure}
    \begin{subfigure}[t]{0.19\textwidth}    
        \includegraphics[width=\textwidth]{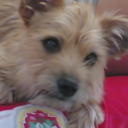}
    \end{subfigure}\\
    \begin{subfigure}[t]{0.19\textwidth}
        \includegraphics[width=\textwidth]{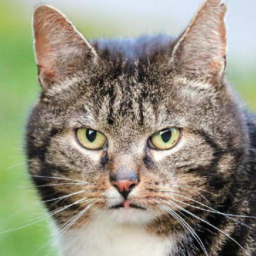}
        \caption{Reference}
    \end{subfigure}
    \begin{subfigure}[t]{0.19\textwidth}
        \includegraphics[width=\textwidth]{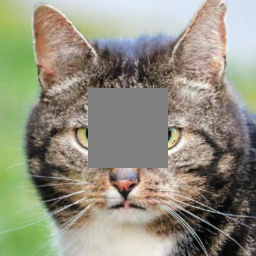}
        \caption{Distorted}
    \end{subfigure}
    \begin{subfigure}[t]{0.19\textwidth}
        \includegraphics[width=\textwidth]{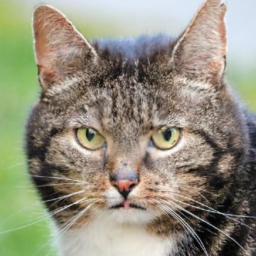}
        \caption{OT-ODE}
    \end{subfigure}
    \begin{subfigure}[t]{0.19\textwidth}
        \includegraphics[width=\textwidth]{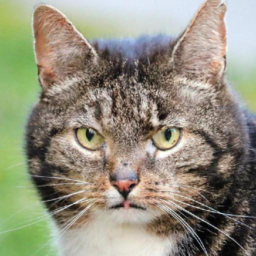}
        \caption{OT-ODE-NRSD}
    \end{subfigure}
    \begin{subfigure}[t]{0.19\textwidth}    
        \includegraphics[width=\textwidth]{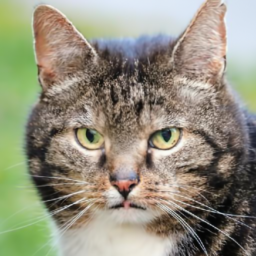}
        \caption{$\Pi$GDM}
    \end{subfigure}
    \caption{Comparison of inpainting (center mask) via OT-ODE sampling with and without null and range space decomposition (NRSD) for (\textbf{first row}) face-blurred ImageNet-64, (\textbf{second row}) face-blurred ImageNet-128, and (\textbf{third row}) AFHQ.  We use VP-SDE model and $\sigma_y=0$.}
    \label{fig:ddnm-inpainting-ddpm-vis}
\end{figure}

\begin{figure}[!htb]
    \begin{subfigure}[t]{0.19\textwidth}
        \includegraphics[width=\textwidth]{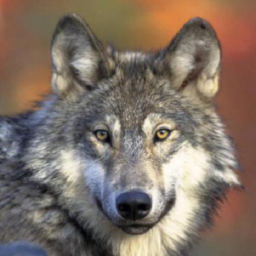}
    \end{subfigure}
    \begin{subfigure}[t]{0.19\textwidth}
        \includegraphics[width=\textwidth]{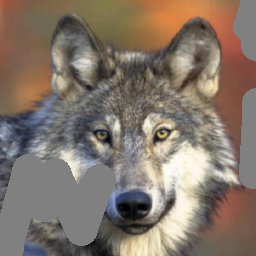}
    \end{subfigure}
    \begin{subfigure}[t]{0.19\textwidth}
        \includegraphics[width=\textwidth]{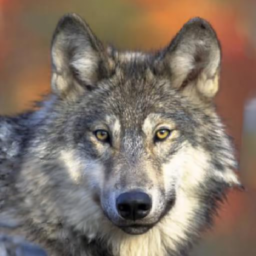}
    \end{subfigure}
    \begin{subfigure}[t]{0.19\textwidth}
        \includegraphics[width=\textwidth]{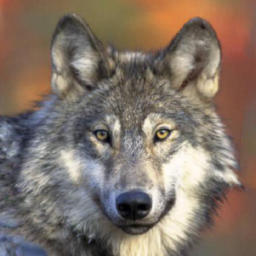}
    \end{subfigure}
    \begin{subfigure}[t]{0.19\textwidth}    
        \includegraphics[width=\textwidth]{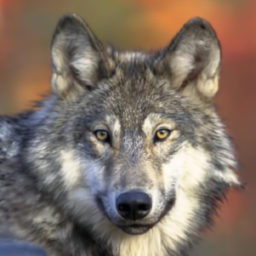}
    \end{subfigure} \\
    \begin{subfigure}[t]{0.19\textwidth}
        \includegraphics[width=\textwidth]{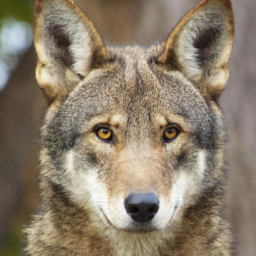}
    \end{subfigure}
    \begin{subfigure}[t]{0.19\textwidth}
        \includegraphics[width=\textwidth]{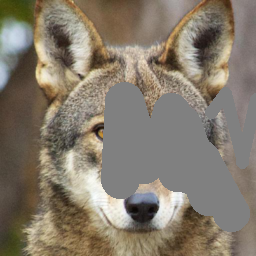}
    \end{subfigure}
    \begin{subfigure}[t]{0.19\textwidth}
        \includegraphics[width=\textwidth]{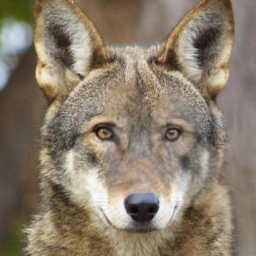}
    \end{subfigure}
    \begin{subfigure}[t]{0.19\textwidth}
        \includegraphics[width=\textwidth]{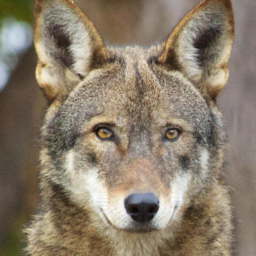}
    \end{subfigure}
    \begin{subfigure}[t]{0.19\textwidth}    
        \includegraphics[width=\textwidth]{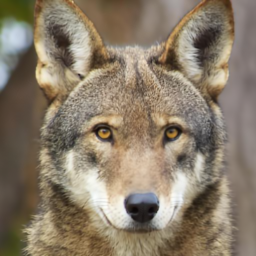}
    \end{subfigure} \\
    \begin{subfigure}[t]{0.19\textwidth}
        \includegraphics[width=\textwidth]{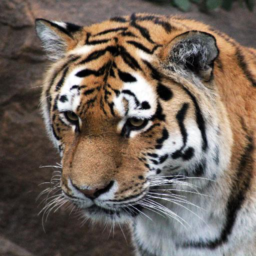}
        \caption{Reference}
    \end{subfigure}
    \begin{subfigure}[t]{0.19\textwidth}
        \includegraphics[width=\textwidth]{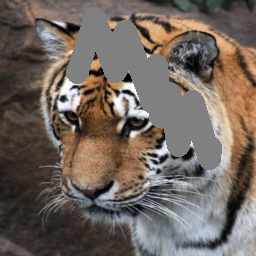}
        \caption{Distorted}
    \end{subfigure}
    \begin{subfigure}[t]{0.19\textwidth}
        \includegraphics[width=\textwidth]{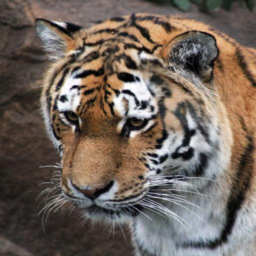}
        \caption{OT-ODE}
    \end{subfigure}
    \begin{subfigure}[t]{0.19\textwidth}
        \includegraphics[width=\textwidth]{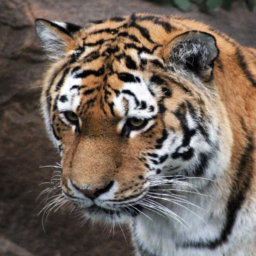}
        \caption{OT-ODE-NRSD}
    \end{subfigure}
    \begin{subfigure}[t]{0.19\textwidth}    
        \includegraphics[width=\textwidth]{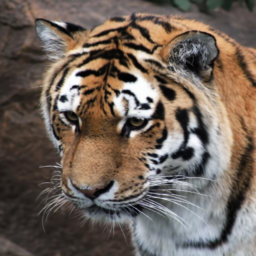}
        \caption{$\Pi$GDM}
    \end{subfigure}
    \caption{Comparison of inpainting (free-form mask) via OT-ODE sampling with and without null and range space decomposition (NRSD) for AFHQ. We use conditional OT model and $\sigma_y=0$.}
    \label{fig:ddnm-inpainting-freeform-ot-vis}
\end{figure}

In our experiments, we find that null space decomposition helps in inpainting but not other measurements. We summarize the results in~\cref{tab:nrsd-imagenet-64-sigma-0-ip,tab:nrsd-imagenet-64-sigma-0-sr-gb,tab:nrsd-imagenet-128-sigma-0-sr-gb,tab:nrsd-imagenet-128-sigma-0-ip,tab:nrsd-afhq-sigma-0-gb-dn,tab:nrsd-afhq-sigma-0-ip} and show qualitative results for inpainting in~\cref{fig:ddnm-inpainting-ddpm-vis,fig:ddnm-inpainting-freeform-ot-vis,fig:ddnm-inpainting-ot-vis}.

\begin{table*}[thbp!]
    \centering
    \caption{Comparison of performance OT-ODE sampling and OT-ODE sampling with null and range space decomposition (NRSD) on face-blurred ImageNet-$64\times64$. For inpainting, OT-ODE sampling with null and range space decomposition outperforms simple OT-ODE sampling.}
    \label{tab:nrsd-imagenet-64-sigma-0-ip}
    \resizebox{0.65\textwidth}{!}{
    \begin{tabular}{ccccccc}
    \toprule 
    \multirow{3}{*}{Model} & \multirow{3}{*}{Inference} & \multirow{3}{*}{NFEs $\downarrow$} &\multicolumn{4}{c}{Inpainting-\textit{Center}, $\sigma_y=0$} \\
    \cmidrule(lr){4-7}
    & & & FID $\downarrow$ & LPIPS $\downarrow$ & PSNR $\uparrow$ &  SSIM $\uparrow$ \\
    \midrule
    OT & OT-ODE  & 80 & 4.94 & 0.080 & 37.42 & 0.885 \\
    OT & OT-ODE-NRSD  & 80 & \textbf{3.84} & \textbf{0.072} & \textbf{38.23} & \textbf{0.888} \\
    OT & VP-ODE & 80 & 7.85 & 0.120 & 34.24 & 0.858 \\
    \midrule
    VP-SDE & OT-ODE  & 80 & 4.85 & 0.079 & 37.64 & 0.887 \\
    VP-SDE & OT-ODE-NRSD  & 80 & \textbf{3.77} & \textbf{0.072} & \textbf{38.24} & \textbf{0.888} \\
    VP-SDE & VP-ODE & 80 & 7.21 & 0.117 & 34.33 & 0.860 \\
    \bottomrule
    \end{tabular}
    }
\end{table*}

\begin{table*}[thbp!]
    \centering
    \caption{Comparison of performance OT-ODE sampling and OT-ODE sampling with null and range space decomposition (NRSD) on face-blurred ImageNet-$64\times64$. For tasks like super-resolution and Gaussian deblurring, OT-ODE sampling without null and range space decomposition outperforms other methods.}
    \label{tab:nrsd-imagenet-64-sigma-0-sr-gb}
    \resizebox{\textwidth}{!}{
    \begin{tabular}{ccccccccccc}
    \toprule 
    \multirow{3}{*}{Model} & \multirow{3}{*}{Inference} & \multirow{3}{*}{NFEs $\downarrow$} &\multicolumn{4}{c}{SR 2$\times$, $\sigma_y=0$} &\multicolumn{4}{c}{Gaussian deblur, $\sigma_y=0$} \\
    \cmidrule(lr){4-7}\cmidrule(lr){8-11}
    & & & FID $\downarrow$ & LPIPS $\downarrow$ & PSNR $\uparrow$ &  SSIM $\uparrow$ & FID $\downarrow$ & LPIPS $\downarrow$ & PSNR $\uparrow$ &  SSIM $\uparrow$\\
    \midrule
    OT & OT-ODE & 80 &  \textbf{6.46} & \textbf{0.119} & \textbf{31.59} & \textbf{0.839} & \textbf{2.59} & \textbf{0.038} & \textbf{35.31} & \textbf{0.961} \\
    OT & OT-ODE-NRSD & 80 &  7.37 & 0.134 & 31.05 & 0.799 & 3.05 & 0.044 & 35.19 & 0.956 \\
    OT & VP-ODE & 80 & 8.29 & 0.147 & 31.20 & 0.817 & 6.13 & 0.083 & 33.31 & 0.929  \\
    \midrule
    VP-SDE & OT-ODE  & 80 & \textbf{6.32} & \textbf{0.118} & \textbf{31.60} & \textbf{0.839} & \textbf{2.61} & \textbf{0.037} & \textbf{35.45} & \textbf{0.963} \\
    VP-SDE & OT-ODE-NRSD & 80 &  7.13 & 0.133 & 31.06 & 0.798 & 2.99 & 0.044 & 35.24 & 0.956 \\
    VP-SDE & VP-ODE  & 80 & 7.76 & 0.145 & 31.21 & 0.817 & 5.68 & 0.080 & 33.37 & 0.931 \\
    \bottomrule
    \end{tabular}
    }
\end{table*}

\begin{table*}[thbp!]
    \centering
    \caption{Comparison of performance OT-ODE sampling and OT-ODE sampling with null and range space decomposition (NRSD) on face-blurred ImageNet-$128\times128$.}
    \label{tab:nrsd-imagenet-128-sigma-0-ip}
    \resizebox{\textwidth}{!}{
    \begin{tabular}{ccccccccccc}
    \toprule 
    \multirow{3}{*}{Model} & \multirow{3}{*}{Inference} & \multirow{3}{*}{NFEs $\downarrow$} &\multicolumn{4}{c}{SR 2$\times$, $\sigma_y=0$} &\multicolumn{4}{c}{Gaussian deblur, $\sigma_y=0$} \\
    \cmidrule(lr){4-7}\cmidrule(lr){8-11}
    & & & FID $\downarrow$ & LPIPS $\downarrow$ & PSNR $\uparrow$ &  SSIM $\uparrow$ & FID $\downarrow$ & LPIPS $\downarrow$ & PSNR $\uparrow$ &  SSIM $\uparrow$\\
    \midrule
    OT & OT-ODE & 70 & 4.46 & \textbf{0.097} & \textbf{33.88} & \textbf{0.903} & 2.09 & 0.048 & 37.49 & 0.961 \\
    OT & OT-ODE-NRSD & 70 & \textbf{3.62} & 0.099 & 33.24 & 0.876 & \textbf{1.42} & \textbf{0.036} & \textbf{38.35} & \textbf{0.969} \\
    OT & VP-ODE & 70 & 7.69 & 0.144 & 32.93 & 0.871 & 6.02 & 0.108 & 34.73 & 0.925\\
    \midrule
    VP-SDE & OT-ODE & 70 & 4.62 & \textbf{0.096} & \textbf{33.95} & \textbf{0.906} & 2.26 & 0.046 & 37.79 & 0.967 \\
    VP-SDE & OT-ODE-NRSD & 70 & \textbf{3.44} & 0.098 & 33.28 & 0.877 & \textbf{1.36} & \textbf{0.035} & \textbf{38.44} & \textbf{0.969} \\
    VP-SDE & VP-ODE & 70 & 7.91 & 0.144 & 32.87 & 0.869 & 5.64 & 0.105 & 34.81 & 0.928  \\
    \bottomrule
    \end{tabular}
    }
\end{table*}

\begin{table*}[th!]
    \centering
    \caption{Comparison of performance OT-ODE sampling and OT-ODE sampling with null and range space decomposition (NRSD) on face-blurred ImageNet-$128\times128$}
    \label{tab:nrsd-imagenet-128-sigma-0-sr-gb}
    \resizebox{0.65\textwidth}{!}{
    \begin{tabular}{ccccccc}
    \toprule 
    \multirow{3}{*}{Model} & \multirow{3}{*}{Inference} & \multirow{3}{*}{NFEs $\downarrow$} &\multicolumn{4}{c}{Inpainting-\textit{Center}, $\sigma_y=0$} \\
    \cmidrule(lr){4-7}
    & & & FID $\downarrow$ & LPIPS $\downarrow$ & PSNR $\uparrow$ &  SSIM $\uparrow$ \\
    \midrule
    OT & OT-ODE & 70 & 5.88 & 0.095 & 37.06 & 0.894 \\
    OT & OT-ODE-NRSD & 70 & \textbf{3.95} & \textbf{0.074} & \textbf{38.27} & \textbf{0.906} \\
    OT & VP-ODE & 70 & 8.63 & 0.144 & 34.48 & 0.864 \\
    \midrule
    VP-SDE & OT-ODE & 70 & 5.93 & 0.094 & 37.31 & 0.898 \\
    VP-SDE & OT-ODE-NRSD & 70 & \textbf{3.84} & \textbf{0.073} & \textbf{38.27} & \textbf{0.906} \\
    VP-SDE & VP-ODE & 70 & 8.08 & 0.142 & 34.55 & 0.865 \\
    \bottomrule
    \end{tabular}
    }
\end{table*}

\begin{table*}[th!]
    \centering
    \caption{Comparison of performance OT-ODE sampling and OT-ODE sampling with null and range space decomposition (NRSD) on AFHQ-$256\times256$}
    \label{tab:nrsd-afhq-sigma-0-gb-dn}
    \resizebox{\textwidth}{!}{
    \begin{tabular}{ccccccccccc}
    \toprule 
    \multirow{3}{*}{Model} & \multirow{3}{*}{Inference} & \multirow{3}{*}{NFEs $\downarrow$} &\multicolumn{4}{c}{SR 4$\times$, $\sigma_y=0$} &\multicolumn{4}{c}{Gaussian deblur, $\sigma_y=0$} \\
    \cmidrule(lr){4-7}\cmidrule(lr){8-11}
    & & & FID $\downarrow$ & LPIPS $\downarrow$ & PSNR $\uparrow$ &  SSIM $\uparrow$ & FID $\downarrow$ & LPIPS $\downarrow$ & PSNR $\uparrow$ &  SSIM $\uparrow$\\
    \midrule
    OT & OT-ODE & 100 & 5.75 & \textbf{0.169} & \textbf{32.25} & \textbf{0.792} & \textbf{6.63} & \textbf{0.213} & \textbf{31.29} & \textbf{0.722} \\
    OT & OT-ODE-NRSD & 100 &  \textbf{5.73} & 0.179 & 31.69 & 0.753 & 7.32 & 0.237 & 30.72 & 0.665 \\
    OT & VP-ODE & 100 & 6.14 & 0.194 & 31.93 & 0.773 & 7.38 & 0.231 & 31.10 & 0.705 \\
    \midrule
    VP-SDE & OT-ODE & 100 & \textbf{6.58} & \textbf{0.178} & \textbf{32.18} & \textbf{0.789} & \textbf{8.24} & \textbf{0.226} & \textbf{31.21} & \textbf{0.717} \\
    VP-SDE & OT-ODE-NRSD & 100 & 6.99 & 0.195 & 31.65 & 0.752 & 10.19 & 0.255 & 30.66 & 0.662 \\
    VP-SDE & VP-ODE & 100 & 8.00 & 0.225 & 31.48 & 0.742 & 9.19 & 0.252 & 30.91 & 0.688 \\
    \bottomrule
    \end{tabular}
    }
\end{table*}

\begin{table*}[th!]
    \centering
    \caption{Comparison of performance OT-ODE sampling and OT-ODE sampling with null and range space decomposition (NRSD) on AFHQ-$256\times256$}
    \label{tab:nrsd-afhq-sigma-0-ip}
    \resizebox{\textwidth}{!}{
    \begin{tabular}{ccccccccccc}
    \toprule 
    \multirow{3}{*}{Model} & \multirow{3}{*}{Inference} & \multirow{3}{*}{NFEs $\downarrow$} &\multicolumn{4}{c}{Inpainting-\textit{Center}, $\sigma_y=0$} &\multicolumn{4}{c}{Inpainting-\textit{Free-form}, $\sigma_y=0$} \\
    \cmidrule(lr){4-7}\cmidrule(lr){8-11}
    & & & FID $\downarrow$ & LPIPS $\downarrow$ & PSNR $\uparrow$ &  SSIM $\uparrow$ & FID $\downarrow$ & LPIPS $\downarrow$ & PSNR $\uparrow$ &  SSIM $\uparrow$\\
    \midrule
    OT & OT-ODE & 100 & 8.87 & 0.061 & 37.45 & \textbf{0.921} & 4.98 & 0.097 & 36.15 & 0.889 \\
    OT & OT-ODE-NRSD & 100 & \textbf{7.95} & \textbf{0.046} & \textbf{38.01} & \textbf{0.921} & \textbf{4.12} & \textbf{0.083} & \textbf{36.62} & \textbf{0.890} \\
    OT & VP-ODE & 100 & 9.18 & 0.106 & 35.63 & 0.898 & 6.92 & 0.135 & 34.72 & 0.869 \\
    \midrule
    VP-SDE & OT-ODE & 100 & \textbf{9.95} & 0.064 & 37.49 & \textbf{0.918} &  5.39 & 0.099 & 36.15 & \textbf{0.887} \\
    VP-SDE & OT-ODE-NRSD & 100 & 10.96 & \textbf{0.052} & \textbf{37.95} & 0.916 & \textbf{4.87} & \textbf{0.089} & \textbf{36.52} & 0.884 \\
    VP-SDE & VP-ODE & 100 & 10.50 & 0.112 & 35.59 & 0.893 & 7.36 & 0.139 & 34.65 & 0.865 \\
    \bottomrule
    \end{tabular}
    }
\end{table*}

%% file: sections/pigdm_impl.tex
\subsection{$\Pi$GDM}\label{sec:pigdm-impl-details}
\paragraph{Implementation details.} We closely follow the official code available on github while implementing $\Pi$GDM. For noisy case, we closely follow the Algorithm 1 in the appendix of \citet{pigdm}. We use adaptive weighted guidance for both noiseless and noisy cases as in the original work. We always use uniform spacing while iterating the timestep over $100$ steps. We use ascending time from $0$ to $1$. Note that the original paper uses descending time from $T$ to $0$. According to the notational convention used in this paper, this is equivalent to ascending time from $0$ to $1$. For the choice of $r_t^2$, we consider the values derived from both variance exploding formulation and variance preserving formulation.

\paragraph{Value of $r_t^2$.} $\Pi$GDM sets the value of $r_t^2 = \frac{\sigma_{1-t}^2}{1 + \sigma_{1-t}^2}$ for VE-SDE, where $q(\vx_t|\vx_1) = \mathcal{N}(\vx_1, \sigma_{1-t}^2\mI)$. We can follow the same procedure as outlined in~\citet{pigdm}, and solve for $r_t^2$ in closed form for VP-SDE. We know for that VP-SDE, $q(\vx_t|\vx_1) = \mathcal{N}(\alpha_{1-t} \vx_1,(1 - \alpha_{1-t}^2)\mI)$, where $\alpha_t = e^{-\frac{1}{2}T(t)}$, $T(t) = \int_0^t \beta(s) ds$, and $\beta(s)$ is the noise scale function. Using \eqref{eqn:r_t_sqr_flow_value} for VP-SDE gives $r_t^2 = 1 - \alpha_{1-t}^2$. We can also obtain an alternate $r_t^2$ by plugging in value of $\sigma_t^2$ for VP-SDE into the expression of $r_t^2$ derived for VE-SDE, which evaluates to $r_t^2 = \frac{1 - \alpha_{1-t}^2}{2 - \alpha_{1-t}^2}$. Empirically, we find that $r_t^2$ for VE-SDE marginally outperforms VP-SDE. We report performance of $\Pi$GDM with both choices of $r_t^2$ in~\cref{tab:pigdm-ve-vp-imagenet-64-sigma-0.05,tab:pigdm-ve-vp-imagenet-128-sigma-0.05,tab:pigdm-ve-vp-afhq-sigma-0.05}. 

\begin{table*}[thb!]
    \centering
    \caption{Relative performance of $\Pi$GDM on face-blurred ImageNet-$64$ with VE and VP derived $r_t^2$ with $\sigma_y=0.05$}
    \label{tab:pigdm-ve-vp-imagenet-64-sigma-0.05}
    \resizebox{\textwidth}{!}{
    \begin{tabular}{cccccccccc}
    \toprule 
    \multirow{3}{*}{Measurement} & \multirow{3}{*}{Model} &\multicolumn{4}{c}{VP} &\multicolumn{4}{c}{VE} \\
    \cmidrule(lr){3-6}\cmidrule(lr){7-10}
    & & FID $\downarrow$ & LPIPS $\downarrow$ & PSNR $\uparrow$ &  SSIM $\uparrow$ & FID $\downarrow$ & LPIPS $\downarrow$ & PSNR $\uparrow$ &  SSIM $\uparrow$ \\
    \midrule
    SR 2$\times$ & OT & 6.52 & 0.168 & 30.54 & 0.753 & 5.91 & 0.160 & 30.60 & 0.762 \\
    Gaussian deblur & OT & 55.19 & 0.374 & 28.74 & 0.516 & 39.36 & 0.326 & 29.00 & 0.572 \\
    Inpainting-\textit{Center} & OT & 9.25 & 0.111 & 34.13 & 0.863 & 8.70 & 0.109 & 34.17 & 0.864 \\
    Denoising & OT & 16.59 & 0.102 & 34.60 & 0.906 & 16.44 & 0.101 & 34.64 & 0.907 \\
    \midrule
    SR 2$\times$ & VP-SDE & 6.84 & 0.174 & 30.48 & 0.743 & 6.11 & 0.166 & 30.54 & 0.753 \\
    Gaussian deblur & VP-SDE & 54.77 & 0.376 & 28.74 & 0.511 & 39.14 & 0.329 & 28.99 & 0.567 \\
    Inpainting-\textit{Center} & VP-SDE & 9.75 & 0.113 & 34.03 & 0.860 & 9.36 & 0.112 & 34.06 & 0.862 \\
    Denoising & VP-SDE & 17.19 & 0.107 & 34.25 & 0.901 & 15.54 & 0.102 & 34.41 & 0.906  \\
    \bottomrule
    \end{tabular}
    }
\end{table*}

\begin{table*}[thb!]
    \centering
    \caption{Relative performance of $\Pi$GDM on face-blurred ImageNet-$128$ with VE and VP derived $r_t^2$ with $\sigma_y=0.05$}
    \label{tab:pigdm-ve-vp-imagenet-128-sigma-0.05}
    \resizebox{\textwidth}{!}{
    \begin{tabular}{cccccccccc}
    \toprule 
    \multirow{3}{*}{Measurement} & \multirow{3}{*}{Model} &\multicolumn{4}{c}{VP} &\multicolumn{4}{c}{VE} \\
    \cmidrule(lr){3-6}\cmidrule(lr){7-10}
    & & FID $\downarrow$ & LPIPS $\downarrow$ & PSNR $\uparrow$ &  SSIM $\uparrow$ & FID $\downarrow$ & LPIPS $\downarrow$ & PSNR $\uparrow$ &  SSIM $\uparrow$ \\
    \midrule
    SR 2$\times$ & OT & 4.38 & 0.148 & 32.07 & 0.831 & 4.26 & 0.145 & 32.12 & 0.834 \\
    Gaussian deblur & OT & 30.30 & 0.328 & 29.96 & 0.606 & 22.42 & 0.296 & 30.17 & 0.642 \\
    Inpainting-\textit{Center} & OT &  7.99 & 0.122 & 34.57 & 0.867 & 7.64 & 0.120 & 34.61 & 0.869 \\
    Denoising & OT & 9.60 & 0.107 & 35.11 & 0.903 & 9.30 & 0.104 & 35.21 & 0.906 \\
    \midrule
    SR 2$\times$ & VP-SDE & 7.55 & 0.183 & 31.61 & 0.785 & 6.14 & 0.168 & 31.79 & 0.803 \\
    Gaussian deblur & VP-SDE & 55.61 & 0.463 & 28.57 & 0.414 & 41.69 & 0.404 & 28.98 & 0.493 \\
    Inpainting-\textit{Center} & VP-SDE & 9.75 & 0.130 & 34.45 & 0.858 & 9.46 & 0.129 & 34.49 & 0.859 \\
    Denoising & VP-SDE & 10.69 & 0.124 & 34.72 & 0.882 & 10.11 & 0.119 & 34.92 & 0.886 \\
    \bottomrule
    \end{tabular}
    }
\end{table*}
\begin{table*}[thb!]
    \centering
    \caption{Relative performance of $\Pi$GDM on AFHQ with VE and VP derived $r_t^2$ with $\sigma_y=0.05$}
    \label{tab:pigdm-ve-vp-afhq-sigma-0.05}
    \resizebox{\textwidth}{!}{
    \begin{tabular}{cccccccccc}
    \toprule 
    \multirow{3}{*}{Measurement} & \multirow{3}{*}{Model} &\multicolumn{4}{c}{VP} &\multicolumn{4}{c}{VE} \\
    \cmidrule(lr){3-6}\cmidrule(lr){7-10}
    & & FID $\downarrow$ & LPIPS $\downarrow$ & PSNR $\uparrow$ &  SSIM $\uparrow$ & FID $\downarrow$ & LPIPS $\downarrow$ & PSNR $\uparrow$ &  SSIM $\uparrow$ \\
    \midrule
    SR $4\times$ & OT & 12.69 & 0.285 & 30.18 & 0.665 & 12.31 & 0.282 & 30.23 & 0.672 \\
    Gaussian deblur & OT & 24.60 & 0.383 & 28.93 & 0.429 & 19.66 & 0.355 & 29.16 & 0.475 \\
    Inpainting-\textit{Center} & OT & 19.09 & 0.153 & 34.20 & 0.855 & 16.51 & 0.145 & 34.40 & 0.863 \\
    Denoising & OT & 11.20 & 0.159 & 34.49 & 0.876 & 10.92 & 0.153 & 34.78 & 0.883 \\
    \midrule
    SR $4\times$ & VP-SDE & 77.49 & 0.469 & 29.34 & 0.469 & 54.12 & 0.413 & 29.73 & 0.549 \\
    Gaussian deblur & VP-SDE & 116.42 & 0.535 & 28.49 & 0.313 & 95.09 & 0.493 & 28.74 & 0.368 \\
    Inpainting-\textit{Center} & VP-SDE & 57.46 & 0.239 & 32.40 & 0.773 & 56.86 & 0.238 & 32.42 & 0.775 \\
    Denoising & VP-SDE & 81.15 & 0.451 & 29.62 & 0.639 & 35.33 & 0.278 & 31.72 & 0.776 \\
    \bottomrule
    \end{tabular}
    }
\end{table*}

\begin{figure}[!htb]
    \centering
    \begin{minipage}[b]{\textwidth}
        \includegraphics[width=0.24\textwidth]{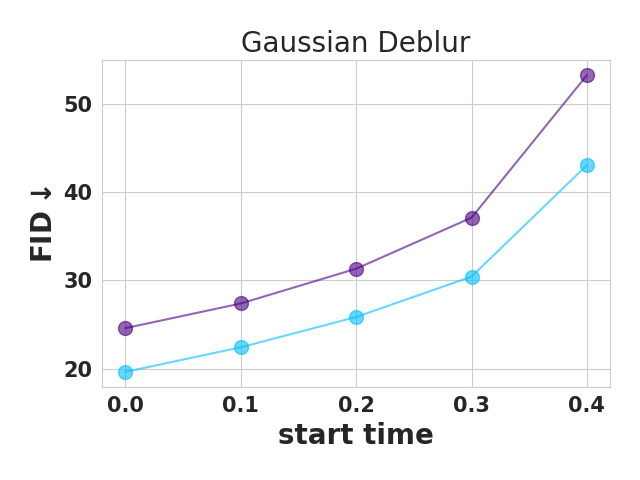}
        \includegraphics[width=0.24\textwidth]{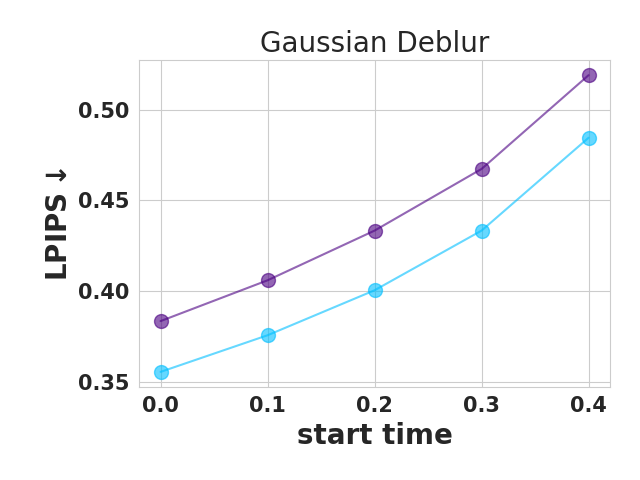}
        \includegraphics[width=0.24\textwidth]{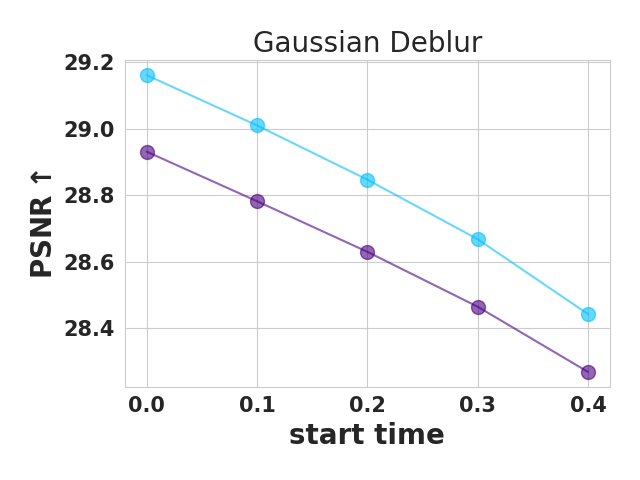}
        \includegraphics[width=0.24\textwidth]{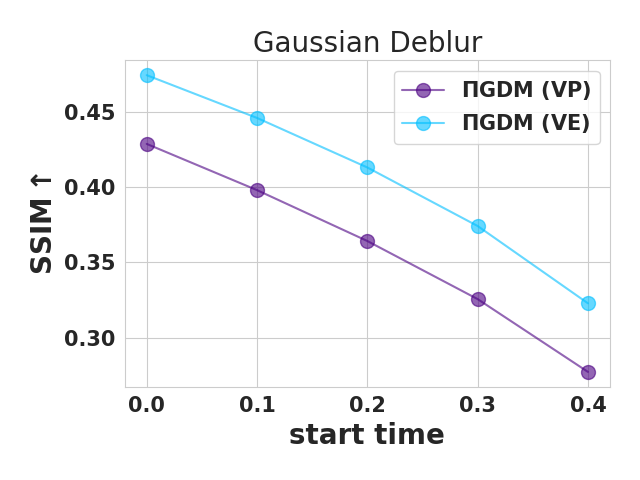}
    \end{minipage}
    \begin{minipage}[b]{\textwidth}
        \includegraphics[width=0.24\textwidth]{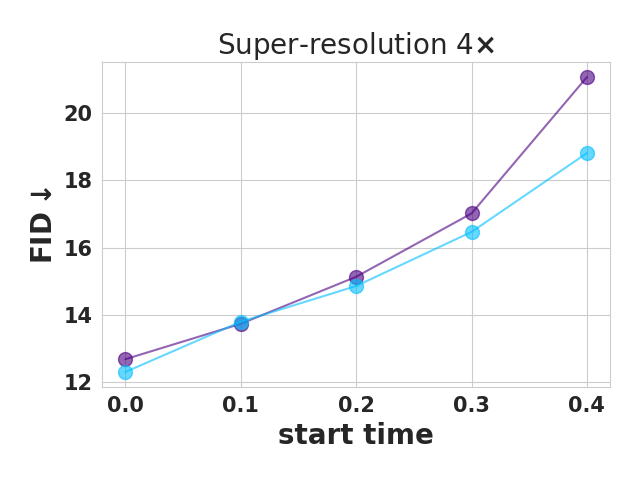}
        \includegraphics[width=0.24\textwidth]{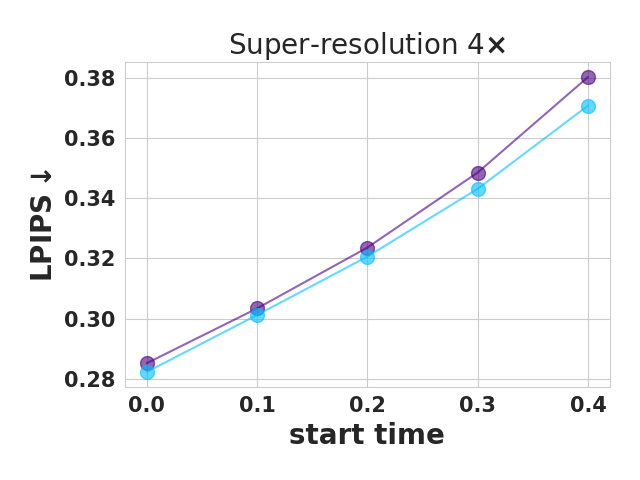}
        \includegraphics[width=0.24\textwidth]{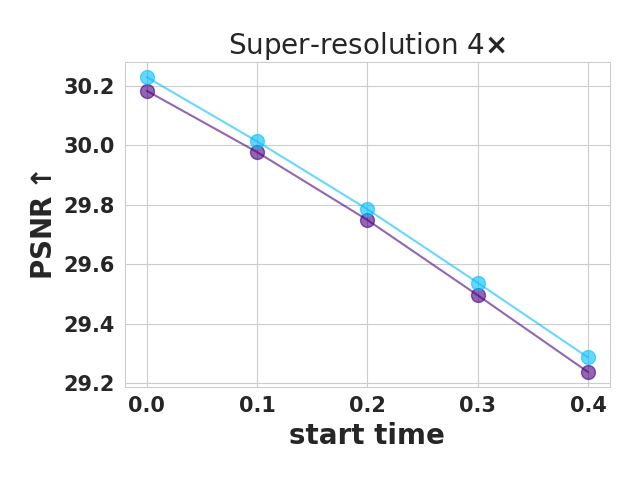}
        \includegraphics[width=0.24\textwidth]{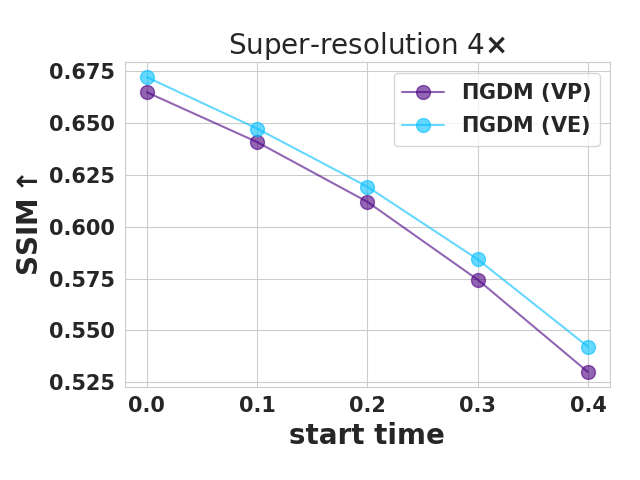}
    \end{minipage}
    \begin{minipage}[b]{\textwidth}
        \includegraphics[width=0.24\textwidth]{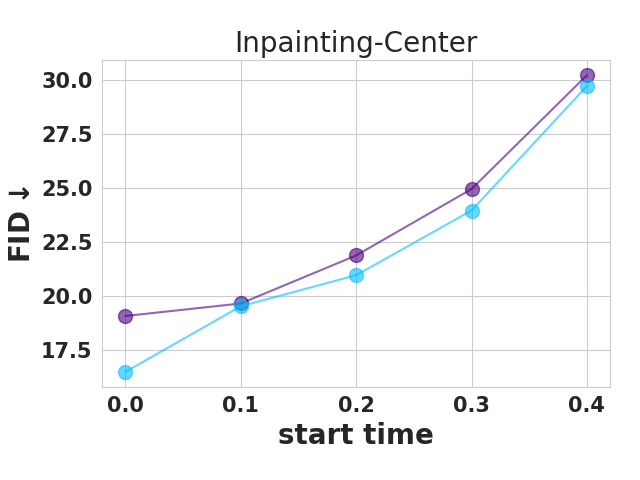}
        \includegraphics[width=0.24\textwidth]{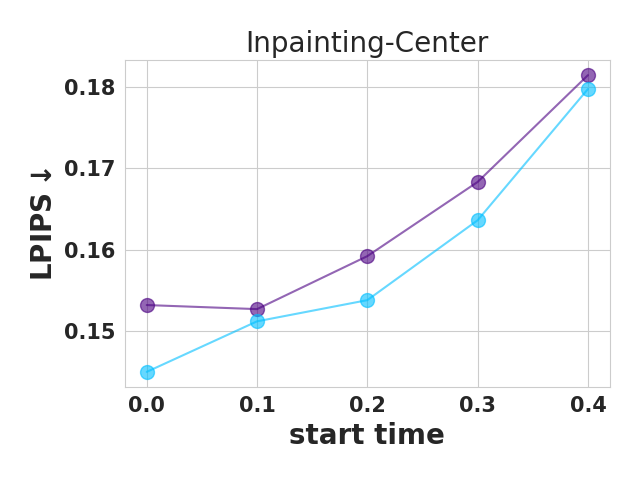}
        \includegraphics[width=0.24\textwidth]{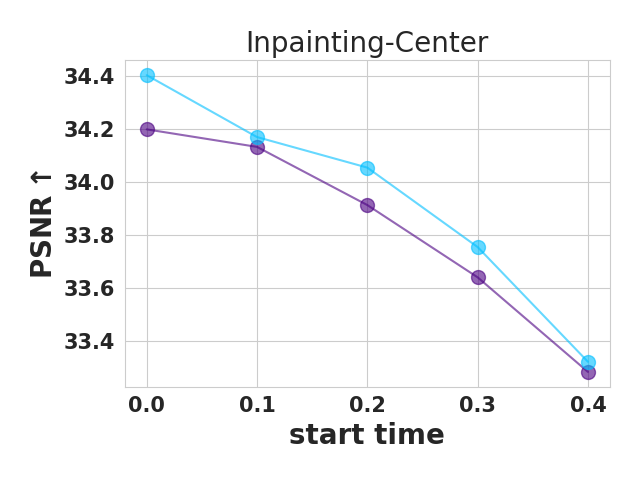}
        \includegraphics[width=0.24\textwidth]{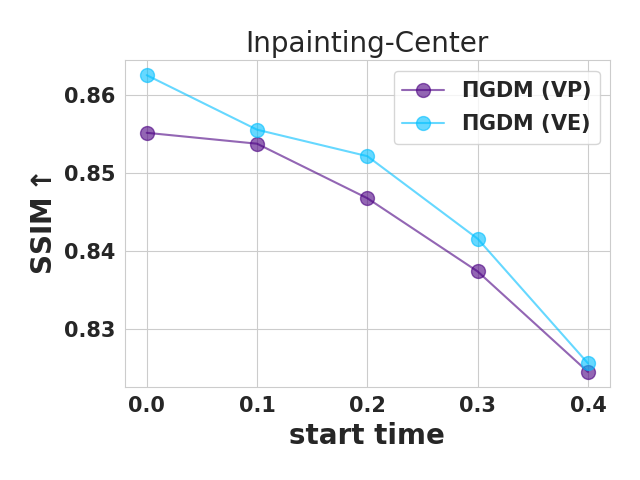}
    \end{minipage}
    \caption{Variation in performance $\Pi$GDM sampling with variation in start times on AFHQ dataset. We use pretrained conditional OT model and set $\sigma_y=0.05$. We observe similar trends with VP-SDE checkpoint. We plot metrics for both choices of  $r_t^2$ that can be derived from variance preserving and variance exploding formulations.}
    \label{fig:pigdm-st-time-ot-noisy-0.05}
\end{figure}

\paragraph{Choice of starting time.} For OT-ODE sampling and VP-ODE sampling, we observe that starting at time $t>0$ improves the performance. We therefore perform an ablation study on $\Pi$GDM baseline, and vary the start time to verify whether starting at $t>0$ helps to improve the performance. We plot the metrics for three different measurements in ~\cref{fig:pigdm-st-time-ot-noisy-0.05}. We observe that starting later at time $t>0$ consistently leads to worse performance compared to starting at time $t=0$. Therefore, for all our experiments with $\Pi$GDM, we always start at time $t=0$.

%% file: sections/reddiff_impl.tex
\subsection{RED-Diff}\label{sec:reddiff-impl-details}

\paragraph{Implementation details.} We use VP-SDE model for all experiments with RED-Diff. We closely follow  the official code available on github while implementing RED-Diff. Similar to \citep{reddiff}, we always use uniform spacing while iterating the timestep over $1000$ steps. We use ascending time from $0$ to $1$. Note that the original paper uses descending time from $T$ to $0$. According to the notational convention used in this paper, this is equivalent to ascending time from $0$ to $1$. We use Adam optimizer and use the momentum pair $(0.9, 0.99)$ similar to the original work. Further, we use initial learning rate of $0.1$ for AFHQ and ImageNet-128, as used in the original work, and learning rate of $0.01$ for ImageNet-64. We use batch size of $1$ for all the experiments. Finally, we extensively tuned the regularization hyperparameter $\lambda$ to find the value that results in optimal performance across all metrics. We summarize the results of our experiments in \cref{tab:reddiff-imagenet-64-sigma-0.05-sr-gb,tab:reddiff-imagenet-64-sigma-0-sr-gb,tab:reddiff-imagenet-128-sigma-0.05-sr-gb,tab:reddiff-imagenet-128-sigma-0-sr-gb,tab:reddiff-afhq-sigma-0.05-sr-gb,tab:reddiff-afhq-sigma-0-sr-gb}. We note that more extensive tuning may be able to find better performing hyperparameters but this goes against the intent of a training-free algorithm.

\begin{table*}[thb!]
    \centering
    \caption{Hyperparameter search for RED-Diff on face-blurred 
 ImageNet-$64\times64$ with $\sigma_y=0.05$. We use learning rate of $0.01$.}
    \label{tab:reddiff-imagenet-64-sigma-0.05-sr-gb}
    \resizebox{0.8\textwidth}{!}{
    \begin{tabular}{ccccccccc}
    \toprule 
    \multirow{3}{*}{$\lambda$} &\multicolumn{4}{c}{SR $2\times$, $\sigma_y=0.05$} &\multicolumn{4}{c}{Gaussian deblur, $\sigma_y=0.05$} \\
    \cmidrule(lr){2-5}\cmidrule(lr){6-9}
    & FID $\downarrow$ & LPIPS $\downarrow$ & PSNR $\uparrow$ &  SSIM $\uparrow$ & FID $\downarrow$ & LPIPS $\downarrow$ & PSNR $\uparrow$ &  SSIM $\uparrow$ \\
    \midrule
    0.1 &  34.09 & 0.224 & 30.12 & 0.798 & \textbf{46.76} & 0.254 & 29.29 & 0.715 \\
    0.25 & 28.45 & 0.206 & 30.40 & 0.814 & 51.20 & \textbf{0.236} & \textbf{30.19} & \textbf{0.776}  \\
    0.75 & \textbf{23.02} & \textbf{0.187} & \textbf{31.22} & \textbf{0.839} & 73.76 & 0.287 & 30.47 & 0.750  \\
    1.5 & 32.35 & 0.243 & 30.80 & 0.792 & 82.26 & 0.335 & 30.29 & 0.705 \\
    2.0 & 40.33 & 0.284 & 30.41 & 0.750 & 86.48 & 0.358 & 30.17 & 0.683 \\
    \bottomrule
    \end{tabular}
    }
    \resizebox{0.8\textwidth}{!}{
    \begin{tabular}{ccccccccc}
    \toprule 
    \multirow{3}{*}{$\lambda$} &\multicolumn{4}{c}{Inpainting-\textit{Center}, $\sigma_y=0.05$} &\multicolumn{4}{c}{Denoising, $\sigma_y=0.05$} \\
    \cmidrule(lr){2-5}\cmidrule(lr){6-9}
    & FID $\downarrow$ & LPIPS $\downarrow$ & PSNR $\uparrow$ &  SSIM $\uparrow$ & FID $\downarrow$ & LPIPS $\downarrow$ & PSNR $\uparrow$ &  SSIM $\uparrow$ \\
    \midrule
    0.1 & 15.71 & 0.155 & 31.74 & 0.840 & 12.47 & 0.085 & 32.24 & 0.907 \\
    0.25 & 15.56 & 0.155 & 31.73 & 0.839 & 11.80 & 0.083 & 32.36 & 0.908 \\
    0.75 & 13.31 & 0.139 & 32.65 & 0.857 & 8.43 & 0.062 & 33.65 & 0.932 \\
    1.5 & \textbf{12.18} & \textbf{0.119} & 33.97 & 0.881 & 6.11 & 0.041 & 35.34 & 0.958 \\
    2.0 & 12.87 & \textbf{0.119} & \textbf{34.19} & \textbf{0.886} & \textbf{6.02} & \textbf{0.041} & \textbf{35.64} & \textbf{0.964} \\
    \bottomrule
    \end{tabular}
    }
\end{table*}

\begin{table*}[thb!]
    \centering
    \caption{Hyperparameter search for RED-Diff on face-blurred 
 ImageNet-$64\times64$ with $\sigma_y=0$. We use learning rate of $0.01$.}
    \label{tab:reddiff-imagenet-64-sigma-0-sr-gb}
    \resizebox{0.8\textwidth}{!}{
    \begin{tabular}{ccccccccc}
    \toprule 
    \multirow{3}{*}{$\lambda$} &\multicolumn{4}{c}{SR $2\times$, $\sigma_y=0$} &\multicolumn{4}{c}{Gaussian deblur, $\sigma_y=0$} \\
    \cmidrule(lr){2-5}\cmidrule(lr){6-9}
    & FID $\downarrow$ & LPIPS $\downarrow$ & PSNR $\uparrow$ &  SSIM $\uparrow$ & FID $\downarrow$ & LPIPS $\downarrow$ & PSNR $\uparrow$ &  SSIM $\uparrow$ \\
    \midrule
    0.1 &  \textbf{11.74} & 0.224 & 30.12 & 0.798 & \textbf{15.39} & \textbf{0.134} & \textbf{31.99} & \textbf{0.879} \\
    0.25 & 12.65 & \textbf{0.130} & \textbf{32.34} & \textbf{0.886} & 29.56 & 0.236 & 30.19 & 0.776 \\
    0.75 & 20.36 & 0.187 & 31.22 & 0.839 & 55.43 & 0.287 & 30.47 & 0.750 \\
    1.5 & 33.13 & 0.243 & 30.80 & 0.792 & 71.64 & 0.335 & 30.29 & 0.705 \\
    2.0 & 41.56 & 0.288 & 30.46 & 0.752 & 78.55 & 0.358 & 30.22 & 0.685 \\
    \bottomrule
    \end{tabular}
    }
    \resizebox{0.45\textwidth}{!}{
    \begin{tabular}{ccccc}
    \toprule 
    \multirow{3}{*}{$\lambda$} &\multicolumn{4}{c}{Inpainting-\textit{Center}, $\sigma_y=0$} \\
    \cmidrule(lr){2-5}
    & FID $\downarrow$ & LPIPS $\downarrow$ & PSNR $\uparrow$ &  SSIM $\uparrow$ \\
    \midrule
    0.1 & \textbf{7.29} & \textbf{0.079} & \textbf{39.14} & \textbf{0.925} \\
    0.25 & 7.40 & 0.155 & 31.73 & 0.839 \\
    0.75 & 8.47 & 0.083 & 38.59 & 0.922 \\
    1.5 & 10.75 & 0.095 & 37.42 & 0.916 \\
    2.0 & 12.54 & 0.119 & 34.19 & 0.886 \\
    \bottomrule
    \end{tabular}
    }
\end{table*}

\begin{table*}[thb!]
    \centering
    \caption{Hyperparameter search for RED-Diff on face-blurred ImageNet-$128\times128$ with $\sigma_y=0.05$. We use learning rate of $0.1$.}
    \label{tab:reddiff-imagenet-128-sigma-0.05-sr-gb}
    \resizebox{0.8\textwidth}{!}{
    \begin{tabular}{ccccccccc}
    \toprule 
    \multirow{3}{*}{$\lambda$} &\multicolumn{4}{c}{SR 2$\times$, $\sigma_y=0.05$} &\multicolumn{4}{c}{Gaussian deblur, $\sigma_y=0.05$} \\
    \cmidrule(lr){2-5}\cmidrule(lr){6-9}
    & FID $\downarrow$ & LPIPS $\downarrow$ & PSNR $\uparrow$ &  SSIM $\uparrow$ & FID $\downarrow$ & LPIPS $\downarrow$ & PSNR $\uparrow$ &  SSIM $\uparrow$ \\
    \midrule
    0.1 & 23.25 & 0.272 & 30.12 & 0.731 & 37.83 & 0.42 & 28.54 & 0.473 \\
    0.75 & 14.56 & 0.224 & 30.71 & 0.782 & \textbf{21.43} & \textbf{0.229} & 31.41 & 0.807 \\
    1.5 & \textbf{10.54} & \textbf{0.182} & 31.82 & 0.852 & 22.85 & 0.247 & \textbf{31.65} & \textbf{0.809} \\
    2.0 & 11.65 & 0.187 & \textbf{31.93} & \textbf{0.859} & 24.71 & 0.259 & 31.61 & 0.802 \\
    \bottomrule
    \end{tabular}
    }
    \resizebox{0.8\textwidth}{!}{
    \begin{tabular}{ccccccccc}
    \toprule 
    \multirow{3}{*}{$\lambda$} &\multicolumn{4}{c}{Inpainting-\textit{Center}, $\sigma_y=0.05$} &\multicolumn{4}{c}{Denoising, $\sigma_y=0.05$} \\
    \cmidrule(lr){2-5}\cmidrule(lr){6-9}
    & FID $\downarrow$ & LPIPS $\downarrow$ & PSNR $\uparrow$ &  SSIM $\uparrow$ & FID $\downarrow$ & LPIPS $\downarrow$ & PSNR $\uparrow$ &  SSIM $\uparrow$ \\
    \midrule
    0.1 & 19.68 & 0.191 & 31.75 & 0.795 & 12.83 & 0.134 & 32.27 & 0.854 \\
    0.75 & 19.03 & 0.202 & 31.36 & 0.779 & 12.69 & 0.14 & 32.09 & 0.846 \\
    1.5 & 16.33 & 0.189 & 31.81 & 0.794 & 10.67 & 0.121 & 32.89 & 0.874 \\
    2.0 & \textbf{14.63} & \textbf{0.171} & \textbf{32.42} & \textbf{0.819} & \textbf{9.19} & \textbf{0.105} & \textbf{33.52} & \textbf{0.895} \\
    \bottomrule
    \end{tabular}
    }
\end{table*}

\begin{table*}[thb!]
    \centering
    \caption{Hyperparameter search for RED-Diff on face-blurred 
 ImageNet-$128\times128$ with $\sigma_y=0$. We use learning rate of $0.1$.}
    \label{tab:reddiff-imagenet-128-sigma-0-sr-gb}
    \resizebox{0.8\textwidth}{!}{
    \begin{tabular}{ccccccccc}
    \toprule 
    \multirow{3}{*}{$\lambda$} &\multicolumn{4}{c}{SR 2$\times$, $\sigma_y=0$} &\multicolumn{4}{c}{Gaussian deblur, $\sigma_y=0$} \\
    \cmidrule(lr){2-5}\cmidrule(lr){6-9}
    & FID $\downarrow$ & LPIPS $\downarrow$ & PSNR $\uparrow$ &  SSIM $\uparrow$ & FID $\downarrow$ & LPIPS $\downarrow$ & PSNR $\uparrow$ &  SSIM $\uparrow$ \\
    \midrule
    0.1 & \textbf{3.90} & \textbf{0.082} & \textbf{34.47} & \textbf{0.922} & \textbf{4.19} & \textbf{0.085} &\textbf{34.68} & \textbf{0.929} \\
    0.75 & 6.52 & 0.105 & 33.54 & 0.905 & 12.59 & 0.177 & 32.71 & 0.864 \\
    1.5 & 10.46 & 0.142 & 32.98 & 0.894 & 19.29 & 0.225 & 32.15 & 0.831  \\
    2.0 & 13.08 & 0.165 & 32.65 & 0.884 & 22.57 & 0.245 & 31.94 & 0.816 \\
    \bottomrule
    \end{tabular}
    }
    \resizebox{0.8\textwidth}{!}{
    \begin{tabular}{ccccccccc}
    \toprule 
    \multirow{3}{*}{$\lambda$} &\multicolumn{4}{c}{Inpainting-\textit{Center}, $\sigma_y=0$} &\multicolumn{4}{c}{Inpainting-\textit{Freeform}, $\sigma_y=0$} \\
    \cmidrule(lr){2-5}\cmidrule(lr){6-9}
    & FID $\downarrow$ & LPIPS $\downarrow$ & PSNR $\uparrow$ &  SSIM $\uparrow$ & FID $\downarrow$ & LPIPS $\downarrow$ & PSNR $\uparrow$ &  SSIM $\uparrow$ \\
    \midrule
    0.1 & \textbf{5.39} & \textbf{0.068} & \textbf{38.91} & \textbf{0.928} & \textbf{8.94} & \textbf{0.162} & \textbf{35.54} & \textbf{0.830} \\
    0.75 & 5.52 & 0.073 & 38.11 & 0.924 & 9.26 & 0.166 & 35.05 & 0.826 \\
    1.5 & 6.09 & 0.079 & 37.32 & 0.920 & 10.13 & 0.172 & 34.58 & 0.821 \\
    2.0 & 6.68 & 0.083 & 36.87 & 0.917 & 10.87 & 0.176 & 34.30 & 0.818 \\
    \bottomrule
    \end{tabular}
    }
\end{table*}

\begin{table*}[thb!]
    \centering
    \caption{Hyperparameter search for RED-Diff on AFHQ with $\sigma_y=0.5$. We use learning rate of $0.1$.}
    \label{tab:reddiff-afhq-sigma-0.05-sr-gb}
    \resizebox{0.8\textwidth}{!}{
    \begin{tabular}{ccccccccc}
    \toprule 
    \multirow{3}{*}{$\lambda$} &\multicolumn{4}{c}{SR $4\times$, $\sigma_y=0.05$} &\multicolumn{4}{c}{Gaussian deblur, $\sigma_y=0.05$} \\
    \cmidrule(lr){2-5}\cmidrule(lr){6-9}
    & FID $\downarrow$ & LPIPS $\downarrow$ & PSNR $\uparrow$ &  SSIM $\uparrow$ & FID $\downarrow$ & LPIPS $\downarrow$ & PSNR $\uparrow$ &  SSIM $\uparrow$ \\
    \midrule
    0.1 & 21.59 & 0.385 & 29.51 & 0.607 & 17.36 & 0.379 & 29.95 & 0.639  \\
    0.25 & 22.47 & 0.374 & 29.66 & 0.635 & \textbf{15.81} & \textbf{0.341} & \textbf{30.15} & \textbf{0.645} \\
    0.75 & \textbf{20.84} & \textbf{0.331} & \textbf{29.97} & \textbf{0.675} & 25.41 & 0.366 & 29.76 & 0.588 \\
    1.5 & 22.46 & 0.355 & 29.68 & 0.642 & 38.66 & 0.409 & 29.34 & 0.525 \\
    2.0 & 25.02 & 0.376 & 29.49 & 0.618 & 45.01 & 0.427 & 29.18 & 0.500 \\
    \bottomrule
    \end{tabular}
    }
    \resizebox{0.8\textwidth}{!}{
    \begin{tabular}{ccccccccc}
    \toprule 
    \multirow{3}{*}{$\lambda$} &\multicolumn{4}{c}{Inpainting-\textit{Center}, $\sigma_y=0.05$} &\multicolumn{4}{c}{Denoising, $\sigma_y=0.05$} \\
    \cmidrule(lr){2-5}\cmidrule(lr){6-9}
    & FID $\downarrow$ & LPIPS $\downarrow$ & PSNR $\uparrow$ &  SSIM $\uparrow$ & FID $\downarrow$ & LPIPS $\downarrow$ & PSNR $\uparrow$ &  SSIM $\uparrow$ \\
    \midrule
    0.1 & \textbf{28.39} & 0.216 & 31.53 & 0.756 & 8.32 & 0.159 & 32.18 & 0.827 \\
    0.25 & 28.85 & 0.217 & 31.51 & 0.755 & 8.35 & 0.161 & 32.16 & 0.826 \\
    0.75 & 28.80 & 0.218 & 31.64 & 0.759 & 7.94 & 0.156 & 32.35 & 0.833 \\
    1.5 & 28.74 & 0.205 & 32.19 & 0.784 & 6.63 & 0.138 & 33.12 & 0.862 \\
    2.0 & 28.55 & 0.190 & 32.63 & 0.802 & 5.71 & 0.124 & 33.70 & 0.882 \\
    2.5 & 28.71 &\textbf{0.177} & \textbf{32.99} & \textbf{0.818} & \textbf{4.93} & \textbf{0.111} & \textbf{34.18} & \textbf{0.899} \\
    \bottomrule
    \end{tabular}
    }
\end{table*}

\begin{table*}[thb!]
    \centering
    \caption{Hyperparameter search for RED-Diff on AFHQ with $\sigma_y=0$. We use learning rate (lr) of $0.1$ unless mentioned otherwise.}
    \label{tab:reddiff-afhq-sigma-0-sr-gb}
    \resizebox{0.8\textwidth}{!}{
    \begin{tabular}{ccccccccc}
    \toprule 
    \multirow{3}{*}{$\lambda$} &\multicolumn{4}{c}{SR $4\times$, $\sigma_y=0$} &\multicolumn{4}{c}{Gaussian deblur, $\sigma_y=0$} \\
    \cmidrule(lr){2-5}\cmidrule(lr){6-9}
    & FID $\downarrow$ & LPIPS $\downarrow$ & PSNR $\uparrow$ &  SSIM $\uparrow$ & FID $\downarrow$ & LPIPS $\downarrow$ & PSNR $\uparrow$ &  SSIM $\uparrow$ \\
    \midrule
    0.005 & 11.67 & 0.197 & \textbf{32.93} & \textbf{0.837} & 14.69 & 0.278 & \textbf{31.73} & \textbf{0.760} \\
    0.05 & \textbf{8.65} & \textbf{0.191} & 32.21 & 0.801 & 11.67 & \textbf{0.268} & 31.30 & 0.731 \\
    0.1 & 9.65 & 0.204 & 31.84 & 0.781 & \textbf{11.53} & 0.273 & 31.05 & 0.711 \\
    0.25 & 11.65 & 0.222 & 31.53 & 0.768 & 13.22 & 0.293 & 30.63 & 0.675 \\
    0.75 & 14.98 & 0.274 & 30.72 & 0.726 & 23.34 & 0.351 & 29.91 & 0.598 \\
    1.5 & 19.40 & 0.332 & 29.95 & 0.665 & 36.96 & 0.402 & 29.39 & 0.529 \\
    2.0 & 22.72 & 0.361 & 29.65 & 0.632 & 43.64 & 0.422 & 29.22 & 0.504 \\
    \bottomrule
    \end{tabular}
    }
    \resizebox{0.8\textwidth}{!}{
    \begin{tabular}{ccccccccc}
    \toprule 
    \multirow{3}{*}{$\lambda$} &\multicolumn{4}{c}{Inpainting-\textit{Center}, $\sigma_y=0$, lr=$0.01$} &\multicolumn{4}{c}{Inpainting-\textit{Freeform}, $\sigma_y=0$} \\
    \cmidrule(lr){2-5}\cmidrule(lr){6-9}
    & FID $\downarrow$ & LPIPS $\downarrow$ & PSNR $\uparrow$ &  SSIM $\uparrow$ & FID $\downarrow$ & LPIPS $\downarrow$ & PSNR $\uparrow$ &  SSIM $\uparrow$ \\
    \midrule
    0.005 & \textbf{8.53} & \textbf{0.050} & \textbf{38.89} & \textbf{0.951} & 7.22 & 0.091 & \textbf{36.89} & \textbf{0.892} \\
    0.05 & 8.53 & 0.050 & 38.89 & 0.951 & 7.27 & \textbf{0.090} & 36.88 & \textbf{0.892} \\ 
    0.1 & 8.53 & 0.050 & 38.88 & 0.951 & \textbf{7.23} & 0.091 & 36.82 & 0.891 \\
    0.25 & 8.53 & 0.050 & 38.83 & 0.950 & 7.32 & 0.094 & 36.69 & 0.889 \\
    0.75 & 8.88 & 0.056 & 38.60 & 0.948 & 7.74 & 0.102 & 36.26 & 0.884  \\
    1.5 & 10.32 & 0.071 & 38.04 & 0.942 & 8.41 & 0.112 & 35.69 & 0.877 \\
    2.0 & 11.62 & 0.084 & 37.54 & 0.937 & 8.76 & 0.119 & 35.37 & 0.872 \\
    \bottomrule
    \end{tabular}
    }
\end{table*}

%% file: sections/additional_background.tex
\clearpage
\section{Additional Background}
In this section, we follow the notation used in the prior work by ~\citet{flowmatching}.  
\paragraph{Continuous Normalizing Flows (CNFs).} A Continuous Normalizing Flow ~\citep{chen2018neural} is a time-dependent diffeomorphic map $\phi_t: [0, 1] \times \mathbb{R}^d \rightarrow \mathbb{R}^d$ that is defined by the ODE:
\begin{equation}
    \dfrac{d}{dt} \phi_t(\vx) = \vv_t(\phi_t(\vx)) ;\quad
    \phi_0(\vx) = \vx \label{eq:cnf-ivp}
\end{equation}
where $\vx \in \mathbb{R}^d$ and $\vv_t: [0, 1] \times \mathbb{R}^d \rightarrow \mathbb{R}^d$ is a time-dependent vector field that is usually parametrized with a neural network. The generative process of a CNF involves sampling from a simple prior distribution $\vx_0 \sim p_0(\vx_0)$ (e.g. standard Gaussian distribution) and then solving the initial value problem defined by the ODE in \cref{eq:cnf-ivp} to obtain a sample from the target distribution $\vx_1 \sim p_1(\vx_1)$. Thus, a CNF reshapes a simple prior distribution $p_0$ to a more complex distribution $p_t$, via a push-forward equation based on the instantaneous change of variables formula. 
\begin{align}
    p_t &= [\phi]_* p_0 \\
    [\phi]_* p_0(\vx) &= p_0(\phi_t^{-1}(\vx)) \text{det}\left[\dfrac{\partial \phi_t^{-1}}{\partial \vx} (\vx)\right]
\end{align}
CNFs are usually trained by optimizing the maximum likelihood objective. As shown in ~\citet{chen2018neural}, the exact likelihood computation can be done via relatively cheap operations despite the  Jacobian term. However, this  requires restricting the architecture of the neural network to constrain the Jacobian term. FFJORD~\citep{ffjord} improves upon this by proposing a method that uses Hutchinson's trace estimator to compute log density, and allows CNFs with free-form Jacobians, thereby removing any restrictions on the  architecture. This approach has difficulties for high-dimensional images where the trace estimator is noisy.  Flow Matching provides an alternative, scalable approach to training CNFs with arbitrary architectures.

\paragraph{Flow Matching.} Suppose we have samples from an unknown data distribution $
\vx_1 \sim q(\vx_1)$. Let $p_t$ denote a probability path from the prior distribution $p_0$ to the data distribution $p_1$ that is approximately equal to $q$. Flow Matching loss is defined as 
\begin{equation}
    \mathcal{L}_{FM} = \E_{t, p_t(\vx)} \| \vv_t(\vx; \theta) - \vu_t(\vx) \|^2\label{eq:fm-loss}
\end{equation}
where $\vu_t(\vx)$ is a vector field that generates the probability path $p_t(\vx)$, and $\theta$ denotes trainable parameters of the CNF. In practice, we usually do not have any prior knowledge on $p_t$ and $u_t$, and thus this objective is intractable. Inspired by diffusion models, \citet{flowmatching} propose Conditional Flow Matching, where both the probability paths and the vector fields are conditioned on the sample $\vx_1 \sim q(\vx_1)$. The exact objective for Conditional Flow matching is given by 
\begin{equation}
    \mathcal{L}_{CFM} = \E_{t, q(\vx_1), p_t(\vx|\vx_1)} \| \vv_t(\vx; \theta) - \vu_t(\vx | \vx_1) \|^2\label{eq:cfm-loss}
\end{equation}
where, $p_t(\vx|\vx_1)$ denotes a conditional probability path, and $\vu_t(\vx|\vx_1)$ denotes the corresponding conditional vector field that generates the conditional probability path. Interestingly, both the loss objectives in \cref{eq:cfm-loss} and \cref{eq:fm-loss} have identical gradients w.r.t. $\theta$. More importantly, past research has proven that $\vu_t(\vx) = \mathbb{E}[\vu_t(\vx | \vx_1) | \vx_t = \vx]$.  The optimal solution to the conditional Flow Matching recovers $\vu_t(\vx)$ and therefore $\vv_t(\vx; \theta)$ generates the desired probability path $p_t(\vx)$.Thus, we can train a CNF without access to the  marginal vector field $\vu_t(\vx)$ or probability path $p_t(\vx)$. Compared to the prior approaches to train flow models, Flow Matching allows simulation-free training with unbiased gradients, and scales easily to high dimensions.